\documentclass{article}
\usepackage{arxiv}

\usepackage[utf8]{inputenc} 
\usepackage[T1]{fontenc}    
\usepackage{hyperref}       
\usepackage{url}            
\usepackage{booktabs}       
\usepackage{multicol}
\usepackage{amsfonts}       
\usepackage{nicefrac}       
\usepackage{microtype}      
\usepackage{lipsum}
\usepackage{graphicx}
\usepackage{subcaption}
\usepackage{siunitx}

\usepackage{threeparttable}
\usepackage{siunitx}
\usepackage{booktabs}
\usepackage{hyperref}
\usepackage{bm}
\usepackage{amsmath}
\usepackage{multirow}
\usepackage{tabularx}
\usepackage{array}

\usepackage[ruled,vlined]{algorithm2e}

\usepackage{xspace} 

\usepackage{cleveref}

\newtheorem{remark}{Remark}
\newcommand{\norm}[1]{\left\lVert #1 \right\rVert}
\usepackage{lineno}

\title{Single- to multi-fidelity history-dependent learning with uncertainty quantification and disentanglement: application to data-driven constitutive modeling}

\author{
  Jiaxiang Yi \\ Faculty of Mechanical Engineering\\
  Delft University of Technology\\
  Mekelweg 2, Delft, 2628 CD, The Netherlands \\
  \And
   Bernardo P. Ferreira \\
  School of Engineering\\
  Brown University\\
  184 Hope St., Providence, RI 02912, USA\\
  \And
  Miguel A. Bessa \\
  School of Engineering\\
  Brown University\\
  184 Hope St., Providence, RI 02912, USA \\
  \texttt{miguel$\_$bessa@brown.edu} \\
}

\begin{document}
\maketitle
\begin{abstract}

Data-driven learning is generalized to consider history-dependent multi-fidelity data, while quantifying epistemic uncertainty and disentangling it from data noise (aleatoric uncertainty). This generalization is hierarchical and adapts to different learning scenarios: from training the simplest single-fidelity deterministic neural networks up to the proposed multi-fidelity variance estimation Bayesian recurrent neural networks. The versatility and generality of the proposed methodology is demonstrated by applying it to different data-driven constitutive modeling scenarios that include multiple fidelities with and without aleatoric uncertainty (noise). The method accurately predicts the response and quantifies model error while also discovering the noise distribution (when present). This opens opportunities for future real-world applications in diverse scientific and engineering domains; especially, the most challenging cases involving design and analysis under uncertainty.

\end{abstract}


\section{Introduction}
\label{sec:introduction}

Data-driven learning focuses on the use of machine learning methods to learn non-trivial models from data. Although these models can include Physics constraints \cite{lagaris1998artificial,raissi2019physics}, in general, they require large quantities of data \cite{lecun2015deep}. Unfortunately, creating large datasets with \textbf{high-fidelity (HF)} data can be challenging, as it involves high-throughput or a vast number of parallel low-throughput experiments or simulations. By definition, HF datasets require costly simulations or experiments that are time-consuming or resource-intensive to conduct.\footnote{This article associates ``cost'' to time and other resources (computational or experimental) that are needed to run a computer simulation or a physical experiment, such that a sample can be collected into a dataset. High cost implies that it is time-consuming or resource-intensive to generate a large dataset.} Conversely, \textbf{low-fidelity (LF)} datasets can be generated faster but have higher error and can be stochastic (noisy). In practice, \textbf{multi-fidelity (MF)} datasets may be necessary to balance the advantages and disadvantages of each fidelity.

Without loss of generality, the accuracy-efficiency paradox addressed by MF modeling will be illustrated in this article by focusing on data-driven constitutive modeling of materials. Although the idea of using neural networks as material models can be traced back to the early 90s \cite{Ghaboussi1991}, it remained largely dormant for more than two decades. However, the advent of high-performance automatic differentiation libraries and high-throughput data generation strategies, such as simulations of \textbf{representative volume elements (RVEs)}\footnote{Recall that an RVE is a sufficiently large domain of a material that is representative of that material's mechanical behavior, i.e., it leads to the same response even after randomizing the microstructure \cite{Bessa2017}.} of materials, were shown to create sufficiently large datasets for training accurate models \cite{Bessa2017}. This can be achieved by finite element analyses of thousands of RVEs when they are composed of elastic or hyperelastic materials \cite{le2015computational}, but it becomes more challenging when they undergo history-dependent phenomena such as plasticity because the simulations are slower \cite{Bessa2017}. A viable pathway is to consider faster simulation methods, e.g., the \textbf{self-consistent clustering analysis (SCA)} \cite{Liu2016_SCA}, at the expense of lower accuracy (lower fidelity). Once a sufficiently large dataset is available, recurrent neural networks have been found to learn the complete history-dependent constitutive behavior of materials \cite{Mozaffarplasticity2019}. Since then, several researchers followed data-driven strategies to learn constitutive models for a wide range of materials, e.g., \cite{wu:2020a,he:2022a, eghbalian:2023a, kalina:2025a, jadoon:2025a}, including their use to accelerate topology optimization \cite{vijayakumaran2024consistentmachinelearningtopology}.

\begin{figure}[h]
    \centering
    \includegraphics[width=0.95\textwidth]{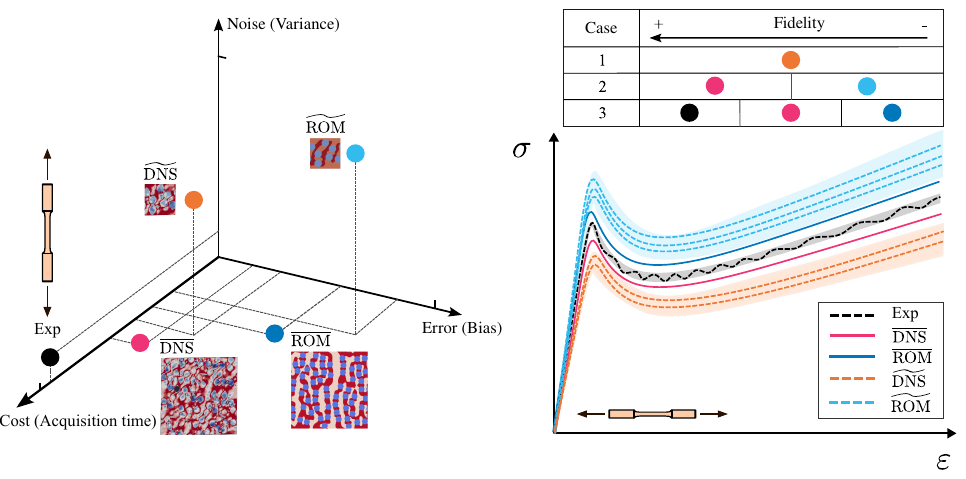}
    \caption{Schematic summarizing different data generation methods for discovering the constitutive behavior of materials. Each method and corresponding dataset from which it originates is organized according to cost, noise, and accuracy. A method or a dataset is considered to have \textbf{high-fidelity (HF)} if it has lower error than another method, which is considered to have \textbf{low-fidelity (LF)}. Experiments are usually the most expensive and slowest method to acquire data (high cost), but provide the lowest error, despite the presence of some noise (aleatoric uncertainty). A \textbf{deterministic direct numerical simulation ($\overline{\text{DNS}}$)} is obtained in this work by \textbf{finite element analysis (FEA)} of a \textbf{representative volume element (RVE)}, i.e., a sufficiently large material domain that generates noiseless data. A \textbf{stochastic DNS ($\widetilde{\text{DNS}}$)} is obtained by FEA of a \textbf{stochastic volume element (SVE)}, i.e., a smaller material domain that is faster to simulate but generates noisy data. A \textbf{deterministic reduced order model ($\overline{\text{ROM}}$)} is obtained by \textbf{self-consistent clustering analysis (SCA)} of an RVE (noiseless data with lower-fidelity than FEA but faster to acquire). And lastly, a \textbf{stochastic ROM ($\widetilde{\text{ROM}}$)} is obtained by SCA of an SVE (noisy data with the lowest fidelity and fastest acquisition time).
    }
    \label{fig:multi_fidelity_concept}
\end{figure}

\Cref{fig:multi_fidelity_concept} exemplifies five different methods of generating datasets with different fidelities, and organizes them according to three characteristics: error, cost, and noise. As indicated in the figure, data can be collected from physical experiments but also from synthetic experiments via deterministic (noiseless) or stochastic (noisy) computer simulations\footnote{Stochastic computer simulations can simulate a noisy response, e.g., by randomizing the microstructure in a material domain while keeping the same descriptors of geometry, constituent properties and boundary conditions.}. In addition, computer simulations can be divided into two types: \textbf{direct numerical simulations (DNSs)} and \textbf{reduced order models (ROMs)}.

Usually, the most accurate but least efficient method of data collection is to perform a real experiment of the material (indicated by a black circle in \Cref{fig:multi_fidelity_concept}). Data collected from experiments has low error, high cost, and low noise. Noise is illustrated as the shaded black average stress-strain response in the right part of \Cref{fig:multi_fidelity_concept}, and it arises from testing different specimens with the same geometry, properties, and boundary conditions. Experimental data is usually scarce and may not be available in sufficient quantity to train a neural network constitutive law \cite{ferreira2025automatically}.

In addition to conducting real experiments, performing a deterministic DNS (denoted as $\overline{\text{DNS}}$) of a material is the best way to collect high-fidelity data. Deterministic DNSs have low error, high cost, and no noise (no aleatoric uncertainty), and they aim to simulate the experimental response as closely as possible. \Cref{fig:multi_fidelity_concept} illustrates this type of simulation in magenta. A deterministic DNS is typically obtained by finite element analysis (FEA) of an RVE.

A stochastic DNS (denoted as $\widetilde{\text{DNS}}$) of a material provides a simple way to reduce cost when compared to a deterministic DNS because it considers a smaller material domain that is no longer representative. Such material domains are usually called \textbf{stochastic volume elements (SVEs)} because randomization of the microstructure leads to different responses (indicated in orange in \Cref{fig:multi_fidelity_concept}). Therefore, a stochastic DNS has noise and typically higher error than a deterministic DNS, but leads to faster data collection.

Typically, DNSs are not sufficiently efficient to create large enough databases when the material undergoes history-dependent phenomena like plasticity deformation, even when considering SVEs instead of RVEs. In that case, \textbf{reduced-order models (ROMs)} must be considered. In this article, we choose the ROM to be the previously mentioned SCA method \cite{Liu2016_SCA}, as it can predict the elastoplastic responses of RVEs and SVEs faster than the finite element method used as DNS. The SCA method and its adaptive formulation \cite{Ferreira2022} recently became open-source \cite{Ferreira2023}, so these (and all other) results obtained in this article can be fully replicated. By definition, a deterministic ROM (denoted as $\overline{\text{ROM}}$) is noiseless and leads to faster predictions than the corresponding deterministic DNS (less cost), but its predictions have a higher error. A stochastic ROM ($\widetilde{\text{ROM}}$) leads to the fastest predictions because it uses a fast simulation method (like SCA) for estimating the behavior of a small material domain, but introduces noise and is associated with the highest error. 

\textit{Summary of contribution.} This article generalizes data-driven learning to all types of datasets exemplified in \Cref{fig:multi_fidelity_concept}. This requires the ability to learn deterministically or probabilistically from datasets that are noiseless or noisy, single- or multi-fidelity, history-dependent or independent. Importantly, the ability to estimate both aleatoric and epistemic uncertainties\footnote{There are two types of uncertainties: epistemic (or model) uncertainty, and aleatoric uncertainty (or noise). Epistemic uncertainty quantifies the error of model predictions; therefore, it reduces as more data is used to train the model (for infinite data, there is no model error). Aleatoric uncertainty refers to the noise that is inherent to the data; therefore, a measurement of a given system under the same input conditions does not lead to the same output when there is aleatoric uncertainty (noise). Aleatoric uncertainty is irreducible because even if we had infinite data, we could not eliminate noise as it results from unknown factors that affect the measurements.} is also considered because data-driven models are difficult to interpret, and their trustworthiness improves with appropriate uncertainty quantification. Disentangling these uncertainties not only facilitates the updating of constitutive models based on experimental data \cite{ferreira2025automatically}, but also supports the design of robust materials by optimizing mechanical performance while accounting for data variation arising from unknown or uncontrolled manufacturing factors. This capability further enables scalable active learning and Bayesian optimization \cite{pmlr-v70-gal17a}, although this is not explored herein.

The proposed generalization requires merging three separate machine learning contributions: (1) scalable \textbf{Bayesian recurrent neural networks (BRNNs)}, (2) efficient uncertainty disentanglement, and (3) multi-fidelity neural network architectures trainable on large datasets. We have recently contributed to these fields separately \cite{yi2024practicalmultifidelitymachinelearning,yi2025cooperativebayesianvariancenetworks} with the aim of merging them into a general learning paradigm, as presented herein.

\textit{Related work on uncertainty quantification.} Uncertainty quantification has become an important topic in machine learning \cite{Wilson2020, Abdar2021, Psaros2023}, as it facilitates decision-making and design optimization \cite{Bessasuppercompressible2019,Shin, Wang2020}. Initially, most solutions relied on Gaussian process regression \cite{williams2006gaussian}, and assumed homoscedastic aleatoric uncertainty (constant noise). Even when considering heteroscedastic noise \cite{Ozbayram2024}, the main bottleneck of Gaussian processes is their lack of scalability due to the ``curse of dimensionality''. In principle, Bayesian neural networks (BNNs) do not have this limitation but their adoption in Engineering practice has been slow because Bayesian inference is challenging \cite{Neal2011,papamarkou2024position}. Various approximate Bayesian inference methods have been proposed, such as Markov chain Monte Carlo (MCMC) sampling \cite{Neal2012}, Variational Inference (VI) \cite{Blei2017}, MC-Dropout \cite{Gal2016}, and Deep Ensembles \cite{Lakshminarayanan2017}. This has enabled the recent application of BNNs to cases without history-dependency, such as estimating material properties \cite{Olivier2021_UQ_Laws, Pasparakis2024} and learning hyperelastic constitutive models \cite{linka2025discovering}. However, enabling uncertainty quantification of history-dependent or path-dependent phenomena, such as material plasticity, requires BRNNs.

\textit{Related work on Bayesian recurrent neural networks and uncertainty disentanglement.} Literature on BRNNs is scarce, but the interested reader is referred to \cite{Chatzis2015, Fortunato2017, Gan2017} for early work on the topic. Two recent methodologies have improved their scalability \cite{yi2025cooperativebayesianvariancenetworks,coscia2025barnn}. In this work, we use the method we developed -- \textbf{Variance estimation Bayesian recurrent neural networks (VeBRNNs)} \cite{yi2025cooperativebayesianvariancenetworks} -- because it is also able to separate epistemic and aleatoric uncertainties. VeBRNNs are scalable because inference is performed by \textbf{preconditioned stochastic gradient Langevin dynamics (pSGLD)} \cite{Li2015}, and they effectively disentangle uncertainties due to the cooperative training of a variance estimation network \cite{nix1994estimating} and a BRNN. To the best of our knowledge, this method is the first that is capable of learning history-dependent constitutive behavior, estimating noise (aleatoric uncertainty), and predicting model (epistemic) uncertainty for small or large single-fidelity datasets.

\textit{Related work on multi-fidelity machine learning.} Reliable data-driven constitutive laws require a substantial amount of HF data \cite{Bessa2017}, but gathering it is often time-consuming and impractical. As previously referred, this is particularly relevant when involving complex experiments or expensive simulations \cite{fare2022multi}. Effective MF approaches can mitigate this challenge by leveraging the useful information embedded in LF data, thereby reducing the dependence on HF data \cite{Park2017_MF_review, Liu2018_multioutput}. MF machine learning has been successfully applied in various domains, such as solving expensive partial differential equations \cite{Meng2020_MFPINN}, optimization \cite{Li2020_MFNNBO, cheng2022enhanced}, and material design \cite{Shin}. Initially, MF machine learning was limited to data scarce scenarios and low-dimensional problems by considering MF Gaussian process regression \cite{williams2006gaussian, Forrester2007, Meng2021_MFBNN}. However, the literature has adopted deterministic methods when dealing with large datasets, e.g., using MF recurrent neural networks \cite{Conti2023}, or continual learning of subnetworks \cite{Dekhovich2023} for transferring knowledge without forgetting. The seamless integration of MF Bayesian machine learning for history-dependent problems remains unexplored. Moreover, existing studies do not explicitly address stochastic data in LF datasets, and its repercussions on the HF predictions.

\section{Proposed datasets: four different single- and multi-fidelity learning scenarios}
\label{sec:datasets}

\begin{figure}[h]
    \centering
    \begin{subfigure}[b]{0.25\textwidth}
        \centering
        \includegraphics[height=4.8cm]{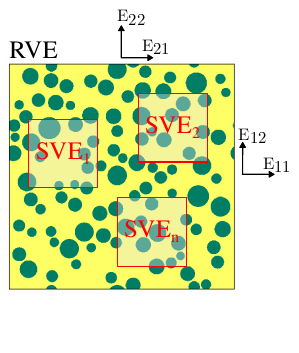}
        \caption{RVE vs SVE}
        \label{fig:rve_vs_sve}
    \end{subfigure}
    \begin{subfigure}[b]{0.33\textwidth}
        \centering
        \includegraphics[height=4.8cm]{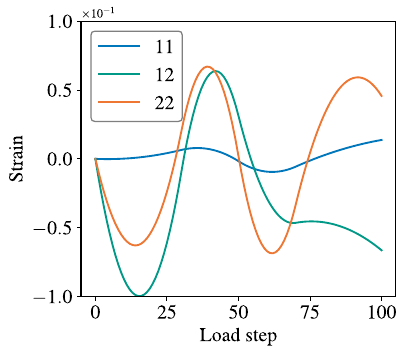}
        \caption{Strain inputs}
        \label{fig:strain_inputs}
    \end{subfigure}
    \begin{subfigure}[b]{0.33\textwidth}
        \centering
        \includegraphics[height=4.8cm]{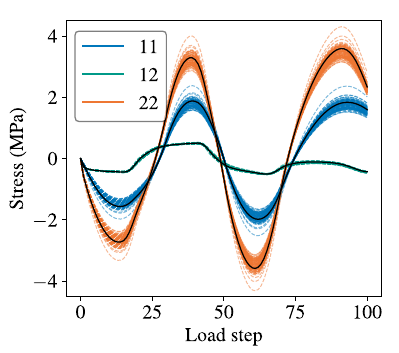}
        \caption{History-dependent stress outputs}
        \label{fig:stress_outputs}
    \end{subfigure}
    \caption{RVE and SVE strain-stress path data:  (a) illustration of biphasic material RVE and three SVEs; (b) example of random polynomial strain paths for the three strain components; and (c) comparison between the RVE (deterministic) homogenized stress response shown in solid lines, and the stochastic stress responses of 100 SVEs shown in dashed lines.}
    \label{fig:data_description}
\end{figure}

We created four different scenarios based on \Cref{fig:multi_fidelity_concept} with the aim of illustrating the generality of the proposed work and to motivate future investigations on this topic by the research community. The datasets for all problems are obtained from simulations that capture history-dependent behavior of material volume elements (RVEs or SVEs for deterministic or stochastic data, respectively). The biphasic material volume elements (\Cref{fig:rve_vs_sve}) are subjected to synthetic random polynomial strain paths (\Cref{fig:strain_inputs}), from which the stress response is obtained (\Cref{fig:stress_outputs}). The biphasic material consists of:
\begin{itemize}
    \item a matrix phase (yellow area of \Cref{fig:rve_vs_sve}) characterized by a von Mises elastoplastic model with isotropic strain hardening with  elastic modulus $E_\text{matrix} = 100 \, \text{MPa}$, Poisson’s ratio $\nu_\text{matrix} = 0.3$, and a von Mises hardening law defined as $\sigma_{y} = 0.5 + 0.5(\Bar{\epsilon}^p)^{0.4}$, where $\Bar{\epsilon}^p$ denotes the accumulated plastic strain; 
    \item and elastic particles (green circles of \Cref{fig:rve_vs_sve}) characterized by a linear elastic isotropic model with $E_\text{fiber} = 15\,\text{MPa}$ and $\nu_\text{fiber} = 0.19$. 
\end{itemize}

As seen in \Cref{fig:rve_vs_sve}, an SVE can be viewed as a subdomain of an RVE (although we still assume periodic microstructures for both). Different SVE realizations obtained by randomizing the microstucture (\Cref{fig:rve_vs_sve}) lead to different average stress responses (dashed lines in \Cref{fig:stress_outputs}), even when subjecting the SVEs to the same input deformation path (\Cref{fig:strain_inputs}).\footnote{Readers unfamiliar with homogenization theory are referred to \cite{Bessa2017, Mozaffarplasticity2019}, where it is explained how to convert an average strain state into a periodic boundary value problem of an RVE or SVE, and then calculate the corresponding average stress state.} In contrast, randomization of the microstructure of an RVE still leads to the same homogenized stress response (see the solid lines in \Cref{fig:stress_outputs}, embedded in the dashed lines corresponding to different realizations of the SVE).

Without loss of generality, all learning problems in this article involve sequence-to-sequence regression where the input sequence is the average strain history applied to the material volume element and the output prediction is the average stress response for that path. Therefore, a nonlinear and history-dependent constitutive model is obtained by training a neural network on strain-stress paths such that:
\begin{equation}
    \bm{\sigma}_{:}  = f(\bm{\varepsilon}_{:}, \bm{\alpha}),
\end{equation}
\noindent where $\bm{\varepsilon}_{:}$ is the strain path, $\bm{\alpha}$ is a set of internal variables (e.g., weights and biases of a neural network $f$), and $\bm{\sigma}_{:}$ is the predicted stress path.

In the inference stage, the model recurrently predicts a stress state $\bm{\sigma}$ for a given input strain state $\bm{\varepsilon}$ (see \cite{Mozaffarplasticity2019}). Additional details on the generation of input strain paths are found in Appendix \ref{sec:strain_stress_paths}, and on the origins of aleatoric uncertainty in this context are in \Cref{fig:data_description} and \Cref{sec:represent_aleatoric_unc}.

\begin{figure}[h]
    \centering
    \begin{subfigure}[t]{0.4\textwidth}
        \centering
        \includegraphics[width=\textwidth]{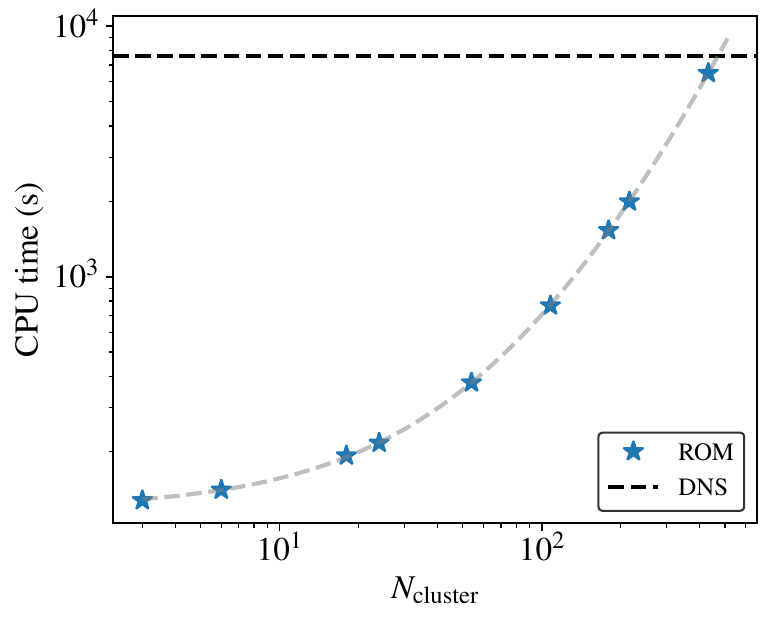}
        \caption{ }
        \label{fig:sca_dns_time}
    \end{subfigure}
    \hspace{0.01\textwidth}
    \begin{subfigure}[t]{0.4\textwidth}
        \centering
        \includegraphics[width=\textwidth]{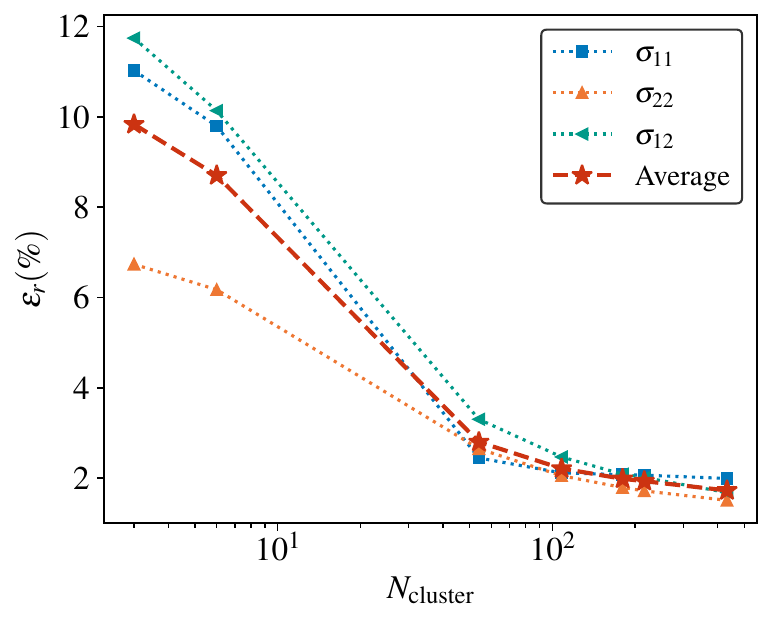}
        \caption{ }
        \label{fig:sca_dns_error}
    \end{subfigure}
    \caption{Performance comparison between FEA-based DNS and SCA in the computational homogenization of biphasic microstructure RVE. The accuracy and computational cost of the SCA method depend on the number of clusters.}
    \label{fig:sca_vs_dns_time_error_linear_solver}
\end{figure}

We also recall that in addition to considering an RVE or an SVE, we can also choose different simulation methods (FEA or SCA). We ensure that the FEA simulations have a sufficiently fine mesh to be considered DNSs, but we use different discretization levels for the SCA to achieve different accuracy-efficiency trade-offs (creating different fidelities from these ROMs). In SCA, the number of clusters is denoted by $N_\text{cluster}$, and it controls the trade-off between accuracy and computational efficiency compared to the DNS (FEA is employed as DNS). A comparison between accuracy and efficiency between $\overline{\text{ROM}}$ and $\overline{\text{DNS}}$ is presented in \Cref{fig:sca_vs_dns_time_error_linear_solver}. The computational cost of SCA scales quadratically with $N_\text{cluster}$, so a larger number of clusters improves accuracy but increases computational expense. For instance, if we use only three clusters, we obtain a large stress prediction error $\epsilon_r$ of approximately $10\%$, yet the corresponding CPU time is very low (almost negligible). Conversely, employing $N_\text{cluster}=512$ achieves an average $\epsilon_r$ of around $2\%$, i.e., slightly less accurate than the DNS obtained by FEA but already having comparable computational cost (that is why we choose FEA as DNS).

In summary, we generate different datasets by considering RVEs or SVEs, and simulating them with the FEA or SCA method. These four choices are used to create six different datasets with different fidelities, as presented in \Cref{tab:mf_datasets_info}. Importantly, we introduce the concept of cost ratio to properly evaluate the computational cost of generating a single random strain-stress path for each dataset when compared to the most time consuming simulation, $\overline{\text{DNS}}$, which takes approximately $\SI{8000}{s}$ to obtain a path in a platform equipped with an AMD Ryzen™ 9 7945HX processor (1 core used), 16 GB RAM. The table also includes the number of paths used for training, for \textbf{in-distribution (ID)} testing, and for \textbf{out-of-distribution (OOD)} testing. The ID testing paths were considered using an applied strain range of $\left(\varepsilon_{ij, \text{min}}, \varepsilon_{ij,\text{max}}\right) = (-0.1, 0.1), \;\; \{i,j\}=1,2$.  The OOD testing paths were obtained considering a wider strain range of $\left(\varepsilon_{ij, \text{min}}, \varepsilon_{ij,\text{max}}\right) = (-0.125, 0.125), \;\; \{i,j\}=1,2$. This means that the OOD paths include $25\%$ extrapolation beyond the training domains. Also note that for the testing paths of stochastic cases (SVEs), we repeat the simulations 100 times (indicated as $\times 100$ in \Cref{tab:mf_datasets_info}) -- this is necessary to have a meaningful estimation of the ground truth aleatoric uncertainty associated to that fidelity, so that we can then compare with the predictions obtained by our models.

We created four different scenarios inspired by real-world applications and by considering data generated by one or two of the six fidelities shown in \Cref{tab:mf_datasets_info}:

\begin{itemize}
    \item \textbf{Dataset 1 (stochastic single fidelity): $\widetilde{\text{DNS}}$} – The only single fidelity dataset considered in this article is created to mimic the \textit{stochastic} experimental case. Therefore, we obtain this dataset by FEA of an SVE \footnote{'An SVE' does not imply a fixed microstructure across the dataset; rather, each strain path corresponds to a random microstructure realization.}, leading to a noisy DNS dataset (recall the tilde on top indicates the data has noise).
    \item \textbf{Dataset 2 (deterministic LF and deterministic HF): $\overline{\text{ROM}}$+$\overline{\text{DNS}}$} – A common scenario in the deterministic MF machine learning literature, where both fidelities are noiseless. HF data is obtained by FEA of an RVE, while LF data is obtained by SCA of the same RVE (recall that the bar on top indicates the data is noiseless).
    \item \textbf{Dataset 3 (stochastic LF and stochastic HF): $\widetilde{\text{ROM}}$+$\widetilde{\text{DNS}}$} – MF dataset where both fidelities are noisy. HF data is the same as in Dataset 1 (DNS of an SVE), while LF data is obtained by SCA of an SVE.
    \item \textbf{Dataset 4 (stochastic LF and deterministic HF): $\widetilde{\text{ROM}}$+$\overline{\text{DNS}}$} - MF dataset where the HF is noiseless and the LF is noisy. This is also a common scenario, as HF data tends to have less (or negligible) noise when compared to LF data.
\end{itemize}

\begin{table}[h]
    \centering
    \caption{Data collected with six different fidelity levels according to the choice of material volume element (RVE vs. SVE), simulation method (DNS vs. ROM) and discretization level (number of clusters used for ROM). The bar on top indicates that data is noiseless (obtained from an RVE). The tilde on top indicates that data is noisy (obtained from an SVE). ID means in-distribution, and OOD means out-of-distribution. Cost ratio indicates the approximate time it takes to simulate one stress-strain path compared to a deterministic DNS. Data types are ordered according to computational cost (decreasing cost from left to right).}
    \resizebox{\textwidth}{!}{%
        \begin{threeparttable}
            \begin{tabular}{m{3.3cm}  m{1cm} m{1.7cm} m{2cm} m{2cm} m{2cm} m{2cm} }
                \toprule
                Fidelity       & $\overline{\text{DNS}}$       & $\widetilde{\text{DNS}}$    & $\overline{\text{ROM}}$ ($N_\text{cluster} = 18$) & $\overline{\text{ROM}}$ ($N_\text{cluster} = 3$) & $\widetilde{\text{ROM}}$ ($N_\text{cluster} = 18$) & $\widetilde{\text{ROM}}$ ($N_\text{cluster} = 3 $) \\
                \midrule
                Cost ratio ($c$)     & 1.0           & $1/20$     & $1/36$                           & $1/60$                            & $1/120$                          & $1/200$                            \\
                Volume element    & RVE & SVE & RVE    & RVE                      & SVE  & SVE                          \\
                Simulation method    & FEA & FEA & SCA    & SCA                      & SCA  & SCA                          \\
                Noise (aleatoric uncert.)   & No & Yes & No                    & No                     & Yes                       & Yes                         \\
                \# Training paths    & 1000          & 2000       & 2000                             & 2000                              & 2000                             & 2000                               \\
                \# ID testing paths  & 100           & 100 ($\times 100$)  & n/a                                & n/a                                 & n/a                                & n/a                                  \\
                \# OOD testing paths & 100           & 100 ($\times 100$)  & n/a                                & n/a                                 & 100 ($\times 100$)                        & 100 ($\times 100$)                          \\

                \bottomrule
            \end{tabular}
        \end{threeparttable}}
    \label{tab:mf_datasets_info}
\end{table}

\section{Proposed methodology}
\label{sec:methodology}

\begin{figure}[h]
    \centering
    \includegraphics[width=0.55\textwidth]{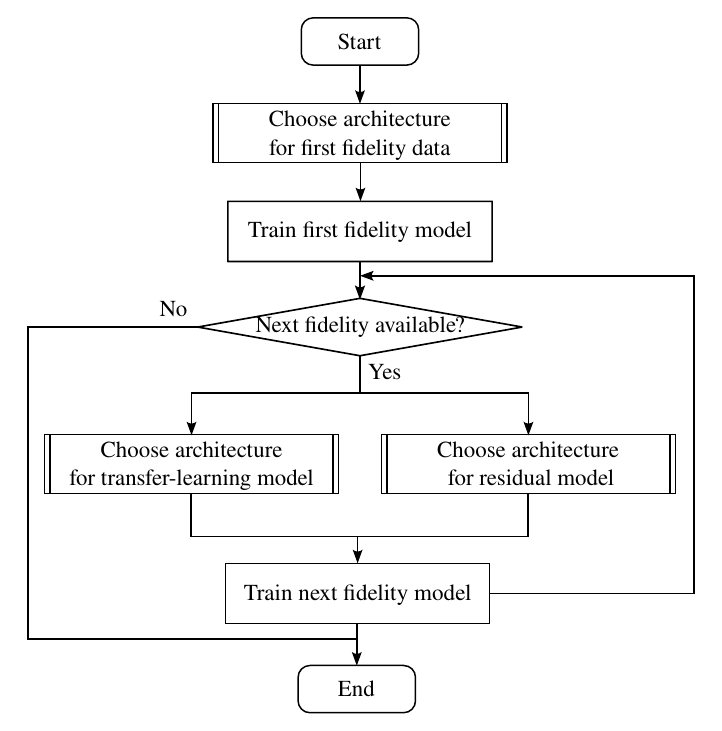}
    \caption{Flowchart of the proposed practical multi-fidelity framework. } 
    \label{fig:training_flowchart}
\end{figure}

We aim to generalize data-driven learning to both single- and multi-fidelity problems, with or without history-dependency, and with the ability to disentangle uncertainties (if needed). Therefore, we created a flowchart (\Cref{fig:training_flowchart}) that highlights the modular nature of the framework, hoping to facilitate navigation among possible choices. In the remainder of the article, we focus on the most general choices, and omit the simpler cases for brevity (those already exist in the literature, as we saw in the Introduction).

\begin{figure}[h]
    \centering
    \includegraphics[width=0.99\textwidth]{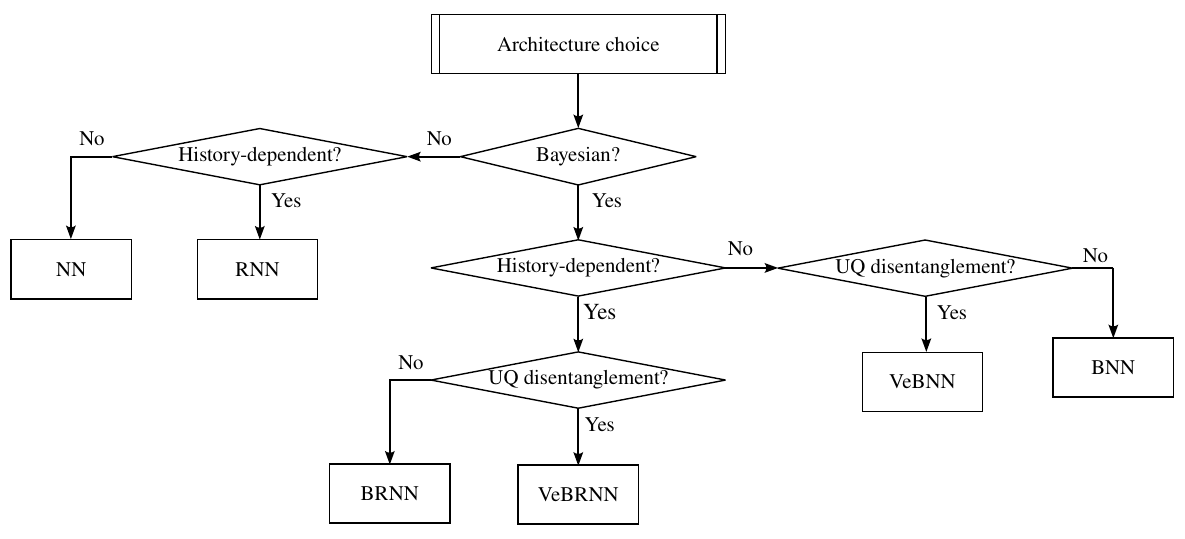}
    \caption{Flowchart for choosing a neural network architecture that learns from data of given fidelity. UQ is an abbreviation for uncertainty quantification.}
    \label{fig:architecture_flowchart}
\end{figure}

\subsection{Neural network architecture choice}
\label{sec:cooperative_uq_training}

\Cref{fig:architecture_flowchart} presents a flowchart for choosing a neural network architecture that learns from data of a given fidelity. The most general architecture, VeBRNN, can be simplified into all other networks. As mentioned above, VeBRNNs are capable of single-fidelity history-dependent predictions with uncertainty quantification and disentanglement. This architecture is trained cooperatively between a variance estimation network and a Bayesian recurrent neural network, enabling multidimensional predictions of (1) mean response, (2) variance of aleatoric uncertainty (noise), and (3) variance of epistemic (model) uncertainty. The method is explained in detail in \cite{yi2025cooperativebayesianvariancenetworks}, but we summarize it here for completeness.

Assuming stress observations at each pseudo-time $t$ (or load step) follow a Gaussian distribution, we can formulate the problem as follows,

\begin{equation}
\label{eq:problem_setup}
    f_j(\mathbf{x}_t) \sim \mathcal{N}\left(y_j(\mathbf{x}_{t}) , s_j^2(\mathbf{x}_{t}) \right) \quad \text{where} \; \quad t= 1,...,T,
\end{equation}

where $y_j(\mathbf{x})$ is the ground truth mean of the output component $j$  (note that $\mathbf{y} \equiv \bm{\sigma}$ and that $j=1,2,3$ because the stress has three components\footnote{From \Cref{sec:datasets}, $\bm{\sigma}$ has three components for two-dimensional problems under plane strain conditions: ${\sigma}_{11}$, ${\sigma}_{12}$, and ${\sigma}_{2}$.}), and where $s_j^2(\mathbf{x})$ is the variance of the aleatoric uncertainty of the appropriate output component (i.e., the covariance matrix of the Gaussian is diagonal). If $s_j^2(\mathbf{x})$ depends on the input $\mathbf{x}$ (strain components $\bm{\varepsilon}$), it is referred as heteroscedastic aleatoric uncertainty; otherwise, it is homoscedastic. \Cref{fig:data_description} and \Cref{fig:aleatoric_uncertainty_sca} clearly show that SVEs have heteroscedastic aleatoric uncertainty (noise).

\Cref{alg:CUQalgorithm} contains the pseudo-code for the cooperative training of the VeBRNN. The algorithm involves training a deterministic RNN in Step 1, then a variance estimation RNN for step 2, and a BRNN in Step 3.\footnote{If the data is history-independent, each network of the 3 steps does not need recurrent units. So, the mean estimation would be with a feedforward neural network (FNN), variance estimation also with an FNN, and final prediction with epistemic uncertainty with a BNN.}. If no aleatoric uncertainty estimation is needed, then Step 2 can be skipped. If no epistemic uncertainty is required, then step 3 can be skipped (and the architecture is simply a deterministic RNN -- Step 1). \Cref{sec:cooperative_uq_training} describes the 3 steps, but a complete description is also provided in the original publication \cite{yi2025cooperativebayesianvariancenetworks}. \Cref{sec:DNS-SVE dataset experiments} presents results obtained from training a VeBRNN on Dataset 1 (obtained by DNS of an SVE).

\begin{algorithm}[h]
    \SetAlgoLined
    \caption{Cooperative training of VeBRNN}
    \label{alg:CUQalgorithm}
    \KwData{Mean RNN network $f(\mathbf{x}; \bm{\theta})$, variance RNN network  $s^2(\mathbf{x}; \bm{\phi})$, training data $\mathcal{D} = \{\mathbf{X}, \mathbf{y}\}$ number of iterations $K$}
    \KwResult{Optimal parameters for variance and Bayesian mean RNN networks}
    \textbf{Step 1: Mean network training:}\\
    \quad \emph{Deterministic training:} find mean estimation by minimizing the MSE loss \Cref{eq:mse_loss} using \emph{Adam} optimizer. \\
    \For{$i=1$ \textbf{to} $K$}{
        \textbf{Step 2: Variance network training (Aleatoric uncertainty)}:\\
        \quad \emph{Deterministic training:} find data variance estimation  by minimizing \Cref{eq:gamma_loss} for fixed mean from Step 1 or Step 3 using \emph{Adam} optimizer.\\
        \textbf{Step 3: Bayesian network training (Epistemic uncertainty)}:\\
        \quad \emph{Bayesian inference:} obtain posterior distribution estimation from \Cref{eq:prediction_equation} to update mean and epistemic variance estimates for fixed aleatoric variance from Step 2 using \emph{pSGLD} sampler illustrated in \Cref{sec:psgld}.\\
        \quad Compute the log marginal likelihood of $\texttt{LMglk}[i]$
    }
    Identify the optimal models by $i^* = \arg\max_{i} \texttt{LMglk}[i]$.
\end{algorithm}

We note that Step 1 and Step 2 involve deterministic training of neural networks, but the respective loss functions are different. Step 1 results from minimizing the conventional mean squared error of the average response (standard training in regression tasks). Step 2 results from training another RNN to learn the standard deviation of the aleatoric uncertainty after the mean response has been determined in Step 1. However, the loss function in Step 2 is different from Step 1, as we derived it from assuming a Gamma distribution of the square of the residual, which can be demonstrated to lead to the variance of the aleatoric uncertainty \cite{yi2025cooperativebayesianvariancenetworks}. The sequential training of Step 1 and Step 2 was shown to stabilize training of the variance network (Step 2), addressing an important knowledge gap in the literature \cite{yi2025cooperativebayesianvariancenetworks}.

In addition, Step 3 concerns ``training'' a BRNN, which is not done by minimizing a loss function but by sampling the posterior distribution -- a process designated by Bayesian inference (\Cref{sec:Bayesian_inference_methods}). This step allows to update the mean response prediction (from Step 1, and for the aleatoric uncertainty predicted in Step 2). More importantly, it also allows to estimate the uncertainty associated to the mean prediction, i.e., it determines the epistemic or model uncertainty. There are many strategies to perform Bayesian inference, and we thoroughly investigated them for different regression problems in \cite{yi2025cooperativebayesianvariancenetworks}. We demonstrated that pSGLD is robust and scalable, and recommend its choice for regression problems involving smooth and unimodal probability distribution functions, such as Gaussian and Gamma distributions.

\subsection{Multi-fidelity framework}
\label{sec:general_description}

In a recent article, we proposed a unification of the MF machine learning literature into a practical MF framework \cite{yi2024practicalmultifidelitymachinelearning}. We postulated that MF machine learning can be expressed through the general formulation,
\begin{equation}
    \label{eq:general_mf_formula}
    f^h(\mathbf{x}) = g(f^l(\mathbf{x}), \mathbf{x}) + r(\mathbf{x}),
\end{equation}

\noindent where  $f^h(\mathbf{x})$ is the HF model, $f^l(\mathbf{x})$ is the LF model, $g(\mathbf{x})$ is a transfer learning model, and $r(\mathbf{x})$ is a residual model. We demonstrate the capability of handling problems of different dimensionality and dataset sizes by selecting suitable machine learning models. In the original work, we adopted a deterministic feedforward neural network for the LF model, a linear transfer learning model (linear regression), and a Bayesian neural network for the residual model. This was demonstrated to be effective in finding the homogenized properties of a biphasic microstructure material (without history-dependency).

To the best of our knowledge, RNNs have not yet been considered in a MF setting. Therefore, in this work, we investigate how to incorporate architectures with recurrent units in the above-mentioned MF framework. As we have to consider inputs that arrive as sequences (contain history), the MF sequence dataset $\mathcal{D}$ consist of HF inputs $\bm{X}^{h}_t \in \mathbb{R}^{N^h \times d^{in}}$ and outputs $\bm{Y}^{h}_t \in \mathbb{R}^{N^h \times d^{out}}$, as well as LF inputs $\bm{X}^{l}_t \in \mathbb{R}^{N^l \times d^{in}}$ and outputs $\bm{Y}^{l}_t \in \mathbb{R}^{N^l \times d^{out}}$. Here, $N^h$ and $N^l$ are the number of HF and LF sequences, respectively, while $d^{in}$ and $d^{out}$ represent input and output dimensions, and $t$ indicates time (or pseudo-time) increment in the sequence. Accordingly, we rewrite \Cref{eq:general_mf_formula} to explicitly account for time-series data as

\begin{equation} \label{eq:general_framework}
    f^h(\mathbf{x}^h_t) = g(f^l(\mathbf{x}^h_t), \mathbf{x}_t^h) + r(\mathbf{x}^h_t);  \quad t=1,\cdots, T \;,
\end{equation}

\noindent where $f^h(\mathbf{x})$, $f^l(\mathbf{x})$, $g(\mathbf{x})$, and $r(\mathbf{x})$ are models for the HF, LF, transfer learning, and residual, respectively, including recurrent units.\footnote{We consider two fidelity levels (HF and LF) for the convenience of illustration. The framework can naturally be extended to accommodate problems involving more than two fidelity levels. This is shown in \Cref{fig:training_flowchart} by recurrently adding new fidelities.}

\begin{figure}[h]
    \centering
    \includegraphics[width=0.45\textwidth]{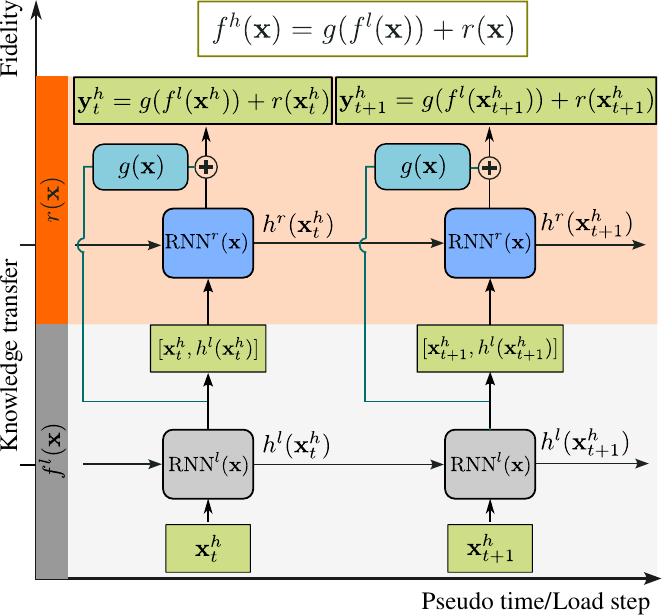}
    \caption{Schematic of the proposed multi-fidelity recurrent neural network architectures for predicting uncertainty-aware history-dependent plasticity.}
    \label{fig:proposed_framework}
\end{figure}

\Cref{eq:general_framework} looks deceptively simple, but it accepts many architectural choices to relate LF and HF data. For example, the literature often considers a linear transfer learning model $g$ (effectively not doing transfer learning), or uses a neural network as transfer learning without considering an additional residual model $r$.  We carefully investigated different possibilities and included in \Cref{sec:mf_model_comparison} four solutions for comparison. For brevity, the main text only includes the best solution -- illustrated in \Cref{fig:proposed_framework}.

The MF model in \Cref{fig:proposed_framework} can be regarded as a particular realization of  \Cref{fig:training_flowchart}. The LF model is trained first (independently) using an RNN for the LF data, then the linear transfer learning model $g(\mathbf{x})$ is trained, and finally another RNN is trained to learn the residual $r(\mathbf{x})$ using HF data. Therefore, there are three models that need to be defined in a MF prediction: $f^l(\mathbf{x})$, $g(\mathbf{x})$, and $r(\mathbf{x})$. These models can be deterministic or Bayesian, depending on whether uncertainty quantification is desired or not. Also note that \Cref{fig:proposed_framework} should be interpreted along two dimensions: one corresponding to the time increment (from left to right), and the other to the fidelity level (from bottom to top). This structure enables the model to capture both the temporal evolution of material responses and transfer useful information across different fidelity levels, facilitating accurate predictions.

Interestingly, we found that transferring the hidden states from the LF model to the HF model was more effective than directly transferring the LF outputs (i.e., the stresses that are decoded at the last layer of the LF RNN, denoted as $\text{RNN}^l$ in \Cref{fig:proposed_framework}). This is reflected in \Cref{fig:proposed_framework} by noticing that the input to the residual model, which is also an RNN (denoted as $\text{RNN}^r$), is the hidden state of the LF model $h^l(\mathbf{x}_t^h)$ together with the input HF data $\mathbf{x}^h$, instead of using the decoded output $\mathbf{y}^l$ of the $\text{RNN}^l$. While this may seem like a subtle difference, we found it to be significant -- see \Cref{sec:mf_model_comparison}. Moreover, we also found that using a linear transfer learning model (i.e., $g(\mathbf{x})$ is simply an identity map in \Cref{fig:proposed_framework}) together with the referred RNN for the residual model (or a Bayesian version like a VeBRNN) is marginally better than using an RNN transfer learning model and removing the residual model -- see \Cref{sec:mf_model_comparison}. We did not find any benefit in using expressive models for both transfer learning and residual learning. Nevertheless, we do not claim that \Cref{fig:proposed_framework} represents the best MF model for all possible datasets. \Cref{sec:mf_model_comparison} simply shows that it is the most effective solution for the four datasets that we consider in this article, and that we discuss in the next sections.

\begin{remark}
\label{rmk:coop_training_rmk}
For simplicity, this subsection and \Cref{fig:proposed_framework} refer to RNNs. However, as shown in the next sections, any other architecture with recurrent units can be used. Namely, the VeBRNN architecture that we described previously (see \Cref{fig:training_flowchart} and \Cref{alg:CUQalgorithm}). The VeBRNN architecture,
\begin{itemize}
    \item when applied only to the LF model (replacing $\text{RNN}^l$ by $\text{VeBRNN}^l$), enables uncertainty quantification and disentanglement for LF predictions.
    \item when applied only to the residual learning model (replacing $\text{RNN}^r$ by $\text{VeBRNN}^r$), enables uncertainty quantification and disentanglement for HF predictions.
    \item when applied to both models, enables uncertainty quantification and disentanglement for both fidelities.
\end{itemize}
\end{remark}

\subsection{MF framework with different neural network architectures}
\label{sec:flowchart}

We expect that a fully Bayesian MF model using VeBRNNs is applicable to a wide range of scenarios, but we also ablate that model to consider progressively simpler architectural choices. We believe that the four above-mentioned datasets (\Cref{sec:datasets}) illustrate different scenarios encountered in Engineering practice. \Cref{tab:model_outputs} summarizes the different models, from simpler (top) to more complex (bottom) -- confront with \Cref{fig:training_flowchart} and \Cref{fig:architecture_flowchart}.

\begin{table}[hbt!]
\centering
\caption{Summary of architectural choices for single- and multi-fidelity datasets. See \Cref{fig:training_flowchart} for understanding single- and multi-fidelity training with neural network architectures chosen according to \Cref{fig:architecture_flowchart}. Notation: ``LF model''+``HF model''}
\label{tab:model_outputs}
\begin{small}
\begin{tabular}{ll ccc ccc}
\toprule
 \multirow{2}{*}{Model} & \multirow{2}{*}{Comment} & \multicolumn{3}{c}{LF predictions} & \multicolumn{3}{c}{HF predictions} \\
\cmidrule(lr){3-5} \cmidrule(lr){6-8}
 & & Mean & Aleatoric & Epistemic & Mean & Aleatoric & Epistemic \\
\midrule
 RNN       & Deterministic (single-fidelity) & & & & \checkmark & & \\
 VeBRNN   & Bayesian  (single-fidelity) & & & & \checkmark & \checkmark & \checkmark \\
 RNN+RNN & Deterministic LF \& HF     & \checkmark & & & \checkmark & & \\
 VeBRNN+RNN & Bayesian LF, deterministic HF          & \checkmark & \checkmark & \checkmark & \checkmark & & \\
 RNN+VeBRNN  & Deterministic LF, Bayesian HF      & \checkmark & & & \checkmark & \checkmark & \checkmark \\
 VeBRNN+VeBRNN & Bayesian LF \& HF                    & \checkmark & \checkmark & \checkmark & \checkmark & \checkmark & \checkmark \\
\bottomrule
\end{tabular}
\end{small}
\end{table}

\section{Results}
\label{sec:numerical_experiments}

We analyze the proposed framework (\Cref{fig:proposed_framework}) on four datasets (\Cref{sec:datasets}), and compare the performance of different model architectures (\Cref{tab:model_outputs}) while considering the total computational cost of data generation as, 
\begin{equation}
    \label{eq:total_cost}
    T_c  = N_\text{train}^hc^h +  N_\text{train}^{l}c^l
\end{equation}
where $N_\text{train}^h$ and $N_\text{train}^l$ are the number of training paths for HF and LF, respectively; $c^h$ and $c^l$ are the corresponding cost ratios of HF and LF, which can be found in \Cref{tab:mf_datasets_info}.

Recall that performance evaluation for Bayesian models involves more than one metric, as opposed to evaluating deterministic models. We use five accuracy metrics summarized in \Cref{tab:evaluation_metrics}. A more detailed description of these metrics is provided in \Cref{sec:performance_metrics}. Note that the ground truth mean in the proposed dataset is obtained from $\overline{\text{DNS}}$ (noiseless FEA simulations of an RVE subjected to different strain paths), which is the highest fidelity data that we have available. The confidence intervals are all defined for $\alpha=0.05$, i.e. we use 95\% confidence intervals for all experiments. Hyperparameters used for every model are listed in \Cref{sec:hyper_params}. Additional results are presented in \Cref{sec:additional_experiments}, and we will refer to this appendix when appropriate. 

\begin{table}[h]
    \centering
    \small
   \caption{Performance metrics considered in this work. The last column indicates the desired trend of the metric (e.g., relative error should decrease, PICP should tend to $1-\alpha=95\%$, and test log-likelihood should increase). }
    \label{tab:evaluation_metrics}
    \begin{tabular}{m{6cm}lm{5cm}l}
        \toprule
        \textbf{Metric}                                                    & \textbf{Abbreviation} & \textbf{Description}                                                                                   & \textbf{Trend}           \\
        \midrule
        Relative error (\cite{Dekhovich2023})                                       & $\epsilon_r$                 & Relative error between mean prediction and ground truth.                                               & $\downarrow$             \\
        Test Log-Likelihood (\cite{Deshpande2024})                         & TLL                   & Test log-likelihood of ground truth mean under the predicted epistemic distribution.                  & $\uparrow$               \\
        Wasserstein Distance (\cite{Kantorovich1960})                      & WA                    & Wasserstein distance between predicted aleatoric and ground truth distributions.                      & $\downarrow$             \\
        Prediction Interval Coverage Probability (\cite{Sluijterman_2024}) & PICP                  & Ratio of ground truth mean within the predicted epistemic interval under given confidence level. & $\rightarrow (1-\alpha)$ \\
        Mean Prediction Interval Width (\cite{Sluijterman_2024})           & MPIW                  & Mean width of predicted epistemic uncertainty interval.                                        & $\downarrow$             \\
        \bottomrule
    \end{tabular}
\end{table}

\subsection{Results for Dataset 1: $\widetilde{\text{DNS}}$ (stochastic single fidelity problem from FEA of an SVE) }
\label{sec:DNS-SVE dataset experiments} 

This first dataset includes aleatoric uncertainty and only considers one fidelity ($\widetilde{\text{DNS}}$). It is introduced to demonstrate the ability of the VeBRNN to predict mean, aleatoric uncertainty variance, and epistemic uncertainty variance. This contrasts with the widely used RNNs that only predict the mean response, as they are deterministic models. See \Cref{tab:model_outputs} for a summary of model capabilities.

\begin{figure}[h]
    \centering
    \hspace*{0.04\textwidth} 
    \begin{subfigure}[t]{0.33\textwidth}
        \centering
        \includegraphics[width=\textwidth]{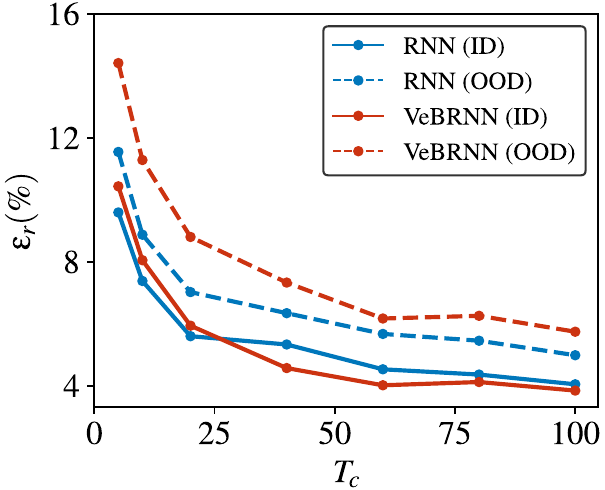}
        \caption{Mean estimate metric ($\downarrow$)}
        \label{fig:dns_sve_a}
    \end{subfigure}
    \begin{subfigure}[t]{0.33\textwidth}
        \centering
        \includegraphics[width=\textwidth]{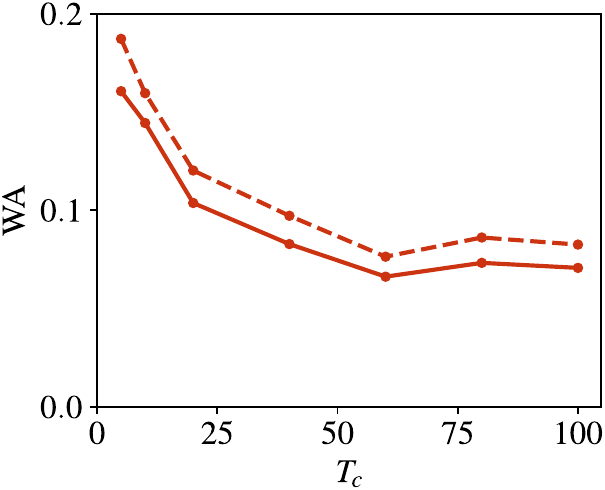}
        \caption{Aleatoric uncertainty metric ($\downarrow$)}
        \label{fig:dns_sve_b}
    \end{subfigure}\\
    \begin{subfigure}[t]{0.33\textwidth}
        \centering
        \includegraphics[width=\textwidth]{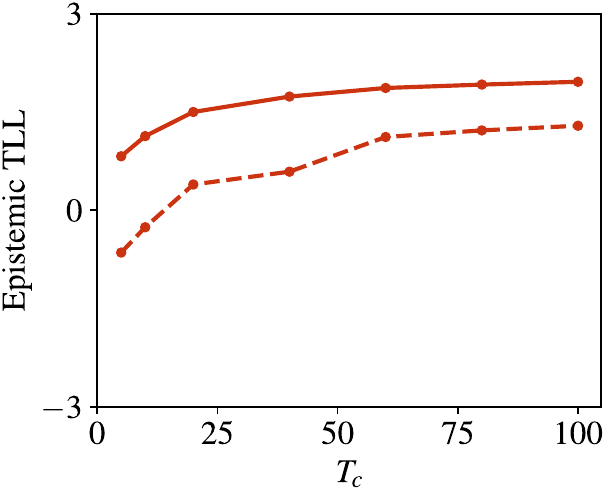}
        \caption{Epistemic uncertainty metric ($\uparrow$)}
        \label{fig:dns_sve_c}
    \end{subfigure}
    \begin{subfigure}[t]{0.33\textwidth}
        \centering
        \includegraphics[width=\textwidth]{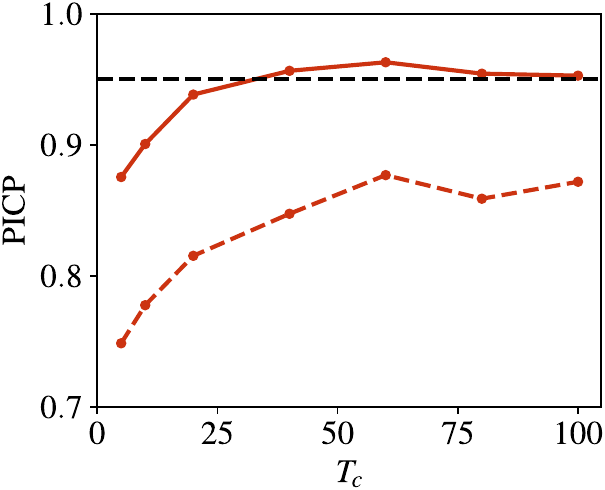}
        \caption{Epistemic uncert. metric ($\rightarrow 0.95$)}
        \label{fig:dns_sve_d}
    \end{subfigure}
    \begin{subfigure}[t]{0.33\textwidth}
        \centering
        \includegraphics[width=\textwidth]{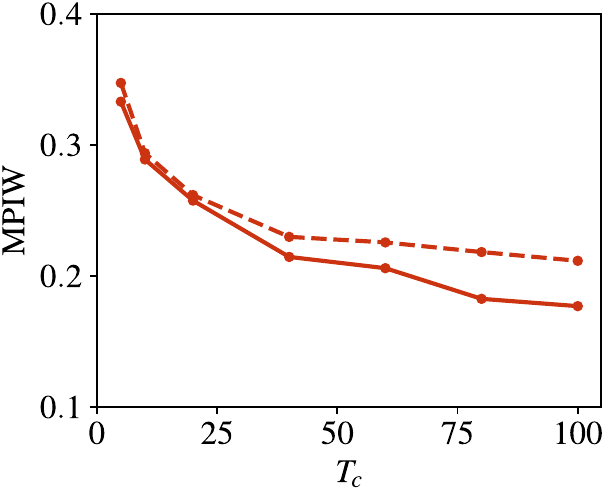}
        \caption{Epistemic uncertainty metric ($\downarrow$)}
        \label{fig:dns_sve_e}
    \end{subfigure}
    \caption{Comparison of test performance metrics on Dataset 1 (single-fidelity problem) based on $\widetilde{\text{DNS}}$ training paths. In this figure, $T_c = 1$ corresponds to 20 training paths, and the horizontal dashed line in (d) indicates the desired confidence level $0.95$. Testing data is considered both in-distribution (ID), and out-of-distribution (OOD).}
    \label{fig:sve_dns_results_comparison}
\end{figure}

As indicated in \Cref{tab:mf_datasets_info}, this dataset has up to 2000 training paths (all of which are considered HF, as there is no LF data). In the results below, we consider experiments with an increasing number of training paths from 100 to 2000. Accordingly, $T_c$ increases from 5 to 100, following the relation $T_c = N^h_\text{train}\times \frac{1}{20} + 0$. The corresponding results are shown in  \Cref{fig:sve_dns_results_comparison}. Note that we also have access to the noiseless data ($\overline{\text{DNS}}$) that we can use to precisely assess the error of this dataset for the relative error $\epsilon_r$ and Epistemic TLL.

\begin{remark}
    \label{rmk:remark single-fidelity}
    The single-fidelity case trivializes the flowchart in \Cref{fig:training_flowchart}, reducing to training a single RNN (or VeBRNN). 
\end{remark}

\begin{figure}[h!]
    \centering
    \begin{subfigure}[t]{0.35\textwidth}
        \centering
\includegraphics[width=\textwidth]{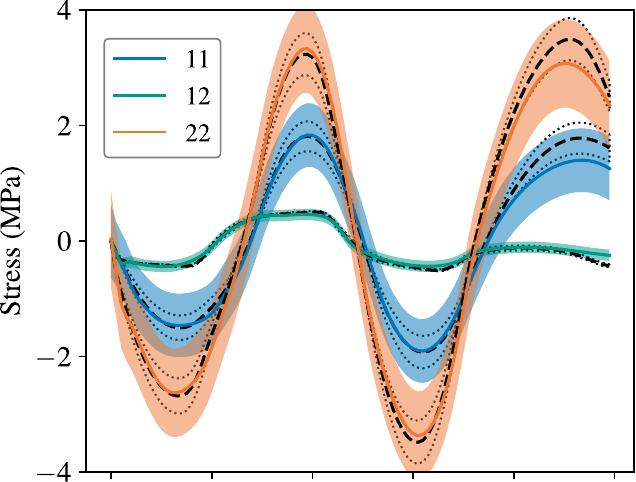}
        \caption{$T_c=5$ (100  $\widetilde{\text{DNS}}$ paths)}
        \label{fig:psgld_id_test_path_a}
    \end{subfigure}
    \hspace{0.02\textwidth}
    \begin{subfigure}[t]{0.33\textwidth}
        \centering
        \includegraphics[width=\textwidth]{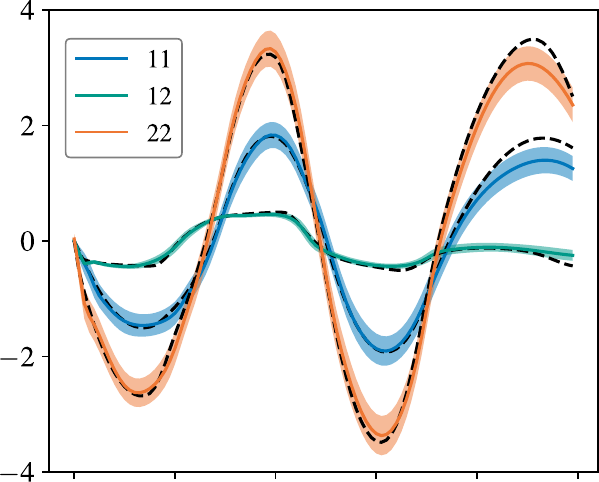}
        \caption{$T_c=5$ (100  $\widetilde{\text{DNS}}$ paths)}
        \label{fig:psgld_id_test_path_b}
    \end{subfigure}

    \vspace{0.3cm}

    \begin{subfigure}[t]{0.35\textwidth}
        \centering
        \includegraphics[width=\textwidth]{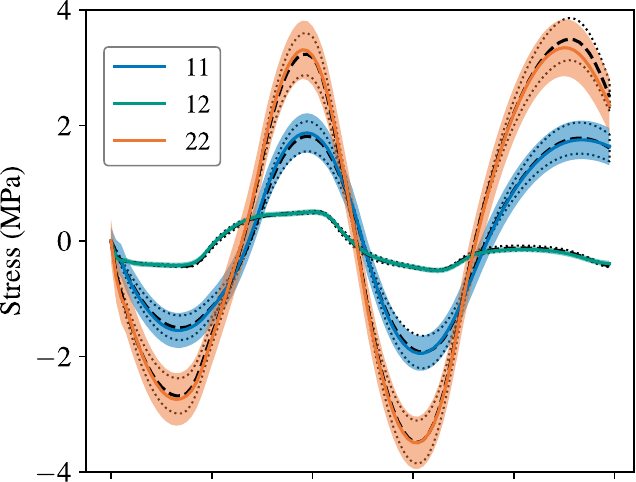}
        \caption{$T_c=40$ (800  $\widetilde{\text{DNS}}$ paths)}
        \label{fig:psgld_id_test_path_c}
    \end{subfigure}
    \hspace{0.02\textwidth}
    \begin{subfigure}[t]{0.33\textwidth}
        \centering
        \includegraphics[width=\textwidth]{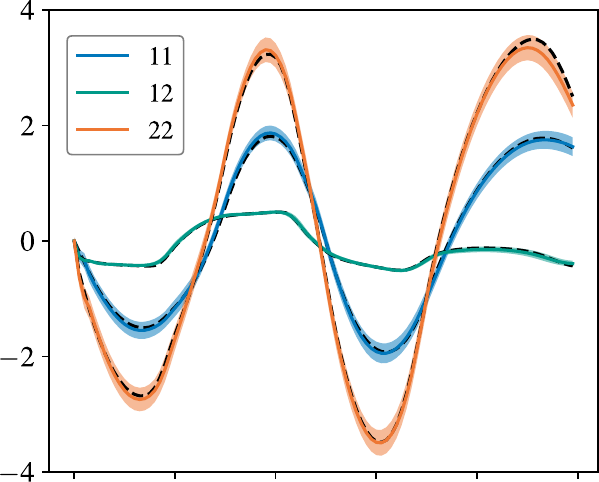}
        \caption{$T_c=40$ (800  $\widetilde{\text{DNS}}$ paths)}
        \label{fig:psgld_id_test_path_d}
    \end{subfigure}

    \vspace{0.3cm}

    \begin{subfigure}[t]{0.35\textwidth}
        \centering
        \includegraphics[width=\textwidth]{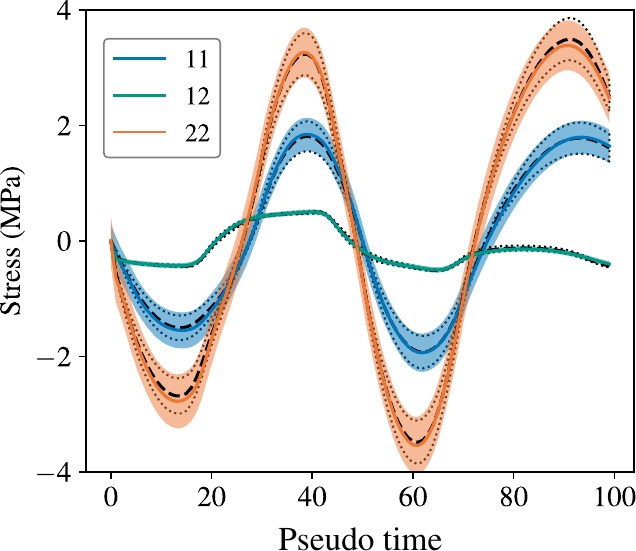}
        \caption{$T_c=100$ (2000  $\widetilde{\text{DNS}}$ paths)}
        \label{fig:psgld_id_test_path_e}
    \end{subfigure}
    \hspace{0.02\textwidth}
    \begin{subfigure}[t]{0.33\textwidth}
        \centering
        \includegraphics[width=\textwidth]{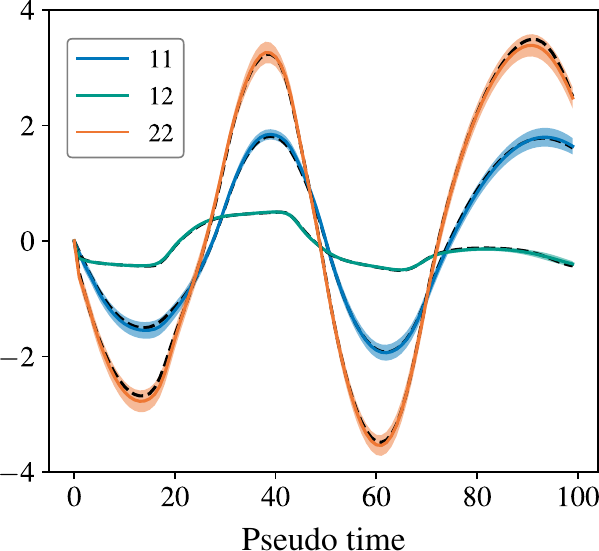}
        \caption{$T_c=100$ (2000  $\widetilde{\text{DNS}}$ paths)}
        \label{fig:psgld_id_test_path_f}
    \end{subfigure}
    
    \vspace{0.4cm}
    \begin{minipage}{\textwidth}
        \centering
        \makebox[0.35\textwidth]{VeBRNN Aleatoric uncertainty}%
        \hspace{0.02\textwidth}
        \makebox[0.33\textwidth]{VeBRNN Epistemic uncertainty}
    \end{minipage}
    
    \vspace{0.1cm} 
    \caption{Predictions of VeBRNN on an ID test path after training with 2000 paths ($T_c = 100$) from Dataset 1 ($\widetilde{\text{DNS}}$). The left column shows the comparison between 0.95 confidence intervals of the predicted aleatoric uncertainty (colored shaded areas) and the ground truth distributions (black dotted shaded areas); the right column gives the predicted epistemic uncertainty distributions. The ground truth mean is shown as black dashed lines in both columns.}
    \label{fig:psgld_id_test_path}
\end{figure}

As shown in \Cref{fig:sve_dns_results_comparison}, the relative error $\epsilon_{r}$ obtained on ID testing paths decreases as the number of training paths increases for both the deterministic model and the proposed VeBRNN method, as expected. The VeBRNN has comparable $\epsilon_{r}$ to the RNN, achieving marginally better mean predictions when $T_c > 20$. The remaining metrics are only relevant to the VeBRNN, as they quantify the quality of uncertainty estimates and their separation (the RNN is deterministic). The Epistemic TLL (\Cref{fig:dns_sve_c}) improves with more training paths, indicating that the VeBRNN reduces model uncertainty with more training data (as expected). The Wasserstein distance (\Cref{fig:dns_sve_b}) to the aleatoric uncertainty distribution also reduces, dropping sharply from $T_c=5$ to $T_c=40$ and then plateauing, indicating that VeBRNN rapidly and effectively captures data noise (aleatoric uncertainty) and separates it from the epistemic uncertainty. Based on the well-captured aleatoric uncertainty, we see that the PICP achieves the desired test coverage of $0.95$ for the ID testing data -- this indicates that 95\% of testing points are within the estimated confidence interval by the VeBRNN. Evidently, in the extrapolation regime, i.e., for OOD testing data, the PICP is worse. Equivalently, the tightening of the epistemic uncertainty confidence interval with increasing data is clear from \Cref{fig:dns_sve_e} (decreasing MPIW value) -- this indicates that the model becomes more confident about its prediction with increasing training data. For the OOD test paths, the performance generally worsens for all performance metrics, as expected. Nevertheless, we observe that MPIW is larger than that for ID test paths, reflecting that the VeBRNN predicts larger epistemic uncertainty for the extrapolation regime (as it should). Interestingly, the $\epsilon_{r}$ of deterministic training is slightly smaller than for the VeBRNN for OOD test paths. Bayesian models have a tendency to revert to the non-trivial prior in the OOD region, as there is no information to create a good quality prediction when there is no data.

\Cref{fig:psgld_id_test_path} exemplifies the predictions of VeBRNN for a randomly selected ID testing path (for all 3 components), and considering different training dataset sizes. The quality of uncertainty quantification and disentanglement achieved by the cooperative training of VeBRNNs is evident. The last row of the figure, corresponding to results obtained for all $2000$ training paths, shows that the predicted aleatoric distribution (shaded areas) closely overlaps with the ground truth distribution (black dotted lines) generated through 100 repeated simulations based on $\widetilde{\text{DNS}}$. Note the nontrivial nature of these predictions, since the model captures the larger variation of $\sigma_{11}$ and $\sigma_{22}$ compared to $\sigma_{12}$.  \Cref{fig:psgld_id_test_path_f} presents the predicted epistemic uncertainty: the band is narrow and yet it covers all three stress components (recall that this path is unseen). The epistemic uncertainty is smaller where the input strain is smaller and increases when the input strain magnitude grows, as expected. For comparison, the prediction from deterministic RNNs is given \Cref{fig:sf_mse_training}, where it can be seen that $\sigma_{12}$ exhibits significant overfitting, emphasizing the advantages of Bayesian modeling in reliable uncertainty quantification and avoiding overconfident predictions.

From top to bottom, \Cref{fig:psgld_id_test_path} indicates the influence of different training set sizes on the model's predictive performance. The results reinforce that fewer training points decrease performance for all accuracy metrics, as also shown in \Cref{fig:sve_dns_results_comparison}. Concerning the uncertainty estimates, few training points lead to less confident predictions for both uncertainties, with wider shaded areas in both aleatoric and epistemic uncertainties. It is worth mentioning that the aleatoric uncertainty predictions for all stress components remain stable from $T_c = 40$ (\Cref{fig:psgld_id_test_path_c} to $T_c = 100$ (\Cref{fig:psgld_id_test_path_e}) while the predicted epistemic uncertainties (\Cref{fig:psgld_id_test_path_d} and \ref{fig:psgld_id_test_path_f} ) decrease, as expected.

\subsection{Results for Dataset 2: $\overline{\text{ROM}}$+$\overline{\text{DNS}}$ (MF problem with deterministic LF \& HF data)}
\label{sec:mf_problem_1}

This MF dataset does not have aleatoric uncertainty (noise); therefore, we can consider RNNs for both fidelities. However, we can also consider VeBRNNs (which reduce to BRNNs because they find zero aleatoric uncertainty), in case we want to predict epistemic uncertainty. We include both cases in this section. Unless otherwise stated, all the available LF training paths indicated in \Cref{tab:mf_datasets_info} are used, and models are assessed for an increasing number of HF training paths.

\begin{figure}[h!]
    \centering
    \begin{subfigure}[t]{0.354\textwidth}
     \includegraphics[width=\textwidth]{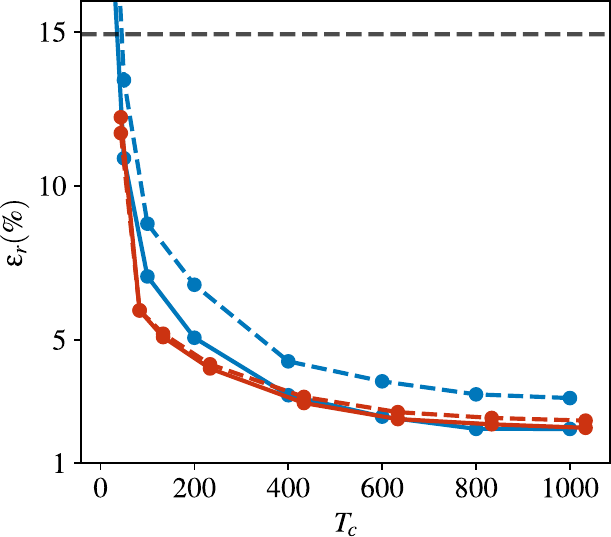}   
     \caption{$N_\text{cluster=3}$ }
     \label{fig:mf_problem_1_deter_comparison_a}
    \end{subfigure}
    \hspace{0.03\textwidth}
    \begin{subfigure}[t]{0.33\textwidth}
     \includegraphics[width=\textwidth]{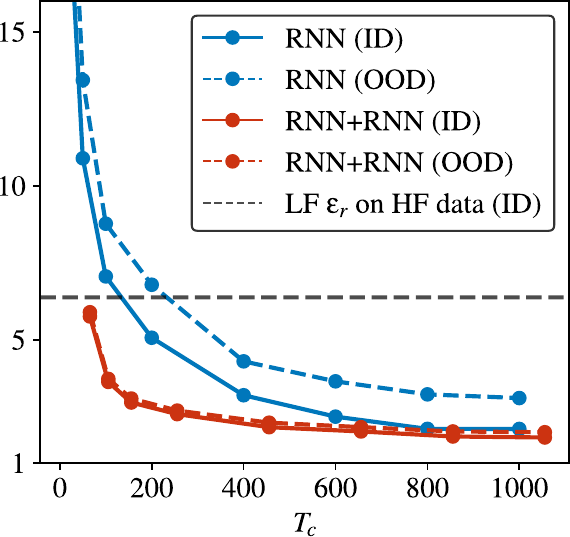}  
     \caption{$N_\text{cluster=18}$}
     \label{fig:mf_problem_1_deter_comparison_b}
    \end{subfigure}
    \caption{Results obtained for the fully deterministic MF model (RNN+RNN) trained on Dataset 2 ($\overline{\text{ROM}}$+$\overline{\text{DNS}}$) compared with a single-fidelity RNN trained on data obtained from $\overline{\text{DNS}}$. (a) results of LF data with $N_\text{cluster}=3$; (b) results of LF data  with $N_\text{cluster}=18$. For the single-fidelity RNN, $T_c$ corresponds to the following number of training paths: $[10, 50, 100, 200, 400, 600, 800, 1000]$. For RNN+RNN model, $T_c$ has offsets of $33.3$ and $55.5$ for $N_\text{cluster}=3$ and $N_\text{cluster}=18$, respectively, because we use all the LF data and only change the HF data size. The dark gray dashed lines on both figures indicate the relative error $\epsilon_r$ of LF models on the HF data.}
    \label{fig:mf_problem_1_deter_comparison}
\end{figure}

\paragraph{MF model with RNN+RNN architectures.}
\label{para:deterministic_training_mf_dataset_1}
A MF model with fully deterministic architectures is trained using \emph{Adam} \cite{Kingma2014}. \Cref{fig:mf_problem_1_deter_comparison} presents the results, including a comparison with a single-fidelity RNN model trained only on HF data ($\overline{\text{DNS}}$). Up to a certain computational cost, the MF model (RNN+RNN) is more accurate than the single-fidelity model, especially for OOD test paths. The MF model advantage is more pronounced when the LF model is better (using a higher number of clusters for the SCA method -- \Cref{fig:mf_problem_1_deter_comparison_b}). For a limit case of only 10 HF training paths, the RNN+RNN model achieves significantly lower $\epsilon_r$ than the single RNN (note that the relative error $\epsilon_r$ is even smaller than the one obtained by the LF model -- shown by the horizontal dark gray bar). These findings confirm that the RNN+RNN model is particularly effective in data-scarce HF scenarios -- encouraging for future real-world Engineering applications. Importantly, the RNN+RNN model significantly decreases the $\epsilon_r$ gap between ID and OOD test paths -- a key benefit of leveraging computationally inexpensive LF data to explore a larger domain and enhancing HF prediction accuracy and generalization.

\begin{figure}[h!]
        \centering
    \hspace*{0.05\textwidth} 
    \begin{subfigure}[t]{0.33\textwidth}
        \centering
        \includegraphics[width=\textwidth]{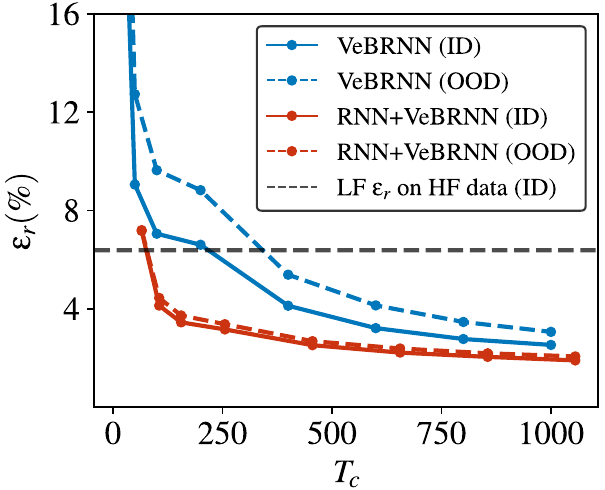}
        \caption{Mean estimate metric ($\downarrow$)}
        \label{fig:mf_problem_1_bayes_comparison_a}
    \end{subfigure}
    \begin{subfigure}[t]{0.33\textwidth}
        \centering
        \includegraphics[width=\textwidth]{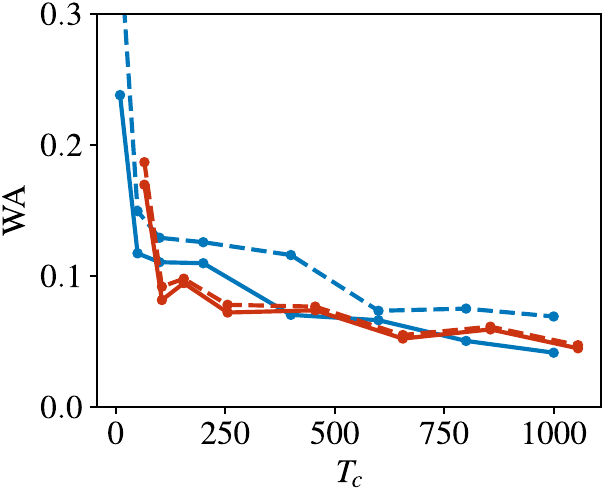}
        \caption{Aleatoric uncertainty metric ($\downarrow$)}
        \label{fig:mf_problem_1_bayes_comparison_b}
    \end{subfigure}
    \vspace{0.3cm}
    \begin{subfigure}[t]{0.33\textwidth}
        \centering
        \includegraphics[width=\textwidth]{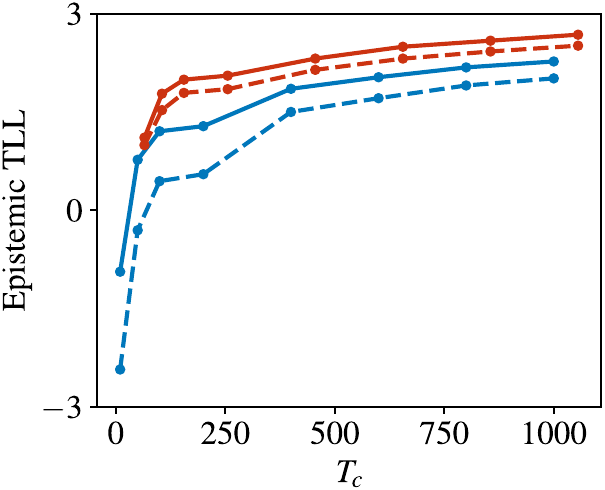}
        \caption{Epistemic uncertainty metric ($\uparrow$)}
        \label{fig:mf_problem_1_bayes_comparison_c}
    \end{subfigure}
    \begin{subfigure}[t]{0.33\textwidth}
        \centering
        \includegraphics[width=\textwidth]{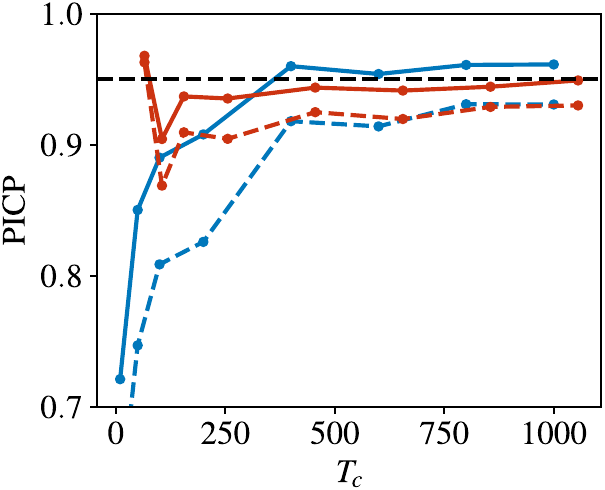}
        \caption{Epistemic uncert. metric ($\rightarrow 0.95$)}
        \label{fig:mf_problem_1_bayes_comparison_d}
    \end{subfigure}
    \begin{subfigure}[t]{0.33\textwidth}
        \centering
        \includegraphics[width=\textwidth]{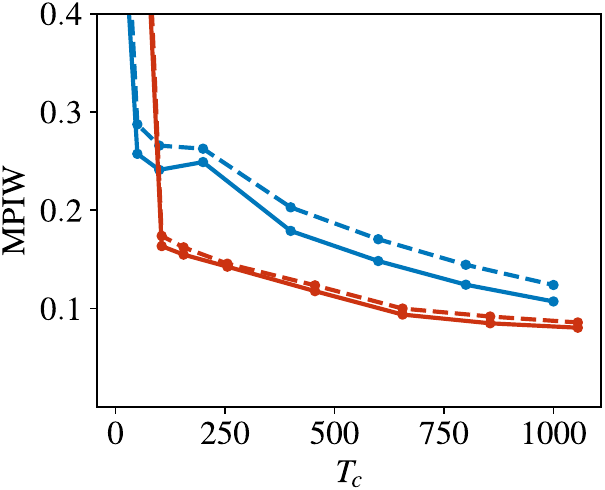}
         \caption{Epistemic uncertainty metric ($\downarrow$)}
        \label{fig:mf_problem_1_bayes_comparison_e}
    \end{subfigure}
    \caption{Results of RNN+VeBRNN model trained on Dataset 2 ($\overline{\text{ROM}}$+$\overline{\text{DNS}}$), compared to single fidelity VeBRNN trained only on HF data ($\overline{\text{DNS}}$). The LF data was obtained by SCA of an RVE using $N_\text{cluster}=18$. The horizontal dashed lines in (d) indicate the desired 95\% confidence interval.}
    \label{fig:mf_problem_1_bayes_comparison}

\end{figure}

\begin{figure}[h!]
    \centering
    \begin{subfigure}[t]{0.371\textwidth}
        \centering
        \includegraphics[width=\textwidth]{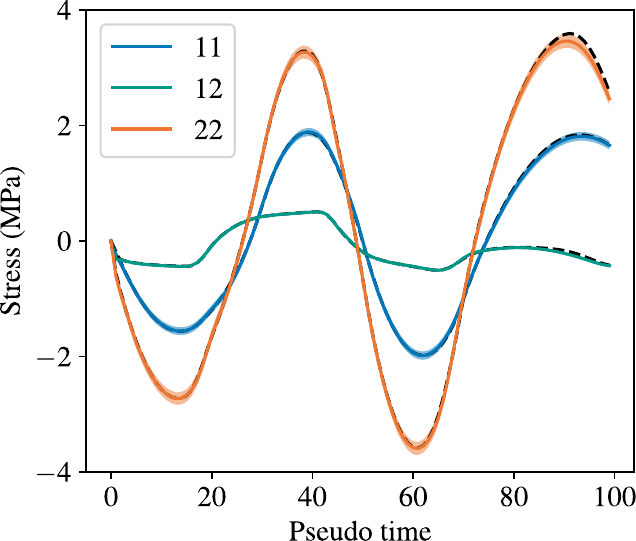}
        \caption{RNN+VeBRNN Aleatoric uncertainty }
        \label{fig:prediction_of_mf_redisual_on_id_path_a}
    \end{subfigure}
    \hspace{0.03\textwidth}
    \begin{subfigure}[t]{0.35\textwidth}
        \centering
        \includegraphics[width=\textwidth]{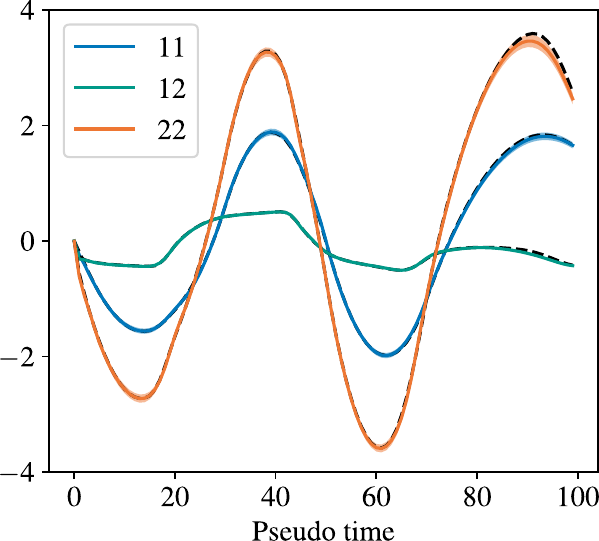}
        \caption{RNN+VeBRNN Epistemic uncertainty}
        \label{fig:prediction_of_mf_redisual_on_id_path_b}
    \end{subfigure}
    \caption{Prediction of RNN+VeBRNN for a training dataset ($\overline{\text{ROM}}$+$\overline{\text{DNS}}$) with $T_c = 1000 + 2000\times\frac{1}{36} =1055.5$, where the LF model was obtained by SCA of an RVE using $N_\text{cluster}=18$. (a) predicted aleatoric distributions (ground-truth is Dirac delta, i.e., zero aleatoric uncertainty), (b) predicted epistemic distributions.}
    \label{fig:prediction_of_mf_redisual_on_id_path}
\end{figure}

\paragraph{MF model with RNN+VeBRNN architectures.} Considering the VeBRNN (which will end up reducing to a BRNN) for the HF data allows to quantify epistemic uncertainty (and to estimate zero aleatoric uncertainty). Results are shown in \Cref{fig:mf_problem_1_bayes_comparison}. Note that now we compare with a single-fidelity model with the VeBRNN, instead of the RNN, so that we can assess if there is an advantage in predicting epistemic uncertainty by using MF data. The relative error evolution with increasing training dataset sizes is consistent with the previous observations for the other models, with a clear advantage for the MF model compared to the SF one. As in \Cref{sec:DNS-SVE dataset experiments}, the PICP and MPIW reveal that the RNN+VeBRNN model estimates a wide epistemic confidence interval when using only 10 HF training paths, corresponding to a high PICP. As the number of HF training paths increases, the MPIW drops rapidly, initially accompanied by a decrease in PICP, followed by a steady recovery to the target confidence level of $95\%$. Regarding the Wasserstein distance, it is calculated with a ground truth of zero variance (as there is no noise), and it can be seen to reduce to a small value rapidly, demonstrating that the RNN+VeBRNN finds that the dataset has negligible noise.\footnote{If the analyst already knows that the dataset is noiseless for both fidelities, then the VeBRNN can simply be replaced by a BRNN (as there is no need to find the noise). We wanted to demonstrate that the method is general, and can be used even when one is not certain about the presence or absence of aleatoric uncertainty.}

\Cref{fig:prediction_of_mf_redisual_on_id_path} presents the RNN+VeBRNN prediction for an ID test path for a model trained with a total computational cost of $T_c = 1000 +  2000 \times \frac{1}{36}=1055.5$. The LF model was obtained by SCA of an RVE using $N_\text{cluster}=18$.  Since both fidelity levels are obtained from RVE simulations, the dataset is noiseless. As a result, the predicted aleatoric uncertainty of the RNN+VeBRNN is near zero in \Cref{fig:prediction_of_mf_redisual_on_id_path_a}. Moreover, the predicted epistemic uncertainty is very tight but still covers the ground truth, as it should. The results demonstrate the advantage of considering a Bayesian model, even for noiseless datasets. They also show that MF models can effectively learn from datasets gathered using fewer computational resources.

\subsection{Results for Dataset 3: $\widetilde{\text{ROM}}$+$\widetilde{\text{DNS}}$ (MF problem with stochastic LF \& HF data)}
\label{sec:mf_problem_2}

\begin{figure}[h!]
    \centering

        \hspace*{0.04\textwidth} 
    \begin{subfigure}[t]{0.33\textwidth}
        \centering
        \includegraphics[width=\textwidth]{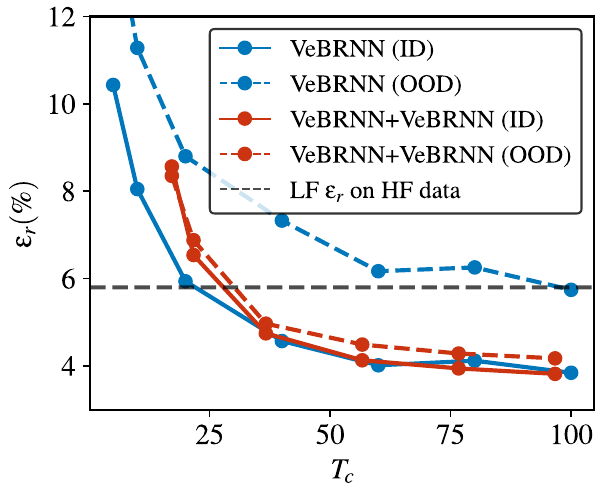}
        \caption{Mean estimate metric ($\downarrow$)}
        \label{fig:mf_uq_dns_sve_dns_sve_a}
    \end{subfigure}
    \begin{subfigure}[t]{0.33\textwidth}
        \centering
        \includegraphics[width=\textwidth]{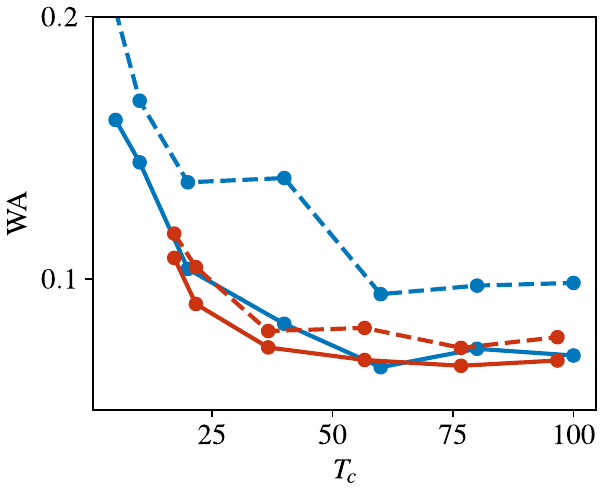}
        \caption{Aleatoric uncertainty metric ($\downarrow$)}
        \label{fig:mf_uq_dns_sve_dns_sve_b}
    \end{subfigure}
    \vspace{0.3cm}
    \begin{subfigure}[t]{0.33\textwidth}
        \centering
        \includegraphics[width=\textwidth]{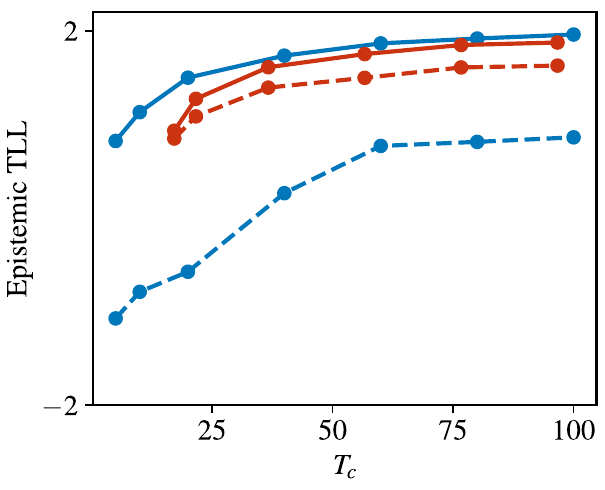}
        \caption{Epistemic uncertainty metric ($\uparrow$)}
        \label{fig:mf_uq_dns_sve_dns_sve_c}
    \end{subfigure}
    \begin{subfigure}[t]{0.33\textwidth}
        \centering
        \includegraphics[width=\textwidth]{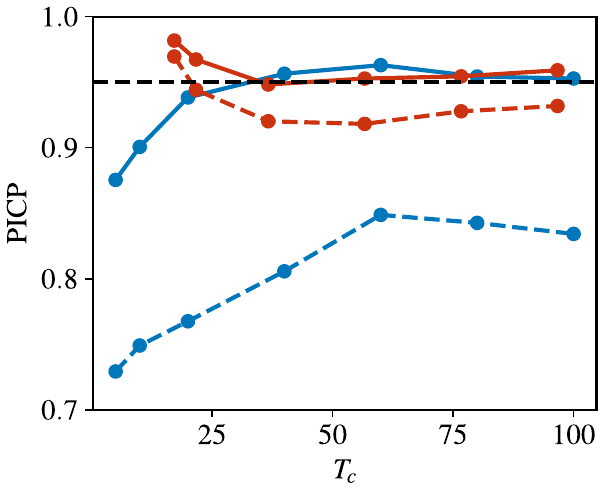}
        \caption{Epistemic uncert. metric ($\rightarrow 0.95$)}
        \label{fig:mf_uq_dns_sve_dns_sve_d}
    \end{subfigure}
    \begin{subfigure}[t]{0.33\textwidth}
        \centering
        \includegraphics[width=\textwidth]{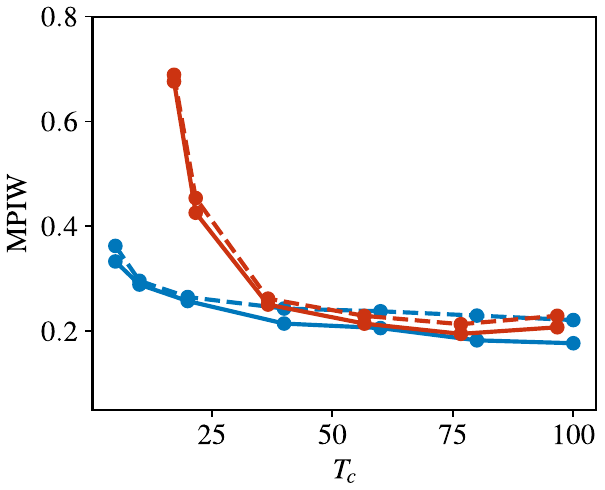}
        \caption{Epistemic uncertainty metric ($\downarrow$)}
        \label{fig:mf_uq_dns_sve_dns_sve_e}
    \end{subfigure}
    \caption{Results for the VeBRNN+VeBRNN model, when compared with a single-fidelity VeBRNN trained on HF data. The LF data is obtained by SCA of an SVE with $N_\text{cluster}=18$. The number of  HF training paths ($\widetilde{\text{DNS}}$) used in MF training is $[10, 100, 400, 800, 1200, 1600]$, and the number of  LF paths is fixed at 2000. This corresponds to an initial total computational cost of $T_c = \frac{10}{20} + \frac{2000}{120} = 17.16$ for the MF dataset.}
    \label{fig:mf_uq_dns_sve_dns_sve}
\end{figure}

This is the most challenging dataset, as both fidelities are stochastic (noisy). \Cref{fig:mf_uq_dns_sve_dns_sve} presents the VeBRNN+VeBRNN model when trained with the LF model using $N_\text{cluster}=18$. We also share the results for an RNN+RNN model (fully deterministic) in \Cref{sec:deter_train_mf_problem_2}, where the LF model uses $N_\text{cluster}=3$ for the SCA method. Furthermore, we also considered a MF model using RNN+VeBRNN, but the predictions are similar to the VeBRNN+VeBRNN model (see \Cref{sec:deter_train_mf_problem_2} for the RNN+VeBRNN model results). 

\Cref{fig:mf_uq_dns_sve_dns_sve} shows that the MF model has similar performance to the single-fidelity model trained on HF data when considering the same total computational cost for ID test paths. This result is not surprising when noticing that the cost ratio between HF and LF data acquisition is only 6:1, but the LF is significantly less accurate and both fidelities are noisy. However, for OOD test paths the MF model is significantly better than the single-fidelity model, as observed by comparing the dashed lines in \Cref{fig:mf_uq_dns_sve_dns_sve}.  

\begin{remark}
    \label{rmk:dataset3}
    In principle, it is not possible to guarantee \textit{a priori} if a given MF dataset will lead to a MF model that performs better than the corresponding single-fidelity model for the same total budget. However, using active learning, it is possible to automate the data acquisition process, allowing to choose between fidelities on-the-fly. Bayesian models like VeBRNNs are a crucial development towards this because active learning becomes possible with accurate epistemic uncertainty estimation. Therefore, we find active learning and Bayesian optimization as an interesting future direction to pursue.
\end{remark}

\subsection{Results for Dataset 4: $\widetilde{\text{ROM}}$+$\overline{\text{DNS}}$ (MF problem with stochastic LF \& deterministic HF)}
\label{sec:mf_problem_3}

\begin{figure}[h!]
    \centering

    \begin{subfigure}[t]{0.354\textwidth}
     \includegraphics[width=\textwidth]{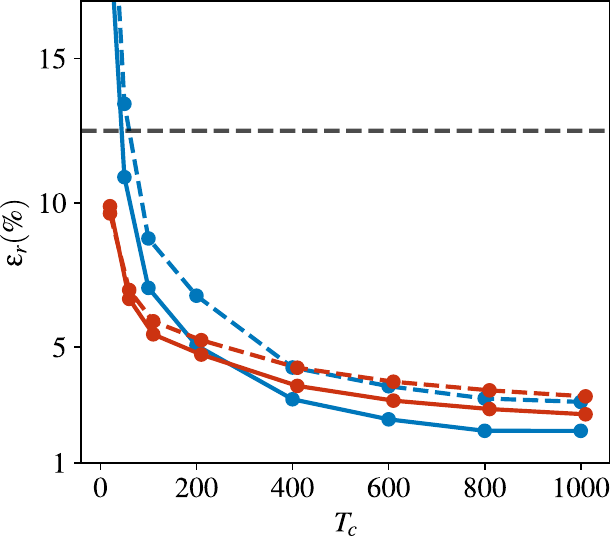}   
     \caption{$N_\text{cluster}=3$ }
     \label{fig:mf_dns_rve_sca_sve_a}
    \end{subfigure}
    \hspace{0.03\textwidth}
    \begin{subfigure}[t]{0.33\textwidth}
     \includegraphics[width=\textwidth]{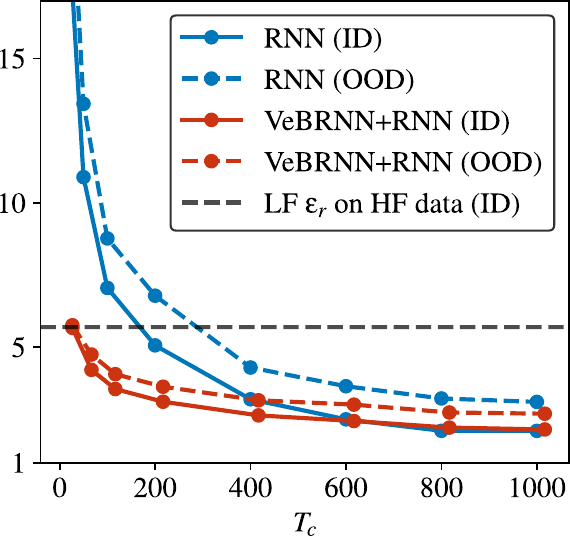}  
     \caption{$N_\text{cluster}=18$ }
     \label{fig:mf_dns_rve_sca_sve_b}
    \end{subfigure}
    \caption{Results for VeBRNN+RNN model trained on Dataset 4 ($\widetilde{\text{ROM}}$+$\overline{\text{DNS}}$) compared to single-fidelity RNN trained on HF data. (a) LF model trained on data obtained by SCA of an SVE with $N_\text{cluster}=3$; (b). LF model trained on data obtained by SCA of an SVE with $N_\text{cluster}=18$.}
    \label{fig:mf_dns_rve_sca_sve}
\end{figure}

\begin{figure}[h!]
    \centering

    \begin{subfigure}[t]{0.371\textwidth}
        \centering
        \includegraphics[width=\textwidth]{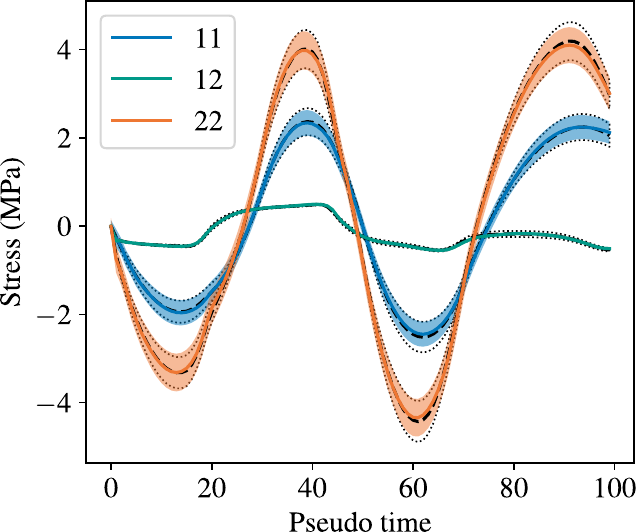}
        \caption{VeBRNN+RNN Aleatoric uncertainty}
    \end{subfigure}
    \hspace{0.03\textwidth}
    \begin{subfigure}[t]{0.35\textwidth}
        \centering
        \includegraphics[width=\textwidth]{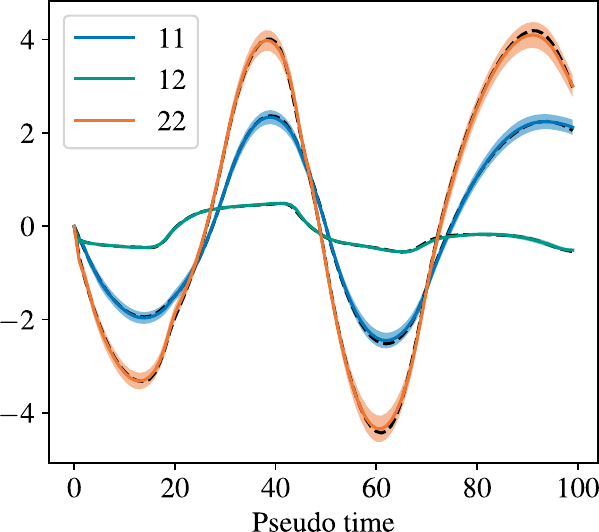}
        \caption{VeBRNN+RNN Epistemic uncertainty}
    \end{subfigure}
    \caption{Predictions of the LF model (VeBRNN) for training data obtained by SCA of an RVE with $N_\text{cluster}=18$ and considering all $2000$ training paths. The ground truth is obtained by repeating the SVE simulation 100 times. (a) LF predicted aleatoric uncertainty; (b) LF predicted epistemic uncertainty}
    \label{fig:lf_pretrained_bayes_prediction}
\end{figure}

\Cref{fig:mf_dns_rve_sca_sve} presents the results obtained for Dataset 4 using a VeBRNN+RNN model (Bayesian LF model and deterministic HF model). The results include a comparison with the single-fidelity RNN trained only on HF data. The results are similar to \Cref{fig:mf_problem_1_deter_comparison}, where a clear advantage is observed when the LF model has a higher accuracy (obtained with the ROM with $N_\text{cluster}=18$, instead of $N_\text{cluster}=3$). Note that in this example the VeBRNN is used to train the LF model (because the LF data is noisy, and we want to determine the aleatoric uncertainty -- see \Cref{tab:model_outputs}). With all the $2000$ training paths used to train the LF model, the VeBRNN achieves $\epsilon_r=3.87\%$, $\text{Epistemic TLL}=1.78$, $\text{WA} = 0.074$, $\text{PICP}=0.949$, and $\text{MPIW}=0.214$. The predictions for a test path are shown in \Cref{fig:lf_pretrained_bayes_prediction}, where excellent predictions for all mean, aleatoric, and epistemic uncertainties are observed. In this case, the estimation of aleatoric uncertainty also reflects the quality of LF data.

\begin{figure}[h!]
    \centering

    \begin{subfigure}[t]{0.33\textwidth}
        \centering
        \includegraphics[width=\textwidth]{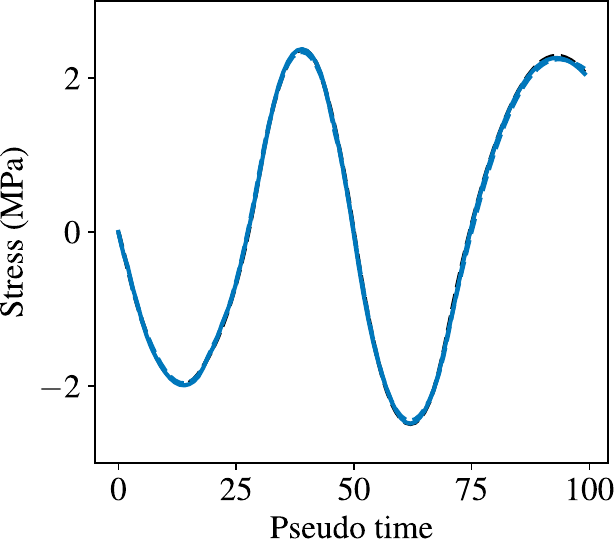}
        \caption{$\sigma_{11}$ }
    \end{subfigure}
    \hspace{0.01\textwidth}
    \begin{subfigure}[t]{0.321\textwidth}
        \centering
        \includegraphics[width=\textwidth]{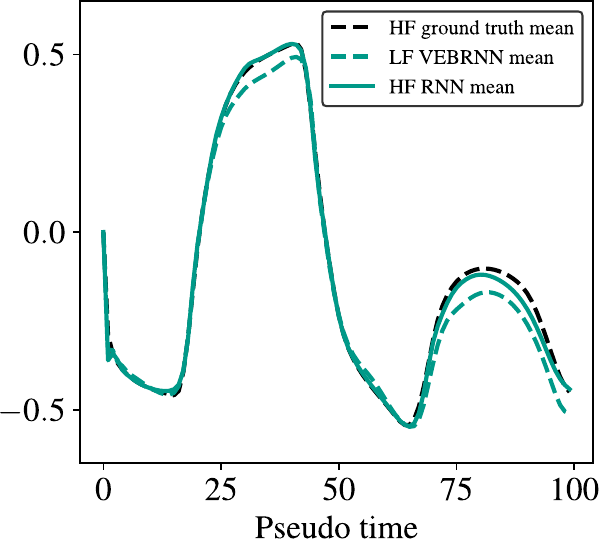}
        \caption{ $\sigma_{12}$ }
    \end{subfigure}
    \hspace{0.01\textwidth}
    \begin{subfigure}[t]{0.308\textwidth}
        \centering
        \includegraphics[width=\textwidth]{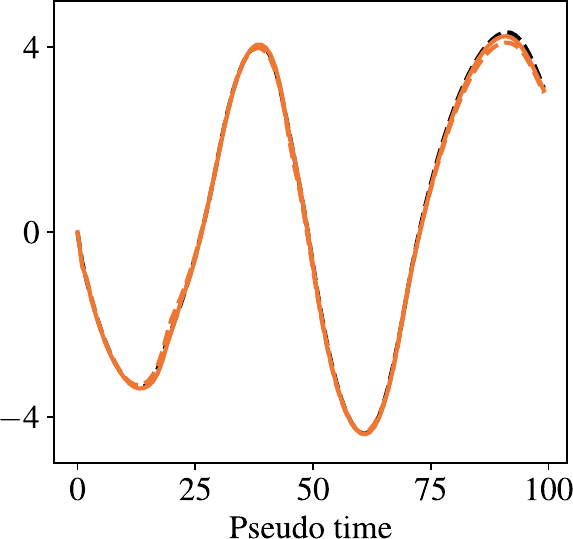}
        \caption{ $\sigma_{22}$ }
    \end{subfigure}
    \caption{Prediction of VeBRNN+RNN model for Dataset 4 ($\widetilde{\text{ROM}}$+$\overline{\text{DNS}}$) on a random HF test path using $2000$ LF training paths (obtained by SCA of an SVE with $N_\text{cluster}=18$) and using $1000$ HF training paths (obtained by FEA of an RVE). The black dashed line is the HF ground truth (from FEA of an RVE), the colored dashed line represents the LF mean prediction on the HF data, and the solid colored line is the final mean prediction by the MF model.}
    \label{fig:lf_hf_prediction}
\end{figure}

\Cref{fig:lf_hf_prediction} shows the HF prediction from the MF model (VeBRNN+RNN). We note that the LF predictions that were transferred to the HF model, shown as colored dashed lines, already provide accurate estimates for $\sigma_{11}$ and $\sigma_{22}$, but not for $\sigma_{12}$. The MF model effectively narrows this gap and predicts the HF accurately (colored solid lines). We find it interesting that a Bayesian model on the LF improves a deterministic model at the HF, although this might not be a common model construction (typically, the analyst intends to predict epistemic uncertainty at the HF and may not be interested in estimating noise at the LF). In addition, there is a potential risk of overfitting to the HF data, particularly in the prediction of $\sigma_{12}$. As discussed in earlier sections, considering a Bayesian model at the HF mitigates this risk.

\section{Discussion}
\label{sec:discussion}

There are three key characteristics that control whether MF models are more accurate than their single-fidelity counterparts: (1) the cost ratio between acquiring HF and LF data; (2) how much error (bias) is introduced by the LF data; and (3) how much aleatoric uncertainty (noise) is present in the LF or HF data. This can be demonstrated by considering Datasets 3 and 4, evaluating performance for different computational budgets, and using different HF and LF data ratios.

\begin{figure}[h!]
    \centering

    \hspace*{0.05\textwidth} 
    \begin{subfigure}[t]{0.33\textwidth}
        \centering
        \includegraphics[width=\textwidth]{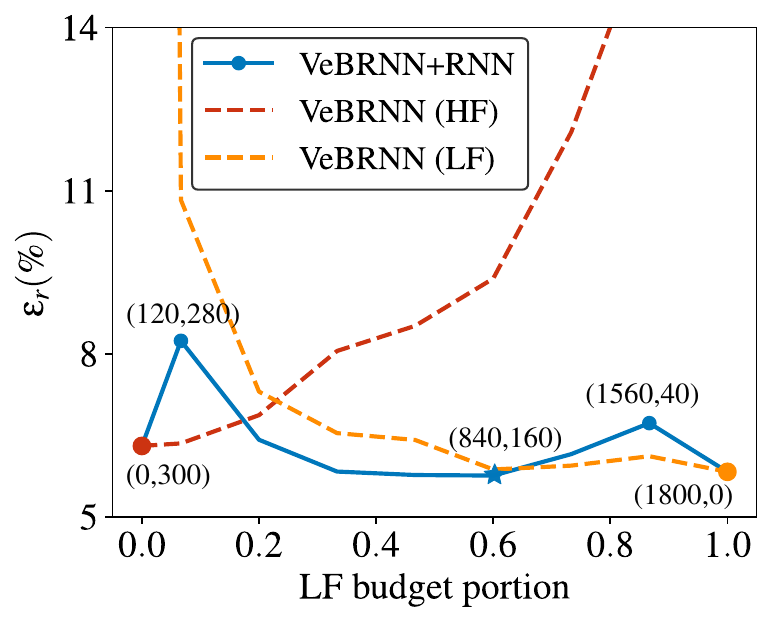}
         \caption{Mean estimate metric ($\downarrow$)}
        \label{fig:computational_allocation_dataset3_a}
    \end{subfigure}
    \hspace*{0.005\textwidth} 
    \begin{subfigure}[t]{0.323\textwidth}
        \centering
        \includegraphics[width=\textwidth]{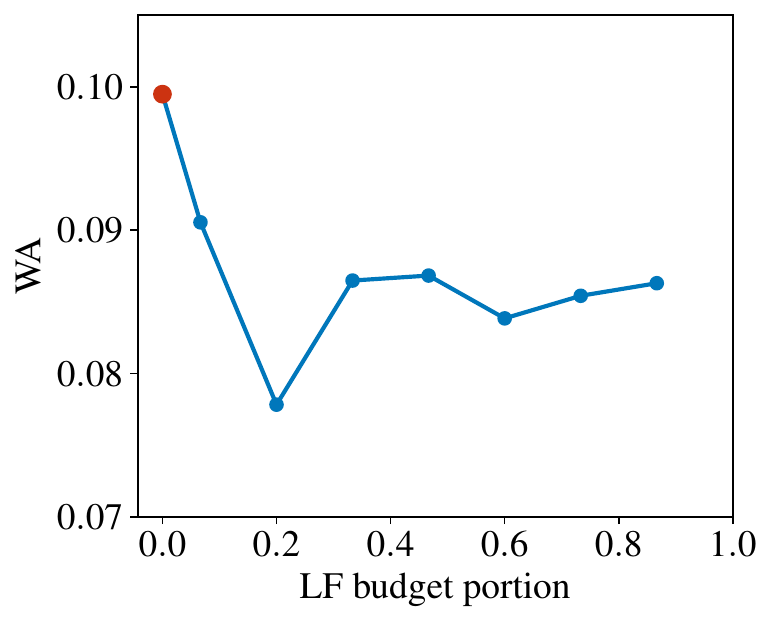}
        \caption{Aleatoric uncertainty metric ($\downarrow$)}
        \label{fig:computational_allocation_dataset3_b}
    \end{subfigure}
    \vspace{0.3cm}
    \begin{subfigure}[t]{0.33\textwidth}
        \centering
        \includegraphics[width=\textwidth]{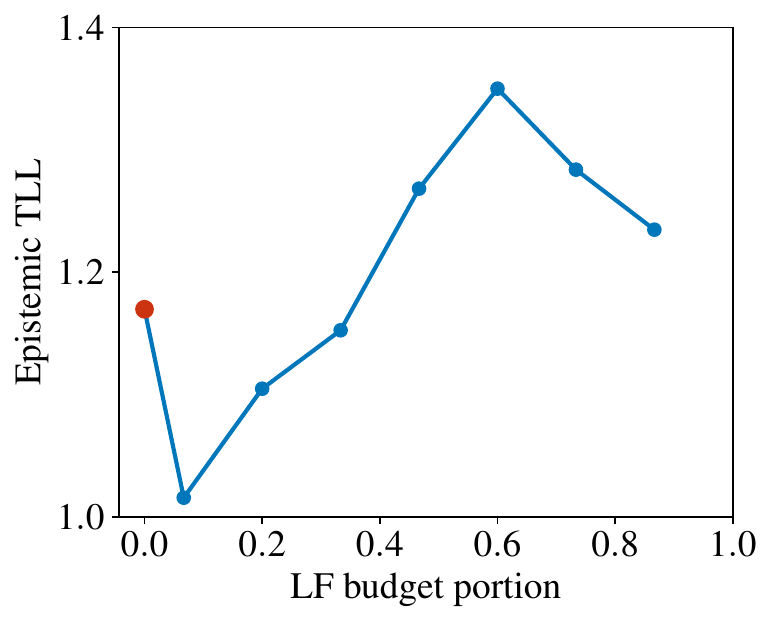}
        \caption{Epistemic uncertainty metric ($\uparrow$)}
        \label{fig:computational_allocation_dataset3_c}
    \end{subfigure}
    \begin{subfigure}[t]{0.33\textwidth}
        \centering
        \includegraphics[width=\textwidth]{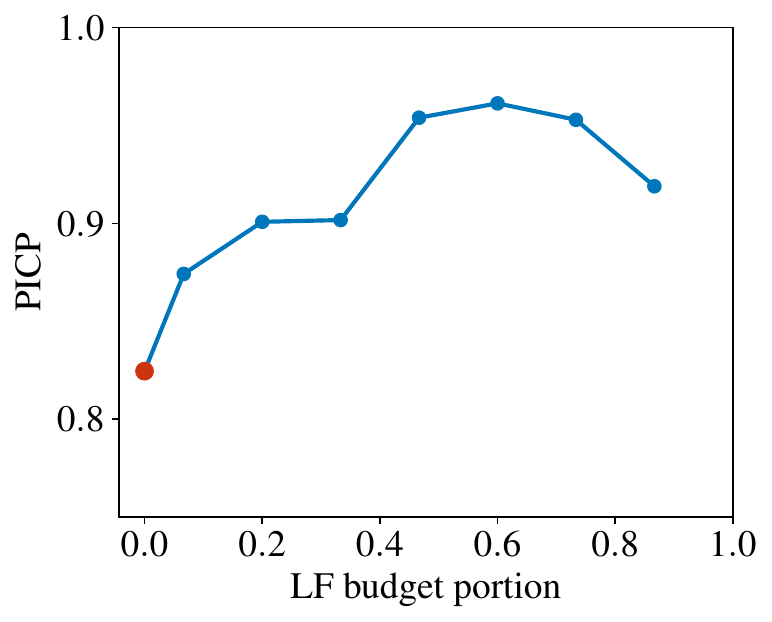}
        \caption{Epistemic uncert. metric ($\rightarrow 0.95$)}
        \label{fig:computational_allocation_dataset3_d}
    \end{subfigure}
    \begin{subfigure}[t]{0.33\textwidth}
        \centering
    \includegraphics[width=\textwidth]{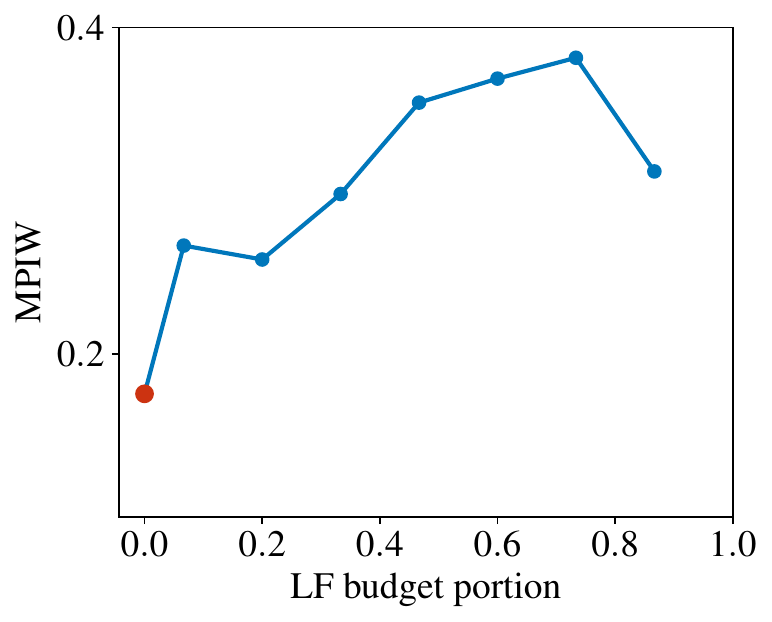}
        \caption{Epistemic uncertainty metric ($\downarrow$)}
        \label{fig:computational_allocation_dataset3_e}
    \end{subfigure}
    \caption{Results for Dataset 3 ($\widetilde{\text{ROM}}$+$\widetilde{\text{DNS}}$ with $N_\text{cluster}=18$) when considering different budget percentages of LF data acquisition for a total budget cost of $T_c=15$. The performance metrics of the VeBRNN+VeBRNN model are compared with the two single-fidelity models obtained when allocating the entire budget to  HF data (red marker) or LF data (orange marker). The red and orange dashed lines in (a) represent the relative errors of using only the HF or LF data from the MF model train setup counterpart shown in the parentheses. Note that we do not include the orange marker for metrics characterizing uncertainty (b--e), as there is no HF prediction in that case.}
    \label{fig:computational_allocation_dataset3}
\end{figure}

Starting with Dataset 3 ($\widetilde{\text{ROM}}$+$\widetilde{\text{DNS}}$), recall that the MF model's performance was not significantly different from a model trained only on HF data for a given total cost when using 2000 training paths for the LF data and testing with ID paths (\Cref{sec:mf_problem_2}). However, \Cref{fig:computational_allocation_dataset3} shows that if we change the percentage of LF data used in training while keeping the total cost at $T_c=15$, then the MF model outperforms both the HF and LF models for most cases (blue markers compared to red and orange markers, respectively). However, these improvements are modest because (1) the accuracy of the HF and LF models is similar, and (2) the cost ratio between the acquisition of the HF and LF data only differs by a factor of 6 (see \Cref{tab:mf_datasets_info}). Therefore, if the HF and LF data have similar acquisition cost, and similar accuracy gain for each additional total cost made available, then the MF model loses usefulness.

\begin{figure}[hbt!]
    \centering

    \begin{subfigure}[t]{0.32\textwidth}
        \centering
        \includegraphics[width=\textwidth]{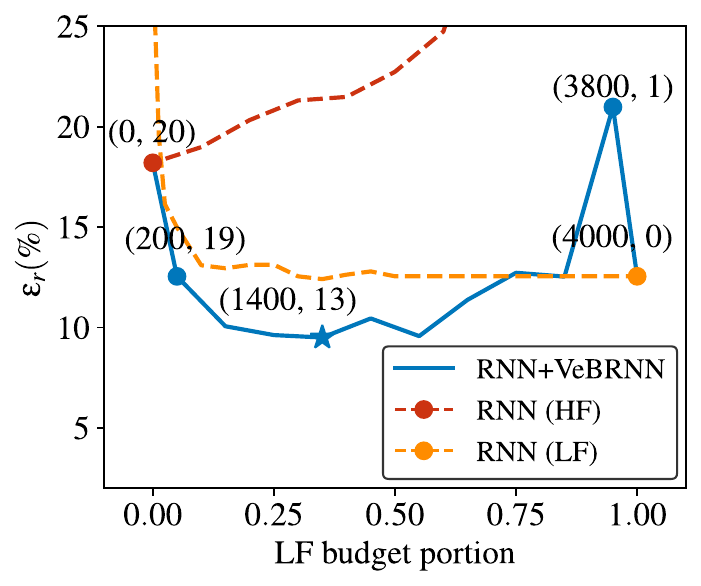}
        \caption{$T_c=20$}
    \end{subfigure}
    \hspace{0.01\textwidth}
    \begin{subfigure}[t]{0.3\textwidth}
        \centering
        \includegraphics[width=\textwidth]{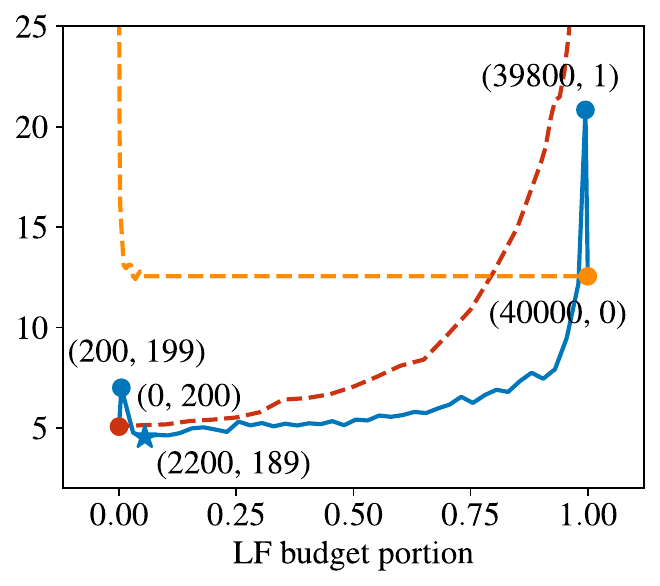}
        \caption{$T_c=200$ }
    \end{subfigure}
    \hspace{0.01\textwidth}
    \begin{subfigure}[t]{0.3\textwidth}
        \centering
        \includegraphics[width=\textwidth]{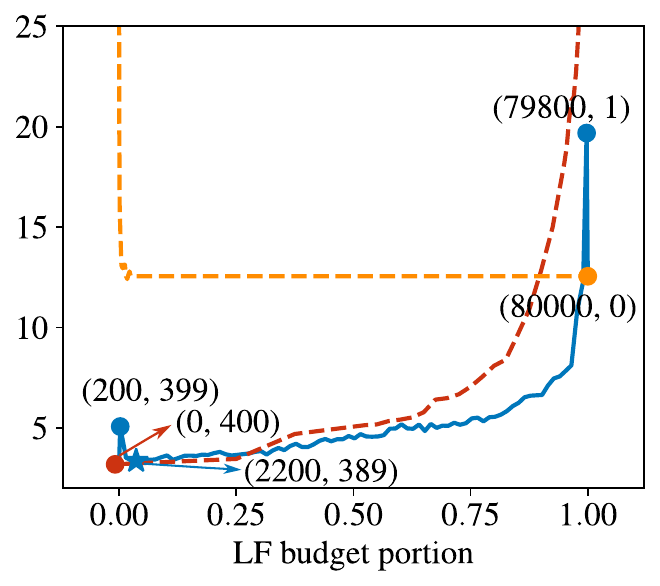}
        \caption{$T_c=400$ }
    \end{subfigure}
    \caption{Results for Dataset 4 ($\widetilde{\text{ROM}}$+$\overline{\text{DNS}}$ with $N_\text{cluster}=18$) when considering different budget percentages of LF data acquisition for a given total budget cost: (a) $T_c=20$, (b) $T_c=200$, and (c) $T_c=400$. Each point is labeled with the number of training paths used at the LF and HF, respectively. The red and orange dashed lines represent the relative errors of using only the HF or LF data from the MF model train setup counterpart shown in the parentheses. Confront these results with \Cref{fig:mf_dns_rve_sca_sve_b}, obtained using 2000 LF training paths and a different number of HF training paths leading to the corresponding total cost in that figure.}
    \label{fig:changing_budget_Dataset4}
\end{figure}

A similar observation arises from considering Dataset 4 ($\widetilde{\text{ROM}}$+$\overline{\text{DNS}}$). \Cref{sec:mf_problem_3} showed that if the LF data is obtained by a reasonably accurate ROM (SCA method discretized by $N_\text{cluster}=18$ clusters), then the MF model is better for a large range of total cost values. This is no longer the case if the ROM is less accurate (SCA discretized by $N_\text{cluster}=3$ clusters). \Cref{fig:changing_budget_Dataset4} complements this observation by showing model error as a function of the percentage of LF budget used for a given total cost. In this figure, we see that for low total cost ($T_c=20$) there is a wide range where the MF model is better than the single-fidelity counterparts (the best MF model is highlighted by a star marker, and curiously it is close to the case considering 2000 training paths for the LF in \Cref{sec:mf_problem_3}). Unsurprisingly, if the total budget is high ($T_c=200$ or $T_c=400$), then there is already enough budget to effectively learn by only using HF data. This explains why the best MF model for $T_c=200$ is only marginally better than the HF model, and why none of the MF models is better than the HF model for $T_c=400$.

We also note that in all cases (\Cref{fig:computational_allocation_dataset3} and \Cref{fig:changing_budget_Dataset4}) we see that if we use very few samples for HF data (or conversely for LF data), then the MF model is worse than the corresponding single-fidelity model (notice the spikes close to the extremes of the plots). This is logical because if there are very few samples of a given fidelity, then the model obtained for that fidelity has a large error (dashed lines). In that case, the information transferred from that fidelity within the MF model is not sufficiently useful (in fact, it disrupts model performance). Interestingly, we see this behavior quickly dissipating, i.e., even a small number of points (but not extremely small) becomes beneficial. Our investigations point to the heuristic conclusion that LF data acquisition should be allocated between 30\% and 70\% of the total budget, although this depends on the three key factors mentioned above.

\section{Conclusions and future work}
\label{sec:conclusion}

We demonstrate that variance estimation Bayesian recurrent neural networks (VeBRNNs) can be used within a multi-fidelity (MF) framework to learn nonlinear, history-dependent phenomena while quantifying and disentangling uncertainties. Our contribution is modular, since the most general MF model that uses VeBRNNs in each fidelity can be simplified (or ablated) to the simplest single-fidelity model that uses a deterministic feedforward neural network model (\Cref{fig:training_flowchart} explains how to reduce a MF model with an arbitrary number of fidelities to a single-fidelity model; \Cref{fig:architecture_flowchart} shows how to progressively choose more complex neural network models up to VeBRNNs).

Furthermore, we highlight that VeBRNNs are trained by a cooperative strategy that also simplifies into standard deterministic training of a neural network when only performing Step 1 of \Cref{alg:CUQalgorithm}. If aleatoric uncertainty does not need to be estimated, we can skip Step 2 of \Cref{alg:CUQalgorithm}. If epistemic uncertainty is also not needed, we can skip Step 3 of \Cref{alg:CUQalgorithm}. Therefore, the presented methodology seamlessly generalizes conventional data-driven learning from the simplest deterministic cases to the most complete Bayesian cases presented to date. We expect this work will open new avenues in data-driven design and analysis, far beyond constitutive modeling.

Without loss of generality, the examples in this article focused on data-driven learning of constitutive models from data generated by simulations of material volume elements with or without aleatoric uncertainty (data noise). Four different scenarios were presented: one single-fidelity case with noisy data, and three bi-fidelity cases considering noisy and noiseless data for different fidelities. The proposed method can quantify and separate epistemic and aleatoric uncertainty at every fidelity (when using VeBRNNs). In other words, the models simultaneously estimate the quality of the mean response (epistemic uncertainty) and the inherent noise present in the data (aleatoric uncertainty).

The proposed method is envisioned to be applied across a wide range of Science and Engineering problems. However, we believe that further generalizing this work to include active learning and applying it to Bayesian optimization is important. This will enable on-the-fly data acquisition, selecting the fidelity from which the next data point should be acquired (for learning or for optimization).

\bibliographystyle{unsrt}
\bibliography{reference}

\newpage
\appendix
\section{Bayesian inference methods} \label{sec:Bayesian_inference_methods}

We have mentioned that MCMC and VI are two fundamental approaches for approximating the PPD (\Cref{eq:Bayes rule}). In the following subsections, we briefly introduce those approaches in \Cref{subsec:MCMC} and \Cref{subsec:VI}, respectively.

\subsection{Markov Chain Monte Carlo sampling} \label{subsec:MCMC}

MCMC is a class of algorithms that can be used to sample from a probability distribution directly \cite{murphy2022probabilistic_advanced}. The most straightforward approach is the random walk Metropolis-Hasting algorithm \cite{Neal2012}, while it suffers from high rejection rates for high-dimensional problems. To address this limitation and accelerate the mixing rate of MCMC, the Hamiltonian Monte Carlo (HMC) was developed by Neal et al. \cite{Neal2012}. HMC leverages the concept of Hamiltonian mechanics by incorporating the  \textbf{gradient} of the target probability density to guide the sampling trajectory. Therefore, the mixing rate can be improved tremendously compared with the random walk Metropolis-Hasting algorithm.

\subsubsection{Hamiltonian Monte Carlo} \label{subsubsec:HMC}

\begin{figure}[h]
    \centering
    \includegraphics{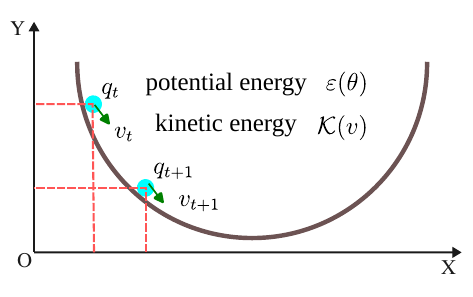}
    \caption{Schematic of Hamiltonian mechanics system}
    \label{fig:schematic_of_hmc}
\end{figure}

The schematic of the Hamiltonian mechanics system is shown in \Cref{fig:schematic_of_hmc}, where the total energy is:
\begin{equation}
    \mathcal{H}\left(\bm{\theta}, \bm{v} \right) =
    \mathcal{\varepsilon(\bm{\theta})} + \mathcal{K}(\bm{v})
\end{equation}
Generally, the potential energy with parameter $\bm{\theta}$ is usually set to be the QoI, specifically, the negative log PPD shown in \Cref{eq:Bayes rule}:

\begin{equation}
    \mathcal{\varepsilon(\bm{\theta})} = - \log \, \tilde{p}(\bm{\theta})
\end{equation}

and the  kinetic energy to be:

\begin{equation}
    \mathcal{K}(\bm{v}) = \frac{1}{2}\bm{v}^{T} {\Sigma}^{-1} \bm{v}
\end{equation}

where ${\Sigma}$ is the mass matrix and $\bm{v}$ is the velocity.

As illustrated in \Cref{fig:schematic_of_hmc},  the bowl-shaped surface represents the target PDF landscape, where the blue dot points are the expected sample points. The HMC can be interpreted as going over the whole landscape in light of the position updating rule of the Hamiltonian mechanics system. To further improve the accuracy of the sampling process, the leapfrog technique is adopted \cite{Neal2012}, which is a symplectic integration operator that can preserve the energy of the system. Overall, the algorithm is summarized in Algorithm \ref{alg:hmc}.

\begin{algorithm}[h]
    \SetAlgoLined
    \KwData{target $p(\bm{\theta})$,  proposed $q(\bm{\theta}^\prime |\bm{\theta}) \sim \mathcal{N}(\bm{\theta}^\prime  |\bm{\theta}, \eta^{2} \bm{I})$, burn-in iteration $N_{b}$, total iteration $N_{t}$,
        step size $\eta$, leapfrog steps $L$ }
    \KwResult{Collection of $\bm{\Theta} =\{\bm{\theta}^{1},\bm{\theta}^{2}, ... \bm{\theta}^{N_{a}} \}$}
    initialize $\bm{\theta}^{0}$ \\
    \For{$t=0,1,2,...,T$}{
    Sample random momentum $\bm{v}_{t-1} \sim N(\bm{0}, \bm{\Sigma})$ \\
    Set ($\bm{\theta}_{0}^{'}, \bm{v}_{0}^{'}) =(\bm{\theta}_{t-1}, \bm{v}_{t-1}) $   \\
    Half step for momentum: $\bm{v}_{\frac{1}{2}}^{'} =\bm{v}_{t}^{'}-\frac{\eta }{2}\nabla \mathcal{\varepsilon}(\bm{\theta}_0^{'})$ \\
    \For{$l=1:L-1$}{
    $\bm{\theta}_{l}^{'} =  \bm{\theta}_{l-1}^{'}+ \eta \bm{\Sigma}^{-1}\bm{v}_{l-\frac{1}{2}}^{'} $ \\
    $\bm{v}_{l+\frac{1}{2}}^{'} =\bm{v}_{l-\frac{1}{2}}^{'}-\eta \nabla \mathcal{\varepsilon}(\bm{\theta}_l^{'}) $ \\
    }
    Full step for location: $\bm{\theta}_{L}^{'} =  \bm{\theta}_{L-\frac{1}{2}}^{'}+ \eta \bm{\Sigma}^{-1}\bm{v}_{L-\frac{1}{2}}^{'} $ \\
    Half step for momentum:$\bm{v}_{L}^{'} =\bm{v}_{L-\frac{1}{2}}^{'}-\frac{\eta }{2}\nabla \mathcal{\varepsilon}(\bm{\theta}_L^{'})$ \\
    Compute $\alpha =min(1,\exp\left[-\mathcal{H}(\bm{\theta}_{L}, \bm{v}_{L}) + \mathcal{H}(\bm{\theta}_{t-1}, \bm{v}_{t-1}) \right]) $ \\
    Set $\bm{\theta}_t = \bm{\theta}_L$ with probability $\alpha$, otherwise  $\bm{\theta}_t = \bm{\theta}_{t-1}$

    }
    \caption{Hamiltonian Monte Carlo Algorithm}
    \label{alg:hmc}
\end{algorithm}

As shown in \Cref{alg:hmc}, HMC still involves an acceptance–rejection step.
Although it significantly improves over the random-walk Metropolis-Hastings algorithm, this step can still hinder its efficiency in high-dimensional problems, such as obtaining the posterior distribution of BNNs in \Cref{eq:Bayes rule}.

\subsubsection{Stochastic Gradient Langevin Dynamics}
\label{sec:psgld}

To address the scalability bottleneck of HMC, Welling et al. \cite{Welling2011} developed a novel Bayesian inference approach called Stochastic Gradient Langevin Dynamics (SGLD), leveraging Stochastic Gradient Descent (SGD) \cite{ruder2017overview} and Langevin Monte Carlo (LMC) \cite{murphy2022probabilistic_advanced}. As mentioned in Algorithm \ref{alg:hmc}, the \textit{leapfrog} step is essential to the accuracy of HMC, where a larger $L$ is expected for a higher acceptance rate, while requiring more posterior evaluations. LMC can be regarded as a special case of HMC for $L=1$ whose procedure is listed in Algorithm \ref{alg:lmc}.

\begin{algorithm}[h]
    \SetAlgoLined
    \KwData{target $p(\bm{\theta})$,  proposed $q(\bm{\theta}^\prime |\bm{\theta}) \sim \mathcal{N}(\bm{\theta}^\prime  |\bm{\theta}, \eta^{2} \bm{I})$, burn-in iteration $N_{b}$, total iteration $N_{t}$,
        step size $\eta$  }
    \KwResult{Collection of $\bm{\Theta} =\{\bm{\theta}^{1},\bm{\theta}^{2}, ... \bm{\theta}^{N} \}$}
    initialize $\bm{\theta}^{0}$ \\
    \For{$t=1,2,...,T$}{
        Sample random momentum $\bm{v}_{t-1} \sim N(\bm{0}, \bm{\Sigma})$ \\
        Update the location: $\bm{\theta}^{\prime} =  \bm{\theta}_{t-1}- \frac{\eta^2}{2}\Sigma^{-1}\nabla\bm{\varepsilon} (\bm{\theta}_{t-1}) +  \eta \bm{\Sigma}^{-1}\bm{v}_{t-1} $ \\
        Update the velocity: $\bm{v}^{\prime} =\bm{v}_{t-1}-\frac{\eta }{2}\nabla \mathcal{\varepsilon}(\bm{\theta}_{t-1})- \frac{\eta }{2}\nabla \mathcal{\varepsilon}(\bm{\theta}^{\prime}) $ \\
        Compute $\alpha =min(1,\exp\left[-\mathcal{H}(\bm{\theta}^{\prime}, \bm{v}^{\prime}) + \mathcal{H}(\bm{\theta}_{t-1}, \bm{v}_{t-1}) \right]) $ \\
        Set $\bm{\theta}_t = \bm{\theta}^{\prime}$ with probability $\alpha$, otherwise  $\bm{\theta}_t = \bm{\theta}_{t-1}$

    }
    \caption{Langevin Monte Carlo Algorithm}
    \label{alg:lmc}
\end{algorithm}

The position updating rule of LMC includes an additional term, $\frac{\eta^2}{2}\Sigma^{-1}\nabla\bm{\varepsilon} (\bm{\theta}_{t-1})$, compared with that of \Cref{alg:hmc}. This extra term follows the Langevin diffusion process. Also, there is a simplification to eliminate the MH acceptance step, making the algorithm more efficient in practice:

\begin{align} \label{eq:lmc_update}
    \bm{\theta}_t & = \bm{\theta}_{t-1} - \frac{\eta^2}{2}\bm{\Sigma}^{-1}\nabla\bm{\varepsilon} (\bm{\theta}_{t-1}) +  \eta \bm{\Sigma}^{-1}\bm{v}_{t-1}               \\
                  & = \bm{\theta}_{t-1} - \frac{\eta^2}{2}\bm{\Sigma}^{-1}\nabla\bm{\varepsilon} (\bm{\theta}_{t-1}) +  \eta \bm{\sqrt{\Sigma^{-1}}}\bm{\epsilon}_{t-1}
\end{align}

where $\bm{v}_{t-1} \sim \mathcal{N}(\bm{0}, \bm{\Sigma})$ and $\bm{\epsilon}_{t-1} \sim \mathcal{N}(\bm{0}, \mathbf{I})$. It is noted that the above updating is similar to the SGD by adding noise to it ( $\eta \bm{\sqrt{\Sigma^{-1}}}\bm{\epsilon}_{t-1}$).  When $\bm{\Sigma} = \mathbf{I}$ and $\eta =\sqrt{2}$, the updating process follows the Langevin diffusion written as:

\begin{equation}
    \mathrm{d}\bm{\theta}_t = - \nabla \varepsilon (\bm{\theta}_t) \mathrm{d} t + \sqrt{2}\mathrm{d} \bm{B}_t
\end{equation}

where $ \bm{B}_t$ is a d-dimensional Brownian motion.

Now, let's revisit the SGD with a loss function \cite{ruder2017overview}:
\begin{equation}
    \mathcal{L}(\bm{\theta})  = \sum_{n=1}^{N} \mathcal{L}_n(\bm{\theta})
\end{equation}
By taking a mini-batch, we can reformulate the mini-batch loss function and calculate the corresponding  gradient as
\begin{equation}
    \nabla_B\mathcal{L}(\bm{\theta})  = \frac{1}{B} \sum_{n=1}^{B} \mathcal{L}_n(\bm{\theta})
\end{equation}

There is a difference (diffusion term) between the actual gradient and the mini-batch one, expressed as
\begin{equation}
    \bm{v}_t = \sqrt{\eta} \left( \nabla\mathcal{L}(\bm{\theta})   -  \nabla_B\mathcal{L}(\bm{\theta})  \right)
\end{equation}
To this end, SGD with mini-batch can be rewritten as
\begin{equation} \label{eq:sgd_update}
    \bm{\theta}_{t+1} = \bm{\theta}_t - \eta \nabla_B \mathcal{L} (\bm{\theta}_t) = \bm{\theta}_t - \eta \nabla \mathcal{L} (\bm{\theta}_t) + \sqrt{\eta}\bm{v}_t
\end{equation}
where  $\bm{v}_t$ has a zero mean.

Combining \Cref{eq:lmc_update} and \ref{eq:sgd_update}, the updating process of SGLD is obtained by substituting the
$p(\bm{\theta} | \bm{X}) \propto p(\bm{\theta})\prod_{n=1}^{N} p(y_i | \bm{\theta}) $  to the loss function $\mathcal{L}(\bm{\theta})$ \cite{Welling2011}:
\begin{equation} \label{eq:sgld_update}
    \Delta \theta_t = \frac{\bm{\epsilon}_t}{2} \left(\nabla \log p(\bm{\theta}_t )  +\frac{N}{n} \log p(y_i|\bm{\theta}_t)\right) + \bm{\eta}_t
\end{equation}
where  $\bm{\eta}_t \sim \mathcal{N}(\bm{0}, \bm{\epsilon}_t)$.

However, SGLD assumes that all parameters $\bm{\theta}$ have the same step size, leading to slow convergence or even divergence in the cases where the components of $\bm{\theta}$ have different curvatures. Thus, a refined version of SGLD, called preconditioned SGLD (pSGLD) \cite{Li2015}, was proposed to address this issue. In pSGLD\cite{Li2015}, the update rule incorporates a user-defined preconditioning matrix $G(\bm{\theta}_t)$, which adjusts the gradient updates and the noise term adaptively:

\begin{equation} \label{eq:sgld}
    \Delta \theta_t = \frac{\bm{\epsilon}_t}{2} \left[G(\bm{\theta}_t) \left(\nabla \log p(\bm{\theta}_t )  +\frac{N}{n} \log p(y_i|\bm{\theta}_t)\right)  + \Gamma(\bm{\theta}_t) \right ] + \bm{\eta}_tG(\bm{\theta}_t)
\end{equation}

where  $\Gamma_i = \sum_j \frac{\partial G_{i,j}(\bm{\theta}) }{\partial \theta_j}$ describes how the preconditioner changes with respect to $\bm{\theta}$.

\subsection{Variational Inference (VI)} \label{subsec:VI}

\begin{figure}[h]
    \centering
    \includegraphics[width=0.5\textwidth]{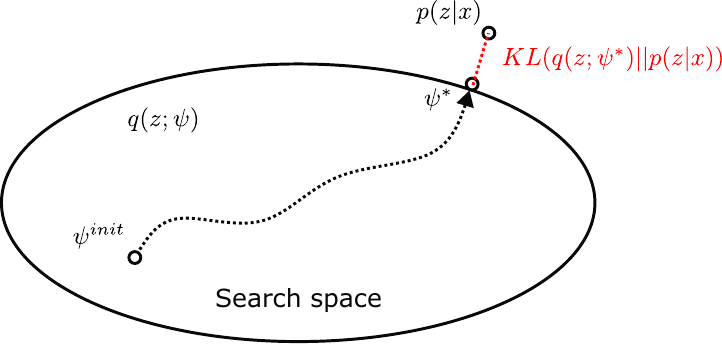}
    \caption{Schematic of VI \cite{murphy2022probabilistic_advanced}: The ellipse represents the search space of the proposed distribution, $p(\bm{\theta}|\mathcal{D})$ is the true posterior distribution, the KL divergence is the distance between the two distributions defined as $\mathrm{D}_\mathbb{KL} \left(q(z)||p_\theta (z|x) \right)$, and the goal is to minimize it.}
    \label{fig:vi}
\end{figure}

VI is another Bayesian inference method that approximates the PPD by a proposed distribution from a parametric family \cite{Blei2017}. The objective of  VI is to minimize the Kullback-Leibler divergence between the true posterior distribution $p(\bm{\theta}|\mathcal{D})$ and the proposed distribution $q(\bm{\theta})$ as illustrated in  \Cref{fig:vi}.

In practice, $\psi$ is regarded as the variational parameter from a parametric family $\Omega$; therefore, the optimal $\psi^*$ can be obtained by minimizing the KL divergence as follows:

\begin{align} \label{eq:vi_inference_equation}
    \psi^* & = \arg \min_{\psi} \mathrm{D}_\mathbb{KL} \left(q_{\psi}(\bm{\theta})||p_\theta(\bm{\theta}|\mathcal{D}) \right)                                                                               \\
           & =  \arg \min_{\psi} \mathbb{E}_{q_\psi(\bm{\theta})} \left[\log q_\psi(\bm{\theta}) - \log p_\theta (\bm{\theta}|\mathcal{D}) \right]                                                          \\
           & =  \arg \min_{\psi} \mathbb{E}_{q_\psi(\bm{\theta})} \left[\log q_\psi(\bm{\theta}) - \log
    \left(\frac{p_\theta(\mathcal{D}|\bm{\theta})p_\theta(\bm{\theta})}{p_\theta(\mathcal{D})} \right) \right]                                                                                              \\
           & =  \arg \min_{\psi} \mathbb{E}_{q_\psi(\bm{\theta})} \left[\log q_\psi(\bm{\theta}) - \log p_\theta(\mathcal{D}|\bm{\theta}) - \log p_\theta(\bm{\theta}) +\log p_\theta(\mathcal{D}) \right]  \\
           & =  \arg \min_{\psi} \mathbb{E}_{q_\psi(\bm{\theta})} \left[\log q_\psi(\bm{\theta}) - \log p_\theta(\mathcal{D}|\bm{\theta}) - \log p_\theta(\bm{\theta})  \right] +\log p_\theta(\mathcal{D})
\end{align}

According to \Cref{eq:vi_inference_equation}, the optimization problem can be decomposed into two terms: the first term is the negative evidence lower bound (ELBO), and the second term is the log marginal evidence. The log marginal evidence is a constant term that does not depend on $\psi$; therefore, we only need to optimize the first term alone:

\begin{align}
    \psi^* & = \arg \min_{\psi} \mathbb{E}_{q_\psi(\bm{\theta})} \left[\log q_\psi(\bm{\theta}) - \log p_\theta(\mathcal{D}|\bm{\theta}) - \log p_\theta(\bm{\theta})  \right]                      \\
           & = \arg \min_{\psi} \mathbb{E}_{q_\psi(\bm{\theta})} \left[ - \log p_\theta(\mathcal{D}|\bm{\theta})\right] + \mathrm{D}_{\mathbb{KL}}(q_{\psi}(\bm{\theta})|| p_\theta(\bm{\theta})) \
\end{align}

where the first term is the negative log likelihood of the given dataset, and the second term is the KL divergence between the proposed distribution and the prior distribution. The optimization problem can be solved by any optimization algorithm, such as SGD \cite{ruder2017overview} or Adam \cite{Kingma2014}.

\newpage
\section{Performance metrics}
\label{sec:performance_metrics}

To evaluate the performance of different methods, we adopt the following metrics evaluated based on the test dataset:

\begin{itemize}
    \item \textbf{Relative error ($\epsilon_r$)}
          \begin{equation} \label{eq:error}
              \epsilon_r = \frac{1}{N_\text{test}T} \sum_{i=1}^{N_\text{test}}  \sum_{t=1}^{T}\Biggl( \frac{\norm{\bar{\mathbf{y}}_{i,t} - \hat{\mathbf{y}}_{i,t}}_F}{\norm{\bar{\mathbf{y}}_{i,t}}_F} \Biggl ) \cdot 100\%,
          \end{equation}
          where $N_\text{test}$ is the number of HF testing points, $\norm{\cdot}_F$ is a Frobenius norm,  $\Bar{\mathbf{y}}_{i,t}$ and $\hat{\mathbf{y}}_{i,t}$ are the mean values of predicted and ground truth of $i^{th}$ test sample and $t^{th}$ time step, correspondingly.

    \item \textbf{Test Log-Likelihood (TLL)}
          \begin{equation} \label{eq:TLL}
              \text{TLL} = \frac{1}{N_\text{test}T} \sum_{i=1}^{N_\text{test}} \sum_{t=1}^{T} \log p(\bm{\Bar{\mathbf{y}}_{i,t}}| \bm{\theta})
          \end{equation}

    \item \textbf{Wasserstein distance (WA)}
          \begin{equation} \label{eq:general wassertein distance}
              \text{WA} =  \frac{1}{N_\text{test}T} \sum_{i=1}^{N_\text{test}} \sum_{t=1}^{T}  \text{W}_2 \left(p(\bar{\mathbf{y}}_{i,t}| \bm{\theta}), q(\bar{\mathbf{y}}_{i,t} | \bm{\theta})\right)
          \end{equation}
          where $p(\bar{\mathbf{y}}_{i,t} | \bm{\theta})$ and $q(\bar{\mathbf{y}}_{i,t} | \bm{\theta})$ are the predicted and ground truth distributions, respectively. If Gaussian assumptions are applied, we can rewrite \Cref{eq:general wassertein distance} into:
          \begin{equation}
              \text{WA} =  \frac{1}{N_\text{test}T} \sum_{i=1}^{N_\text{test}} \sum_{t=1}^{T}  \sqrt{(\bar{\mathbf{y}}_{i,t} - \hat{\mathbf{y}}_{i,t})^2  + (\bar{\mathbf{s}}_{i,t} - \hat{\mathbf{s}}_{i,t})^2 }
          \end{equation}
    \item \textbf{Prediction Interval Coverage Probability (PICP)}
          \begin{equation}
              \text{PICP} = \frac{1}{N_\text{test}T} \sum_{i=1}^{N_\text{test}} \sum_{t=1}^{T} \mathbb{I} \left[ \mathbf{y}_{i,t} \in \left[ \hat{\mathbf{y}}_{i,t}^{\text{lower}}, \hat{\mathbf{y}}_{i,t}^{\text{upper}} \right] \right]
          \end{equation}

    \item \textbf{Mean Prediction Interval Width (MPIW)}
          \begin{equation}
              \text{MPIW} = \frac{1}{N_\text{test}T} \sum_{i=1}^{N_\text{test}} \sum_{t=1}^{T} \left( \hat{\mathbf{y}}_{i,t}^{\text{upper}} - \hat{\mathbf{y}}_{i,t}^{\text{lower}} \right)
          \end{equation}
          where $\hat{\mathbf{y}}_{i,t}^{\text{lower}}$ and $\hat{\mathbf{y}}_{i,t}^{\text{upper}}$ are the predicted lower and upper bounds based on the epistemic uncertainty with a confidence level of 0.95.
\end{itemize}

\newpage
\section{Additional information about the datasets}
\label{sec:addi_datasets}

\subsection{Random polynomial strain-stress paths}
\label{sec:strain_stress_paths}

\begin{figure}[h]
    \centering
    \includegraphics[width=0.5\linewidth]{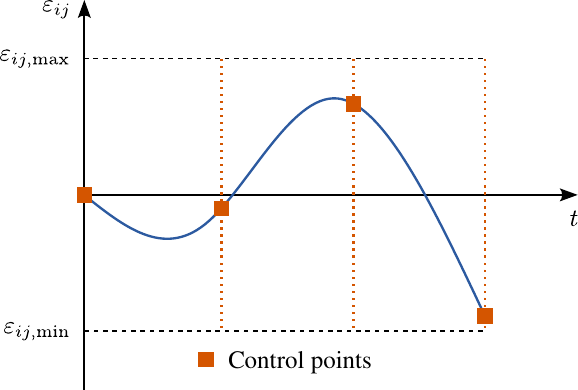}
    \caption{Generation of the random polynomial strain paths.}
    \label{fig:strain_stress_path_generation}
\end{figure}

We adopt the random polynomial strain path developed in our previous work \cite{Mozaffarplasticity2019, Dekhovich2023}, whose generation process is briefly given by \Cref{fig:strain_stress_path_generation}.

As illustrated in \Cref{fig:strain_stress_path_generation}, the control points are randomly sampled within the range $[\varepsilon_{ij, \text{min}}, \varepsilon_{ij, \text{max}}]$ for each strain component $\varepsilon_{ij}$, where ${i,j} = 1,2$. A quadratic curve is then fitted through these points to construct the strain path for each component (blue curve). This continuous path is subsequently discretized into $T$ load steps. In this work, we use 6 control points and set the total number of load steps to $T = 100$. Given the prescribed strain paths, the corresponding history-dependent stress responses can be computed using any solver, where an example is shown in \Cref{fig:data_description}.

\subsection{Convergence analysis to determine the size of an RVE}
\label{sec:convergence_analysis}
\begin{figure}[h]
    \centering
    \includegraphics[width=0.5
        \textwidth]{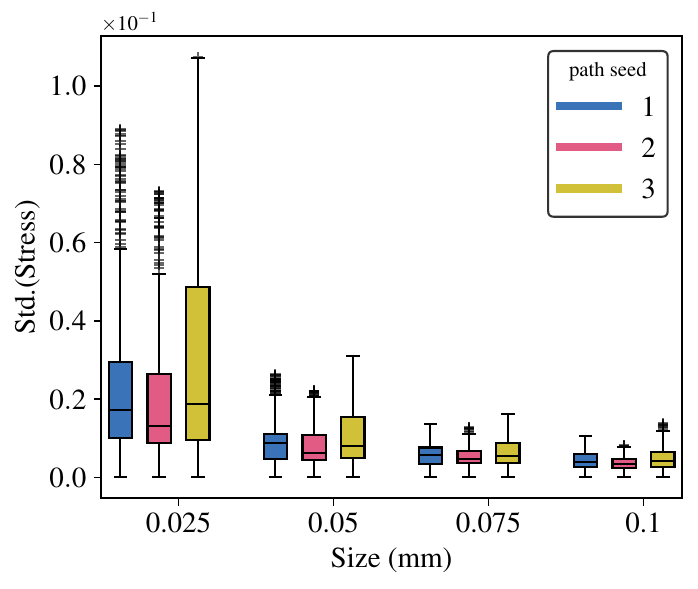}
    \caption{Convergence analysis of the size of the microstructure volume element and the origin of aleatoric uncertainty (noise). Different colors represent different seeds for strain path generation. The size of the volume element varies from $\SI{0.025}{mm}$ to $\SI{0.10}{mm}$, and we repeat the simulation 100 times to get the statistics.}
    \label{fig:influence_of_geometry}
\end{figure}

By definition, a Representative Volume Element (RVE) should have negligible aleatoric uncertainty when randomizing the microstructure. As shown in \Cref{fig:influence_of_geometry}, when reducing the size of the material volume element (\Cref{fig:rve_vs_sve}), the response exhibits aleatoric uncertainty (noise), i.e., we have a Stochastic Volume Element (SVE) instead of an RVE. 

We conduct a convergence analysis based on finite element analyses of different volume element sizes. The results are summarized in \Cref{fig:influence_of_geometry}, where we can see that a volume element with a length of $\SI{0.1}{mm}$ can be considered an RVE. Furthermore, we can also see from \Cref{fig:influence_of_geometry} that the stress variation decreases as the size of the microstructure unit cell increases, highlighting the fact that aleatoric uncertainty originated because of inadequate statistical information. In other words, the aleatoric uncertainty will disappear when RVE is utilized; then it becomes a deterministic dataset as listed in \Cref{sec:datasets}.

\subsection{Aleatoric uncertainty}
\label{sec:represent_aleatoric_unc}

\subsubsection{Note on the aleatoric uncertainty of $\widetilde{\text{DNS}}$ (data obtained from DNS of an RVE) }
As discussed in the main text, RVEs lead to deterministic responses because they contain sufficient morphological and topological information to produce consistent homogenized stress responses \cite{hill1963elastic}. However, aleatoric uncertainty or noise arises from randomizing the microstructure of SVEs because, by definition, they are a small domain of material that is not representative, leading to variations in the homogenized stress responses. Nevertheless, the mean response across multiple SVEs converges to that of the corresponding RVE. These variations correspond to aleatoric uncertainty in our dataset.

Specifically, the particles -- occupying a $30\%$ volume fraction -- are modeled as circles with radii sampled from a normal distribution $\mathcal{N}(0.003, 0.001^2)$. Based on a convergence analysis presented in \Cref{sec:convergence_analysis}, we set the RVE size to $\SI{0.1}{mm}$ with a realization shown in \Cref{fig:rve_vs_sve}. Finite Element Analysis (FEA) is adopted as direct numerical simulation (DNS), where the RVE is discretized into $400 \times 400$ elements. The homogenized stress response undergoing a prescribed random polynomial strain path (\Cref{fig:strain_inputs}) is depicted as the solid lines in \Cref{fig:stress_outputs}.

On the other hand, the SVE is defined with a domain size of $\SI{0.025}{mm}$ as illustrated by the red window in \Cref{fig:rve_vs_sve}. By moving this window within the RVE domain, multiple SVE realizations can be obtained, although we ensure that the microstructure remains periodic. In this case, the local volume fraction varies depending on the window’s position; thus, we model the SVE volume fraction using a normal distribution $\mathcal{N}(0.3, 0.03^2)$ to reflect this spatial variability. In \Cref{fig:stress_outputs}, we observe noticeable variability in the stress components $\sigma_{11}$ and $\sigma_{22}$, while the variation in shear stress $\sigma_{12}$ is significantly smaller. Additionally, the stochastic SVE responses exhibit heteroscedastic aleatoric uncertainty (noise is different for different input values), both across different stress components and along the strain load steps.

\subsubsection{Note on the aleatoric uncertainty of $\widetilde{\text{ROM}}$ (data obtained from SCA of an SVE)}
\label{sec:aleatoric_uncertainty_sca}

\begin{figure}[h]
    \centering
     \begin{subfigure}[t]{0.33\textwidth}
        \centering
        \includegraphics[width=\textwidth]{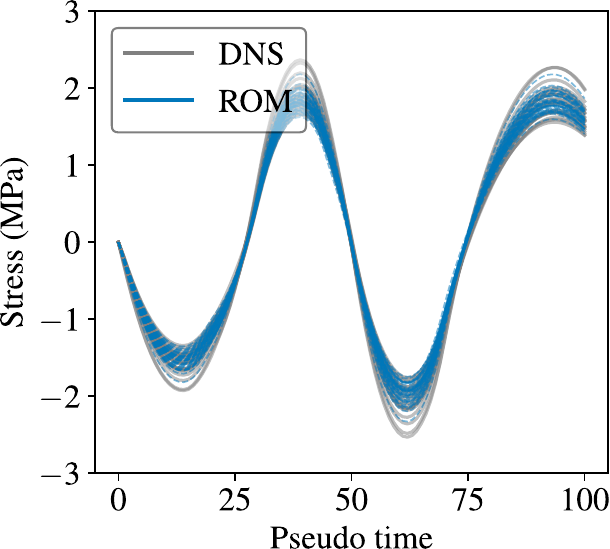}
        \caption{$\sigma_{11}$ }
    \end{subfigure}
    \hspace{0.01\textwidth}
    \begin{subfigure}[t]{0.308\textwidth}
        \centering
        \includegraphics[width=\textwidth]{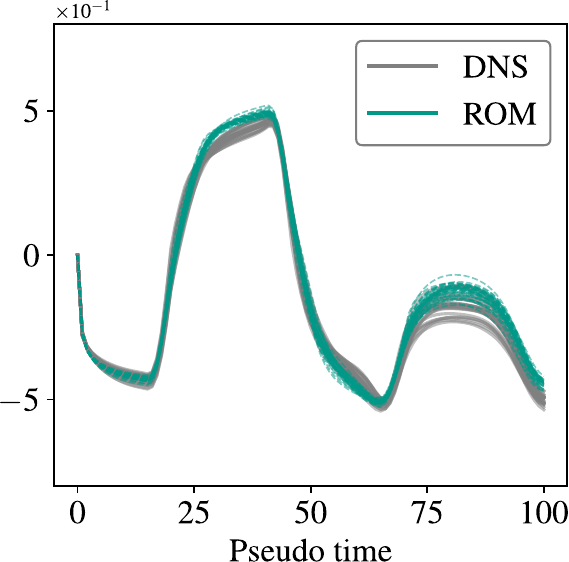}
        \caption{ $\sigma_{12}$ }
    \end{subfigure}
    \hspace{0.01\textwidth}
    \begin{subfigure}[t]{0.308\textwidth}
        \centering
        \includegraphics[width=\textwidth]{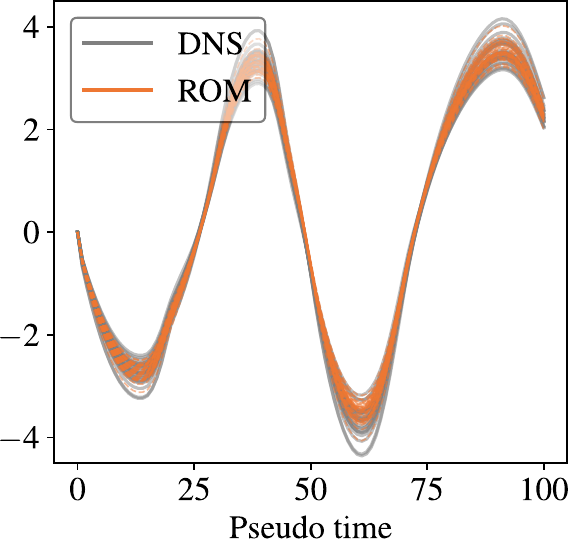}
        \caption{ $\sigma_{22}$ }
    \end{subfigure}
    \caption{Aleatoric uncertainty of \textbf{SCA-SVE}  with $N_\text{cluster}=18$ in comparison to \textbf{DNS-SVE}. From left to right we see the stress components $\sigma_{11}, \sigma_{12}, \sigma_{22}$. Colored lines indicate responses obtained by DNS (FEA), and gray lines by ROM (SCA).}
    \label{fig:aleatoric_uncertainty_sca}
\end{figure}

\Cref{fig:aleatoric_uncertainty_sca} compares the noisy results obtained from $\widetilde{\text{DNS}}$ (i.e., FEA of an SVE) with the noisy results obtained from $\widetilde{\text{ROM}}$ (i.e., SCA of an SVE). This allows for visualizing the differences in aleatoric uncertainty when data is obtained from different simulation methods (FEA vs. SCA). For consistency, note that the stress responses shown in \Cref{fig:aleatoric_uncertainty_sca} are obtained for the same strain path shown in \Cref{fig:rve_vs_sve}.

\Cref{fig:aleatoric_uncertainty_sca} shows that the $\widetilde{\text{ROM}}$ has higher aleatoric uncertainty than the $\widetilde{\text{DNS}}$. This is reasonable because the aleatoric uncertainty in the ROM arises not only from the microstructure variations (the primary source), but also from the discretization level of the ROM, i.e., the number of clusters used in the SCA method (secondary source of aleatoric uncertainty) \cite{ferreira2021adaptive, Liu2016_SCA}. However, in this study, we focus on the total aleatoric uncertainty and do not further decompose its individual contributions.

\subsection{Bias error introduced by $\overline{\text{ROM}}$ compared to $\overline{\text{DNS}}$ (data acquired by SCA of an RVE vs DNS of an RVE)}
\label{sec:irreducible_bias}

Conducting simulations with a fast reduced order model (ROM) when compared to a slow direct numerical simulation is a dominant factor, as illustrated in \Cref{fig:multi_fidelity_concept}. This acceleration is achieved by the self-consistent clustering analysis (SCA) method \cite{Liu2016_SCA, Ferreira2022, Ferreira2023}, a type of reduced order model (ROM) that we developed previously.
We have shown the accuracy and computational time comparison in \Cref{fig:sca_vs_dns_time_error_linear_solver}, and the quadratic scaling of SCA with the discretization level (number of clusters $N_\text{cluster}$). 

\begin{figure}[htbp]
    \centering

    \begin{subfigure}[t]{0.33\textwidth}
        \centering
        \includegraphics[width=\textwidth]{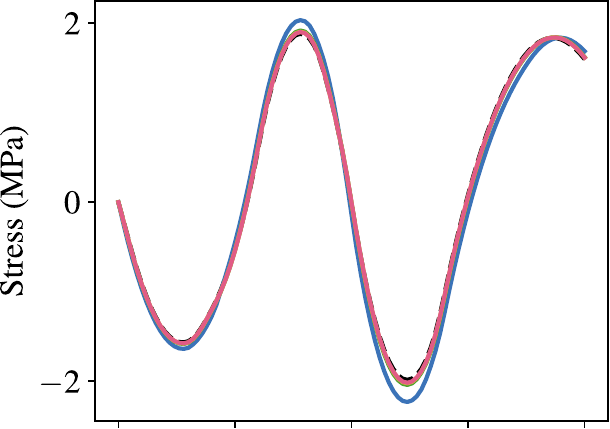}
        \caption{}
        \label{fig:difference_sca_vs_dns_results_linear_solver_a}
    \end{subfigure}
    \hspace{0.005\textwidth}
    \begin{subfigure}[t]{0.322\textwidth}
        \centering
        \includegraphics[width=\textwidth]{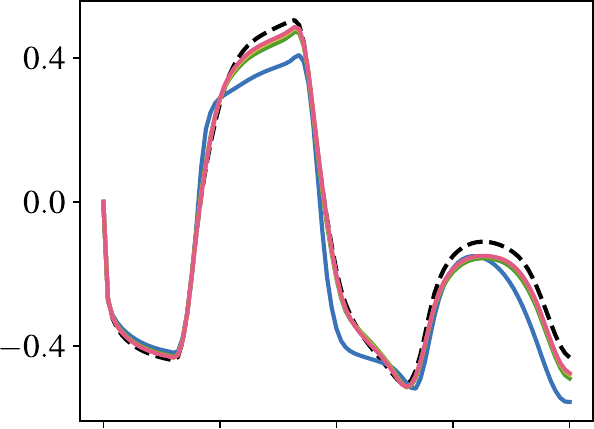}
        \caption{}
        \label{fig:difference_sca_vs_dns_results_linear_solver_b}
    \end{subfigure}
    \hspace{0.01\textwidth}
    \begin{subfigure}[t]{0.308\textwidth}
        \centering
        \includegraphics[width=\textwidth]{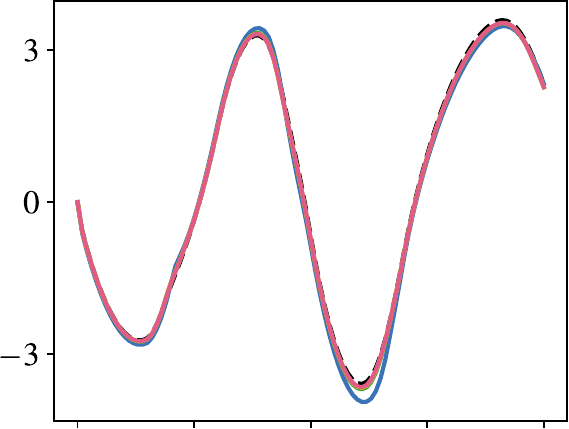}
        \caption{}
        \label{fig:difference_sca_vs_dns_results_linear_solver_c}
    \end{subfigure}

    \begin{subfigure}[t]{0.33\textwidth}
        \centering
        \includegraphics[width=\textwidth]{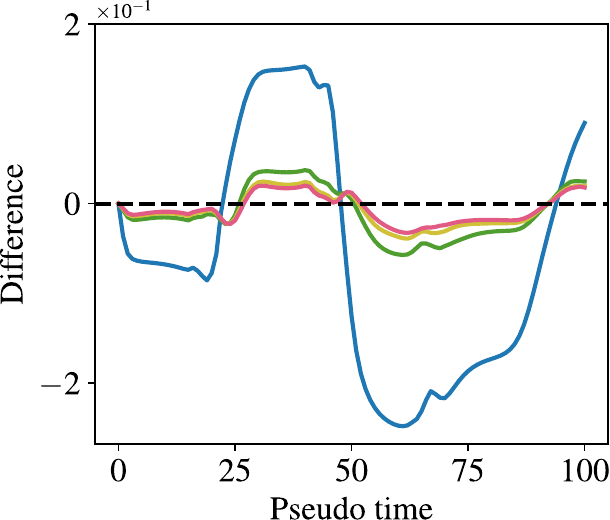}
        \caption{}
        \label{fig:difference_sca_vs_dns_results_linear_solver_d}
    \end{subfigure}
    \hspace{0.005\textwidth}
    \begin{subfigure}[t]{0.322\textwidth}
        \centering
        \includegraphics[width=\textwidth]{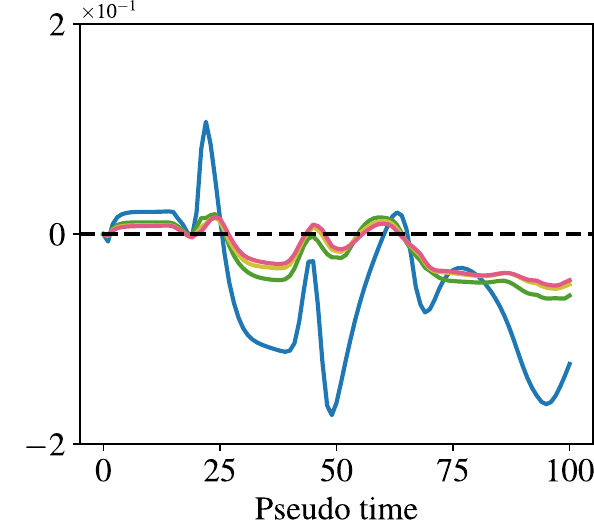}
        \caption{}
        \label{fig:difference_sca_vs_dns_results_linear_solver_e}
    \end{subfigure}
    \hspace{0.01\textwidth}
    \begin{subfigure}[t]{0.308\textwidth}
        \centering
        \includegraphics[width=\textwidth]{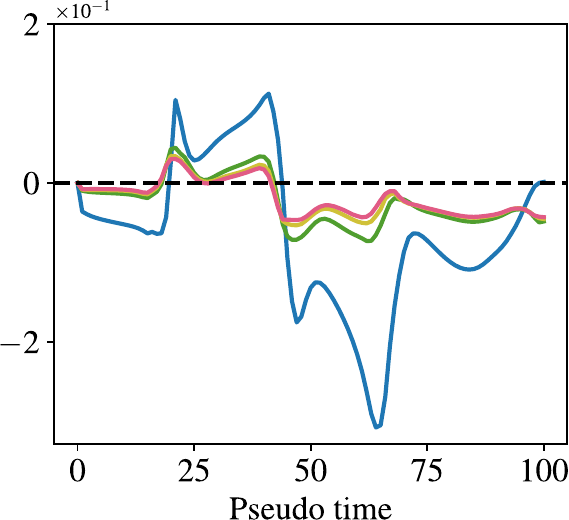}
        \caption{}
        \label{fig:difference_sca_vs_dns_results_linear_solver_f}
    \end{subfigure}
    \vskip 5pt
    \begin{minipage}{0.9\textwidth}
        \centering
        \includegraphics[width=\textwidth]{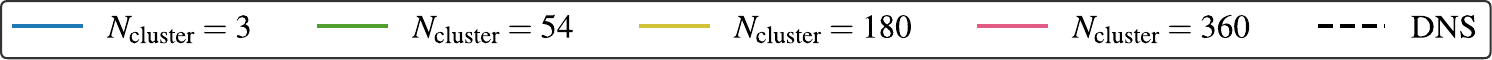}
    \end{minipage}

    \caption{Bias error introduced by \textbf{SCA-RVE} when compared to \textbf{DNS-RVE} and considering different numbers of clusters for the SCA method. The columns from left to right refer to the different stress components: $\sigma_{11}$, $\sigma_{12}$, and $\sigma_{22}$, respectively.}
    \label{fig:difference_sca_vs_dns_results_linear_solver}
\end{figure}

In this section, we further analyze the differences between $\overline{\text{DNS}}$ and $\overline{\text{ROM}}$ when varying the $N_\text{cluster}$ for the SCA method -- see \Cref{fig:difference_sca_vs_dns_results_linear_solver}. Intuitively, the difference reduces rapidly when the number of clusters increases. The difference is large for $N_\text{cluster}=3$, but it reduces significantly for $N_\text{cluster}=54$. After that, the improvement of the SCA with more clusters is limited, despite the dramatic increase in computational cost (\Cref{fig:sca_vs_dns_time_error_linear_solver}). Therefore, we employ  $N_\text{cluster}=3$ and $N_\text{cluster}=18$, serving as lower quality but faster LF data, and as higher quality LF data but relatively slower, respectively.

\newpage

\section{Cooperative training of VeBRNNs}
\label{sec:cooperative_uq_training}

Training Bayesian models to provide good estimates of aleatoric and epistemic uncertainties is not trivial. Although BNNs are reported to be capable of modeling both aleatoric and epistemic uncertainties, state-of-the-art approaches either treat the noise as a fixed hyperparameter under a homoscedastic assumption \cite{Wilson2020}, or embed it directly into the likelihood for end-to-end Bayesian inference \cite{Gal2016}. We recently proposed a more efficient strategy to quantify aleatoric uncertainty \cite{yi2025cooperativebayesianvariancenetworks}. The idea consists of a three-step cooperative training strategy, in which we first train a deterministic neural network for the mean only, then train another variance estimation (Ve) network deterministically for the aleatoric uncertainty, and finally execute Bayesian inference to update the mean and epistemic uncertainty. Furthermore, the second and third steps can be iterated if necessary. The cooperative training strategy for uncertainty disentanglement is summarized in the main text -- see \Cref{alg:CUQalgorithm}.

\subsection{Step 1: Mean network training}

The first step of the cooperative training strategy involves training a mean network that follows the same procedure of conventional RNN training by minimizing the mean squared error (MSE) loss between the RNN outputs and the learning objective. As noted in \Cref{rmk:coop_training_rmk}, the specific inputs and outputs of the RNN may vary depending on the architecture, as summarized in \Cref{tab:architecture_summary}. For clarity, we denote the input as $\mathbf{x}$ and the output as $\mathbf{y}$. Without any loss of generality, we demonstrate the proposed framework by adopting a GRU model \cite{Mozaffarplasticity2019, Dekhovich2023} due to its simplicity and effectiveness, which is written as:

\begin{align}
    \label{eq:gru_cell}
    \mathbf{z}_t         & = \mathrm{sigmoid}(\mathbf{W}_{xz} \mathbf{x}_t + \mathbf{b}_{xz} + \mathbf{W}_{hz} \mathbf{h}_{t-1}  +\mathbf{b}_{hz} ),    \\
    \mathbf{r}_t         & = \mathrm{sigmoid}(\mathbf{W}_{xr} \mathbf{x}_t + \mathbf{b}_{xr} + \mathbf{W}_{hr} \mathbf{h}_{t-1} + \mathbf{b}_{hr}) ,    \\
    \mathbf{\tilde{h}}_t & = \tanh \left(\mathbf{W}_{xh} \mathbf{x}_t + \mathbf{b}_{xh} +  \mathbf{r}_t \odot \left( \mathbf{W}_{hh}\mathbf{h}_{t-1} + \mathbf{b}_{hh} \right) \right), \\
    \mathbf{h}_t         & =  \mathbf{z}_t \odot \mathbf{h}_{t-1} + (1-\mathbf{z}_t) \odot \mathbf{\tilde{h}}_t,
\end{align}

\noindent where $\mathbf{h}_0 = \mathbf{0}$, $\mathbf{x}_t$ is the input vector at the time (or pseudo-time) step $t$, $\mathbf{h}_t$  is the output in the hidden state corresponding to that time step, and the final decoded prediction of the quantity of interest is $f(\mathbf{x}_t)  =  \mathbf{W}_{ho} \mathbf{h}_t + \mathbf{b}_{ho}$. With this step, we use \emph{Adam} \cite{Kingma2014} to optimize the MSE loss between the decoded $f(\mathbf{x}_t)$ and $\mathbf{y}_t$, finding the point estimate of the learnable parameters $\bm{\theta} = [\bm{W}, \bm{b}]$. Thus, the loss function can be given explicitly as:
\begin{equation}
\label{eq:mse_loss}
    \mathcal{L}_\text{MSE}(\bm{\theta})  = \frac{1}{NT}\sum_{n=1}^{N} \sum_{t=1}^{T} \left(\mathbf{y}_{n,t}-f(\mathbf{x}_{n,t};\bm{\theta})\right)^2
\end{equation}
where $N$ is the number of training paths, $T$ is the total pseudo time step.

\subsection{Step 2: Variance network training for aleatoric uncertainty estimation}
\label{sec:variance network}

Once the mean is obtained from the first step, a variance network (also a GRU) is trained with the mean prediction fixed, as described in \Cref{alg:CUQalgorithm}. A crucial detail that facilitates stable training of this variance estimation network lies in its output representation. Specifically, the network does not directly predict the variance $s^2(\mathbf{x})$ as in \Cref{eq:problem_setup}. Instead, it is trained to model a Gamma distribution, enabling a more stable and expressive representation of aleatoric uncertainty. This follows from our previous proof that the square residual $\bm{r}_{t} = \left(f(\mathbf{x}_{t}; \bm{\theta}) - \mathbf{y}_{t}\right)^2$ follows a Gamma distribution since $\mathbf{y}_{t}$ is Gaussian. The corresponding Negative log-likelihood loss to train the variance network is defined by:
\begin{equation}
    \label{eq:gamma_loss}
    \mathcal{L}_\text{Gamma}(\bm{\phi})= \sum_{n=1}^{N}\sum_{t=1}^T \left[ \alpha(\mathbf{x}_{n,t}; \bm{\phi}) \log\frac{\lambda(\mathbf{x}_{n,t}; \bm{\phi})}{\Gamma(\alpha\left(\mathbf{x}_{n,t}; \bm{\phi})\right)} - \left(\alpha(\mathbf{x}_{n,t}; \bm{\phi}\right) - 1) \log \bm{r}_{n,t}
        + \frac{\lambda(\mathbf{x}_{n,t}; \bm{\phi})}{\bm{r}_{n,t}} \right]
\end{equation}
where $\alpha(\mathbf{x}_{t}; \bm{\phi}) > 0$ and $\lambda(\mathbf{x}_{t}; \bm{\phi}) > 0$ are the shape and rate parameters of the Gamma distribution.  After training the variance network, the expected value of the Gamma distribution becomes the desired heteroscedastic variance, which leads to:

\begin{equation}
    \label{eq:aleatoric_variance}
    s^2(\mathbf{x}_{1:T}; \bm{\phi}) = \frac{\alpha(\mathbf{x}_{1:T}; \bm{\phi})}{\lambda(\mathbf{x}_{1:T}; \bm{\phi})}
\end{equation}

\subsection{Step 3: Bayesian recurrent neural network training for updated mean and epistemic uncertainty}
\label{sec:Bayesian_RNN}
After completing \textbf{Step 2} training for the aleatoric variance, we aim to update the mean obtained from \textbf{Step 1} as well as get the epistemic uncertainty through Bayesian inference.  The distinction between deterministic (\textbf{Step 1}) and Bayesian treatments is that the latter places the parameters in probabilistic distributions, whereas the deterministic approach treats them as real numbers \footnote{Theoretically, deterministic training can be regarded as finding the mode of PPD, refers to Maximum A Posterior (MAP); or assume all neural parameters following delta distributions}. The posterior distribution of the parameters can be obtained using Bayes' rule, expressed by \cite{Jospin2022},

\begin{equation} \label{eq:Bayes rule}
    p(\bm{\theta}|\mathcal{D}) = \frac{p(\mathcal{D}| \bm{\theta})p(\bm{\theta})}{p(\mathcal{D})}; \quad  p(\mathcal{D}) = \int p(\mathcal{D} | \bm{\theta}) p(\bm{\theta}) d\bm{\theta}
\end{equation}

\noindent where $p(\mathcal{D}| \bm{\theta})$ denotes the observation likelihood,  $p(\bm{\theta})$ is the prior of the neural parameters, and $p(\mathcal{D})$ is the marginal likelihood. Following common practice in BNNs, we adopt a unit Gaussian distribution prior for neural parameters; moreover, $p(\mathcal{D}| \bm{\theta})$ is assumed to be independent and identically distributed (i.i.d.) Gaussian distribution:

\begin{equation}
    \label{sec:likelihood}
    p(\mathcal{D}|\bm{\theta}) =  \prod_{i=n}^{N}  \prod_{t=1}^{T} \mathcal{N}\left(\mathbf{y}_{i,t}; f(\mathbf{x}_{n,t} ;\bm{\theta} ), s^2(\mathbf{x}_{n,t};\bm{\phi}) \right)
\end{equation}

\noindent where $s^2(\mathbf{x};\bm{\phi})$ is the aleatoric (data) uncertainty (from \textbf{step 2}).
Obtaining the posterior distribution in \Cref{eq:Bayes rule} is difficult because of the intractable high-dimensional probability integral. Thus, approximate methods such as MCMC \cite{murphy2022probabilistic_advanced} or VI \cite{Blei2017, murphy2022probabilistic_advanced} are required; we provide details in the Appendix \ref{sec:Bayesian_inference_methods}.  After obtaining the posterior distribution of the parameters in \Cref{eq:Bayes rule}, the predicted posterior distribution for any unknown point $\mathbf{x^\prime}_{1:T}$ can be computed as

\begin{equation} \label{eq:prediction_equation}
    p\left (\mathbf{y^\prime}_{1:T} \mid \mathbf{x^\prime}_{1:T}, \mathcal{D} \right )
    = \int p \left (\mathbf{x^\prime}_{1:T} \mid \mathbf{x^\prime}_{1:T}, \boldsymbol{\theta}\right)  p \left ( \boldsymbol{\theta} \mid \mathcal{D} \right ) \mathrm{d} \boldsymbol{\theta}.
\end{equation}
\noindent The predicted mean can be approximately obtained as

\begin{equation}
    \mathbb{E} (\mathbf{y^\prime}_{1:T} \mid \mathbf{x^\prime}_{1:T}, \mathcal{D} ) = \mathbb{E}_{p(\bm{\theta} \mid D)} \left[ f(\mathbf{x}_{1:T})\right],
\end{equation}
\noindent and predicted variance as
\begin{align}
    \label{sec:predicted_var}
    \text{Var}(\mathbf{y^\prime}_{1:T} \mid \mathbf{x^\prime}_{1:T}, \mathcal{D}) & = \underbrace{\mathbb{E}_{p(\bm{\theta} \mid D)} \left[s^2(\mathbf{x}_{1:T};\bm{\phi}) \right]}_{\text{Aleatoric Uncertainty}} + \underbrace{\text{Var}_{p(\bm{\theta} \mid D)} \left[ f(\mathbf{x}_{1:T}) \right]}_{\text{Epistemic Uncertainty}} \notag \\
    & = \underbrace{s^2(\mathbf{x}_{1:T};\bm{\phi})}_{\text{Aleatoric Uncertainty}} + \underbrace{\text{Var}_{p(\bm{\theta} \mid D)} \left[ f(\mathbf{x}_{1:T}) \right]}_{\text{Epistemic Uncertainty}}.
\end{align}

\newpage

\section{Alternative multi-fidelity models and comprehensive investigations}
\label{sec:mf_model_comparison}

\subsection{Alternative multi-fidelity models}

\begin{figure}[h]
    \centering
    \includegraphics[width=0.45\textwidth]{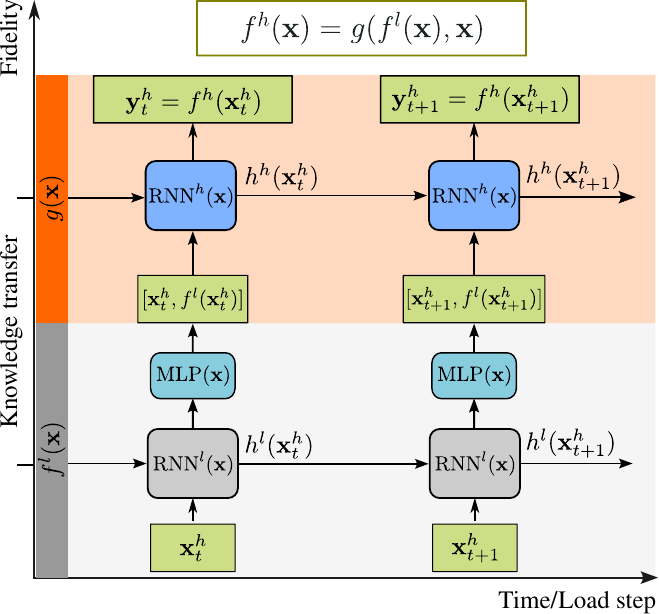}
    \caption{Schematic of the proposed multi-fidelity recurrent neural network architecture: \emph{MF-Nest-Output}. This configuration follows the formulation $f^h(\mathbf{x}) = g(f^l(\mathbf{x}), \mathbf{x})$, where $g(\cdot)$ is a non-linear transfer learning model—specifically, another RNN. This architecture does not include an explicit residual model. Meanwhile, the transfer model takes as input the decoded stress prediction from the low-fidelity RNN, concatenated with the high-fidelity strain inputs.}
    \label{fig:mf_nest_output}
\end{figure}

\begin{figure}[h]
    \centering
    \includegraphics[width=0.45\textwidth]{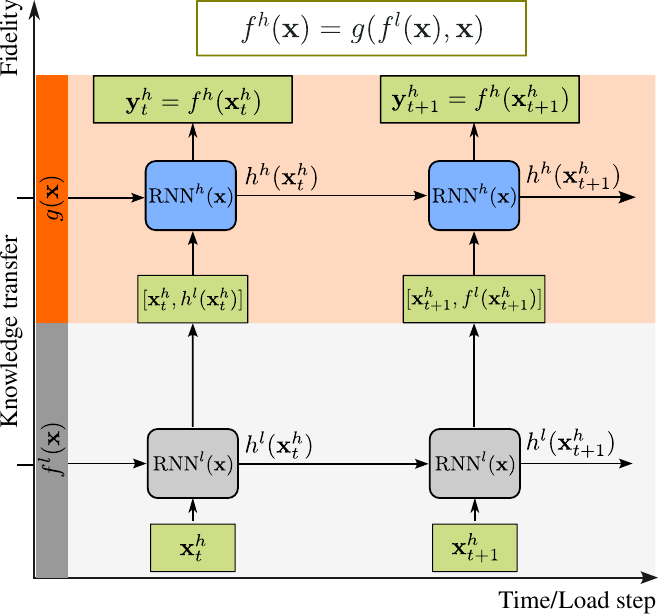}
    \caption{Schematic of the proposed multi-fidelity recurrent neural network architectures: \emph{MF-Nest-Hidden}.   
    This configuration follows the formulation $f^h(\mathbf{x}) = g(f^l(\mathbf{x}), \mathbf{x})$, where $g(\cdot)$ is also another RNN and without a residual model. Instead, the transfer model takes as input the hidden state from the low-fidelity RNN, concatenated with the high-fidelity strain inputs.}
    \label{fig:mf_nest_hidden}
\end{figure}

\begin{figure}[h]
    \centering
    \includegraphics[width=0.45\textwidth]{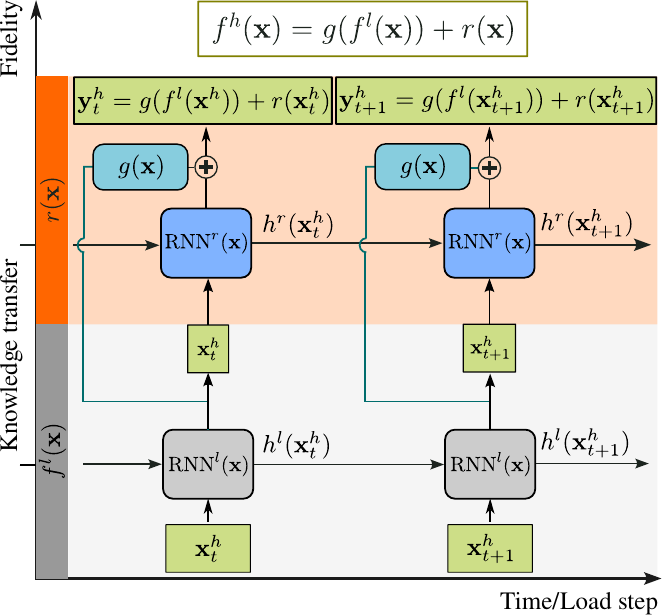}
    \caption{Schematic of the proposed multi-fidelity recurrent neural network architecture: \emph{MF-Residual-Original}. This configuration adopts the formulation $f^h(\mathbf{x}) = g(f^l(\mathbf{x})) + r(\mathbf{x})$, where $g(\cdot)$ is a linear transfer learning model and $r(\mathbf{x})$ is a residual model implemented as another RNN. Unlike the configuration shown in \Cref{fig:proposed_framework}, this variant trains the residual model using only the original high-fidelity strain input.}
    \label{fig:mf_residual_original}
\end{figure}

We described a multi-fidelity approach where an RNN was first trained on the LF data, followed by a linear transfer learning model, and another RNN was used to learn the residual between the high-fidelity response and the transfer-learned prediction. Experiments in \Cref{sec:numerical_experiments} demonstrated its effectiveness across various datasets. However, we also clarified that multi-fidelity architecture is one particular architecture drawn from \Cref{fig:training_flowchart}. Herein, we provide three other variants that can be obtained following \Cref{eq:general_mf_formula}, showing in 
\Cref{fig:mf_nest_output}, \Cref{fig:mf_nest_hidden}, and \Cref{fig:mf_residual_original}.

As shown in \Cref{fig:mf_nest_output}, \Cref{fig:mf_nest_hidden}, and \Cref{fig:mf_residual_original}, the key difference between the configurations using $f^h(\mathbf{x}) = g(f^l(\mathbf{x}), \mathbf{x})$ and $f^h(\mathbf{x}) = g(f^l(\mathbf{x})) + r(\mathbf{x})$ lies in the knowledge transfer. Specifically, for the configurations following $f^h(\mathbf{x}) = g(f^l(\mathbf{x}), \mathbf{x})$ (\emph{MF-Nest-Output} and \emph{MF-Nest-Hidden}), the transfer learning model $g(\cdot)$ is implemented as another RNN, and no residual model is used. In contrast, the configurations following $f^h(\mathbf{x}) = g(f^l(\mathbf{x})) + r(\mathbf{x})$ (\emph{MF-Residual-Original} and \emph{MF-Residual-Hidden}, as shown in \Cref{fig:proposed_framework}) adopt a linear transfer learning model\footnote{We adopt the identity function as the transfer learning model in this work.} and additionally employ a residual model—another RNN—to learn the discrepancy between the high-fidelity and low-fidelity predictions.

Another subtle distinction exists within each configuration regarding the inputs and outputs for the knowledge transfer module. For the configurations of \emph{MF-Nest-Output} and \emph{MF-Nest-Hidden}, the input to the transfer learning model can be formed by concatenating $\mathbf{x}^h$ with the decoded prediction of LF at $\mathbf{x}^h$ (i.e., the LF predicted stress) or with the hidden state at $\mathbf{x}^h$ of the LF model directly. Similarly, \emph{MF-Residual-original} and \emph{MF-Residual-Hidden} can select to concatenate the additional hidden state at $\mathbf{x}^h$ of the LF model or use $\mathbf{x}^h$ only. Based on these variations, we summarize the uniqueness of the inputs and outputs for transfer learning of each architecture in \Cref{tab:architecture_summary}.

\begin{table}[h]
    \centering
    \caption{Summary of inputs and outputs for different multi-fidelity model variants from \Cref{fig:training_flowchart}}
    \label{tab:architecture_summary}
    \begin{tabular}{llll}
        \toprule
        Variation                     & Input                               & Output / Learning target              & Formula                                                \\
        \midrule
        \emph{MF-Nest-Output}     & $[\mathbf{x}^h, f^l(\mathbf{x}^h)]$ & $\mathbf{y}^h$                        & $f^h(\mathbf{x}) = g(f^l(\mathbf{x}), \mathbf{x})$     \\
        \emph{MF-Nest-Hidden}     & $[\mathbf{x}^h, h^l(\mathbf{x}^h)]$ & $\mathbf{y}^h$                        & $f^h(\mathbf{x}) = g(f^l(\mathbf{x}), \mathbf{x})$     \\
        \emph{MF-Residual-Original} & $\mathbf{x}^h$                      & $\mathbf{y}^h - g(f^l(\mathbf{x}^h))$ & $f^h(\mathbf{x}) = g(f^l(\mathbf{x})) + r(\mathbf{x})$ \\
        \emph{MF-Residual-Hidden}   & $[\mathbf{x}^h, h^l(\mathbf{x}^h)]$ & $\mathbf{y}^h - g(f^l(\mathbf{x}^h))$ & $f^h(\mathbf{x}) = g(f^l(\mathbf{x})) + r(\mathbf{x})$ \\
        \bottomrule
    \end{tabular}
\end{table}

\subsection{Comprehensive investigation on multi-fidelity architectures}
\label{sec:mf_architecture_selection}

We proposed four distinct multi-fidelity models in this paper including \Cref{fig:proposed_framework}, namely \emph{MF-Nest-Hidden}, \emph{MF-Nest-Output}, \emph{MF-Residual-Hidden}, and \emph{MF-Residual-Original}, with detailed input and learning objective summarized in \Cref{tab:architecture_summary}. In this section, we conduct a comprehensive investigation to compare the performance of these configurations using \textbf{Dataset 2}. Specifically, an RNN+RNN architecture is first trained in \Cref{sec:configuration_deter_comparison}, and followed by an RNN+VeBRNN architecture trained in \Cref{sec:configuration_bayesian_comparison}.

\subsubsection{RNN+RNN architecture training}
\label{sec:configuration_deter_comparison}

\begin{figure}[h]
    \centering

    \begin{subfigure}[t]{0.375\textwidth}
        \centering
        \includegraphics[width=\textwidth]{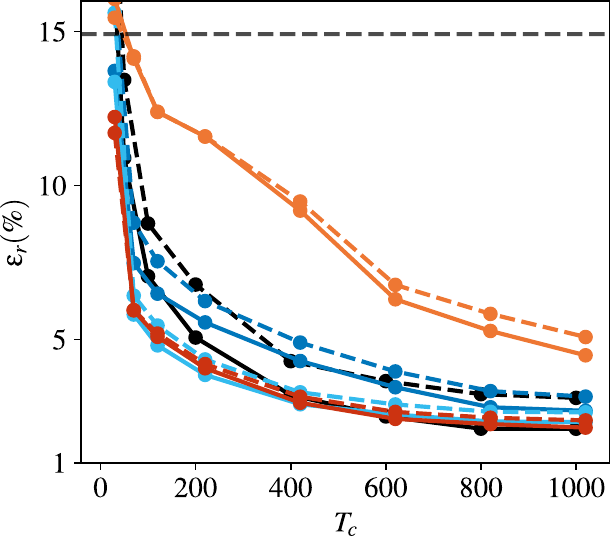}
        \caption{$N_\text{cluster}=3$}
        \label{fig:mf_deter_different_architecture_comparison_appendix_a}
    \end{subfigure}
    \hspace{0.03\textwidth}
    \begin{subfigure}[t]{0.35\textwidth}
        \centering
        \includegraphics[width=\textwidth]{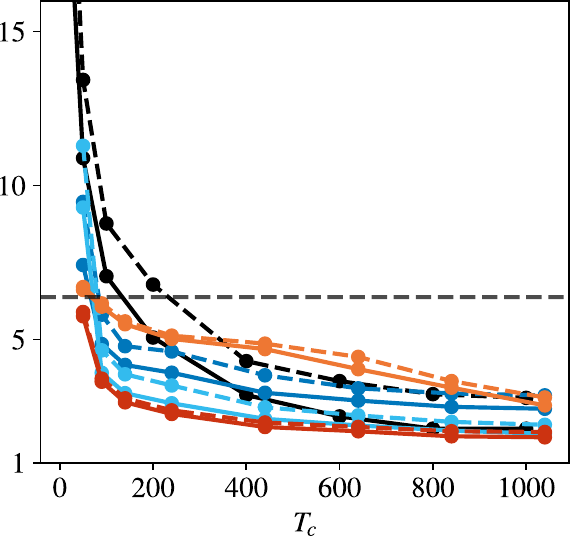}
        \caption{$N_\text{cluster}=18$}
        \label{fig:mf_deter_different_architecture_comparison_appendix_b}
    \end{subfigure}
    \vskip 5pt
    \begin{minipage}{0.8\textwidth}
        \centering
        \includegraphics[width=\textwidth]{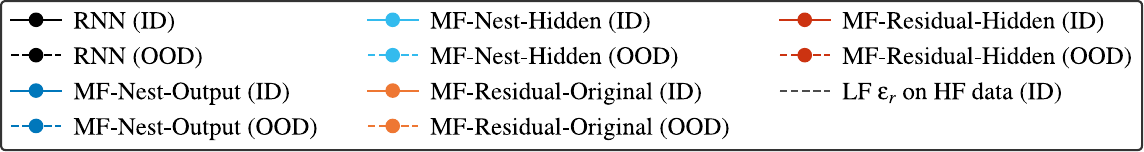}
    \end{minipage}
    \caption{Comparison of different knowledge transfer configurations for \textbf{Dataset 2} with RNN+RNN training. The single-fidelity RNN training based on $\overline{\text{DNS}}$ results are shown with black lines for comparison. Meanwhile, the results of RNN+RNN training have offsets of $33.3$ and $55.5$ for $N_\text{cluster} = 3$ and $N_\text{cluster} = 18$ in the x-axis, considering the computational cost of low-fidelity data.}
    \label{fig:mf_deter_different_architecture_comparison_appendix}
\end{figure}

We follow the same setup as \Cref{fig:mf_problem_1_deter_comparison} that uses all $2000$ low-fidelity training paths to train the low-fidelity RNN model first. Then, we gradually increase the number of high-fidelity training paths from $10$ to $1000$. The results are presented in \Cref{fig:mf_deter_different_architecture_comparison_appendix}.

Different colors in \Cref{fig:mf_deter_different_architecture_comparison_appendix} represent different multi-fidelity model configurations proposed according to \Cref{eq:general_framework} and \Cref{fig:training_flowchart}, where it is evident that a better LF model improves the performance of MF modeling. Among them, \emph{MF-Residual-Hidden} is outperforming other multi-fidelity model configurations across all experiments (achieving the lowest $\epsilon_r$ across all $T_c$) shown in \Cref{fig:mf_deter_different_architecture_comparison_appendix}. Conversely, \emph{MF-Residual-Original} has the worst performance, showing an apparent gap to other multi-fidelity model configurations. As for the other two multi-fidelity model configurations--\emph{MF-Nest-Hidden} and \emph{MF-Nest-Output}--which rely on nonlinear transfer learning without a residual model, struggle to give robust predictions in cases where the number of high-fidelity training points is limited.

Interestingly, we observe that the multi-fidelity model configurations using hidden-state-based knowledge transfer—\emph{MF-Nest-Hidden} and \emph{MF-Residual-Hidden}—consistently outperform their output-state-based counterparts—\emph{MF-Nest-Output} and \emph{MF-Residual-Original}, respectively. This is a noteworthy finding, as it suggests that directly transferring the hidden states from the low-fidelity model to the high-fidelity model is more effective than transferring decoded outputs (i.e., stresses). This observation is counterintuitive to previous studies in data-driven constitutive modeling, which often argue that hidden state representations are elusive and lack interpretability, making them unsuitable for transfer learning. Our results indicate that transferring hidden states bypasses the need to decode low-fidelity outputs and re-encode them into the HF representation space, as shown in \Cref{fig:mf_nest_output}. Furthermore, hidden states in \Cref{fig:mf_nest_hidden} and \Cref{fig:proposed_framework} typically provide a high-dimensional representation of the underlying history-dependent material behavior, and thus may encode richer and more informative features than the observable outputs alone.

We also note that having a linear knowledge transfer, such as \emph{MF-Residual-Original} and \emph{MF-Residual-Hidden}, improves the OOD performance. Specifically, the OOD  $\epsilon_r$ of \emph{MF-Residual-Original} and \emph{MF-Residual-Hidden} are almost the same as the ID $\epsilon_r$, while \emph{MF-Nest-Output} and \emph{MF-Nest-Hidden} have a larger discrepancy between ID and OOD $\epsilon_r$. This observation is consistent with our previous findings based on feedforward neural networks \cite{yi2024practicalmultifidelitymachinelearning}.

\subsubsection{RNN+VeBRNN architecture training}
\label{sec:configuration_bayesian_comparison}

\begin{figure}[h]
    \centering
    \hspace*{0.04\textwidth} 
    \begin{subfigure}[t]{0.33\textwidth}
        \centering
        \includegraphics[width=\textwidth]{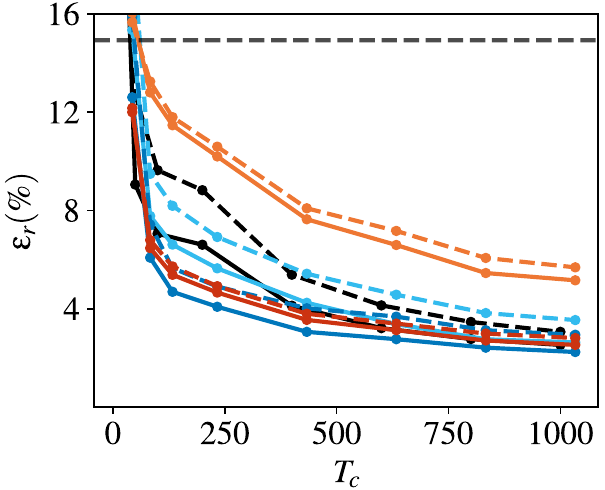}
        \caption{Mean estimate metric ($\downarrow$)}
        \label{fig:comparison_mf_architectures_sca_rve_appendix_2_1_a}
    \end{subfigure}
    \begin{subfigure}[t]{0.33\textwidth}
        \centering
        \includegraphics[width=\textwidth]{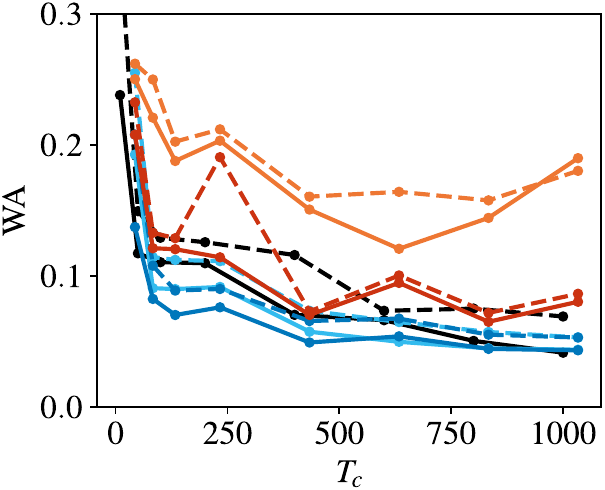}
        \caption{Aleatoric uncertainty metric ($\downarrow$)}
        \label{fig:comparison_mf_architectures_sca_rve_appendix_2_1_b}
    \end{subfigure}
    \vspace{0.3cm}
    \begin{subfigure}[t]{0.33\textwidth}
        \centering
        \includegraphics[width=\textwidth]{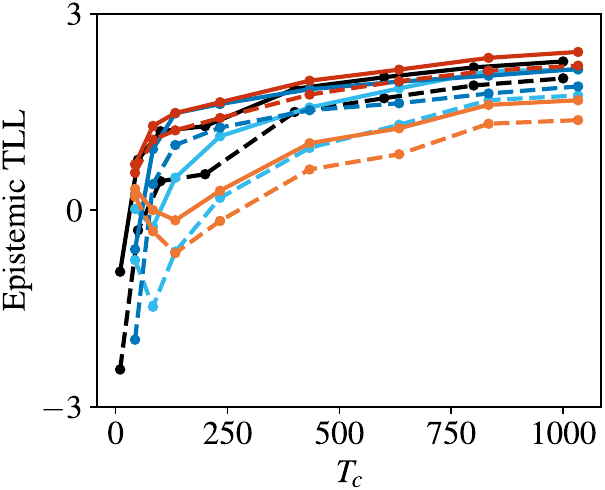}
        \caption{Epistemic uncertainty metric ($\uparrow$)}
        \label{fig:comparison_mf_architectures_sca_rve_appendix_2_1_c}
    \end{subfigure}
    \begin{subfigure}[t]{0.33\textwidth}
        \centering
        \includegraphics[width=\textwidth]{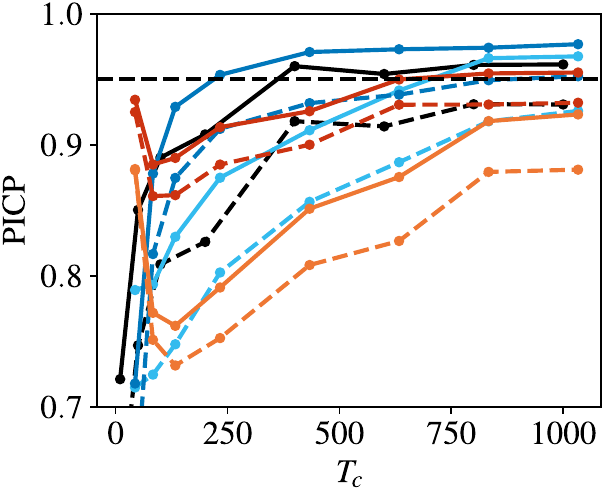}
        \caption{Epistemic uncert. metric ($\rightarrow 0.95$)}
        \label{fig:comparison_mf_architectures_sca_rve_appendix_2_1_d}
    \end{subfigure}
    \begin{subfigure}[t]{0.33\textwidth}
        \centering
        \includegraphics[width=\textwidth]{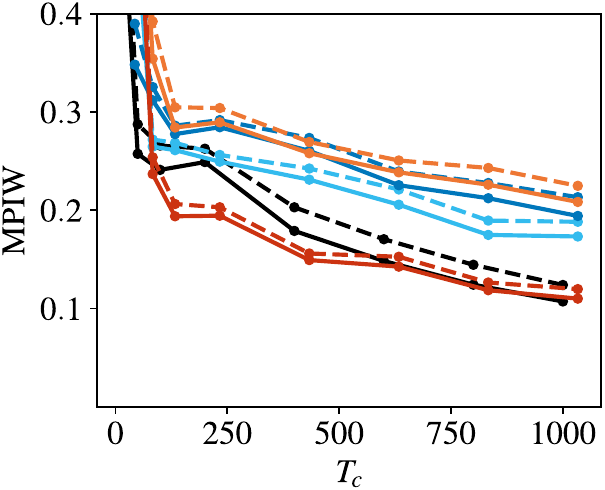}
        \caption{Epistemic uncertainty metric ($\downarrow$)}
        \label{fig:comparison_mf_architectures_sca_rve_appendix_2_1_e}
    \end{subfigure}
    \vskip 5pt
    \begin{minipage}{\textwidth}
        \centering
        \includegraphics[width=\textwidth]{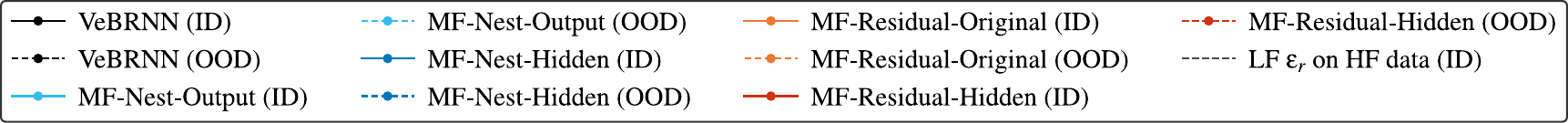}
    \end{minipage}
    \caption{Comparison of different multi-fidelity model configurations proposed in this paper for \textbf{Dataset 2} with RNN+VeBRNN architecture under low-fidelity of $N_\text{cluster}=3$}
    \label{fig:mf_bayes_different_architecture_comparison_appendix_2_1}
\end{figure}

\begin{figure}[h]
    \centering
    \hspace*{0.04\textwidth} 
    \begin{subfigure}[t]{0.33\textwidth}
        \centering
        \includegraphics[width=\textwidth]{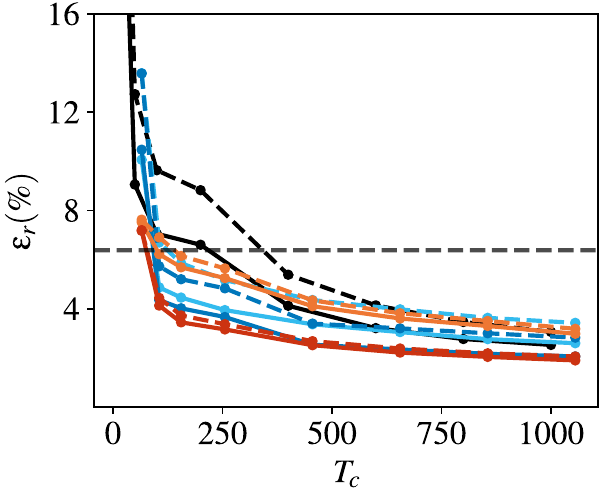}
        \caption{Mean estimate metric ($\downarrow$)}
        \label{fig:comparison_mf_architectures_sca_rve_appendix_12_6_a}
    \end{subfigure}
    \begin{subfigure}[t]{0.33\textwidth}
        \centering
        \includegraphics[width=\textwidth]{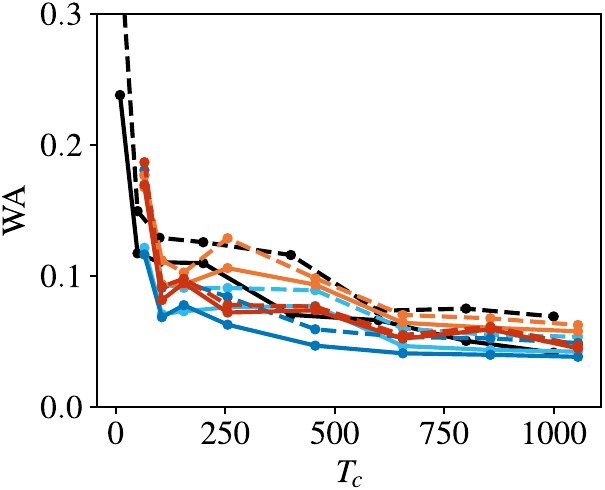}
        \caption{Aleatoric uncertainty metric ($\downarrow$)}
        \label{fig:comparison_mf_architectures_sca_rve_appendix_12_6_b}
    \end{subfigure}
    \vspace{0.3cm}
    \begin{subfigure}[t]{0.33\textwidth}
        \centering
        \includegraphics[width=\textwidth]{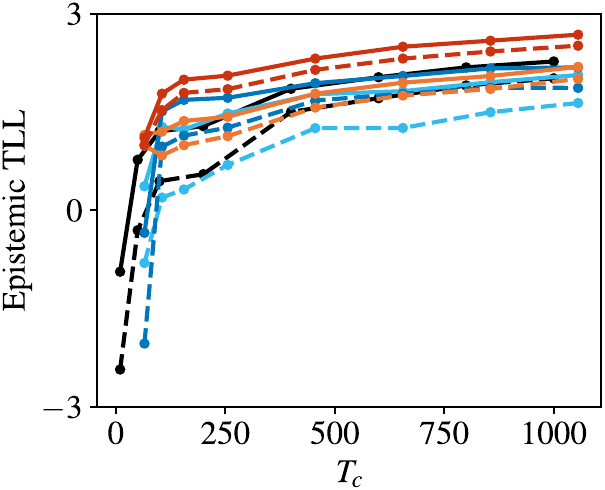}
        \caption{Epistemic uncertainty metric ($\uparrow$)}
        \label{fig:comparison_mf_architectures_sca_rve_appendix_12_6_c}
    \end{subfigure}
    \begin{subfigure}[t]{0.33\textwidth}
        \centering
        \includegraphics[width=\textwidth]{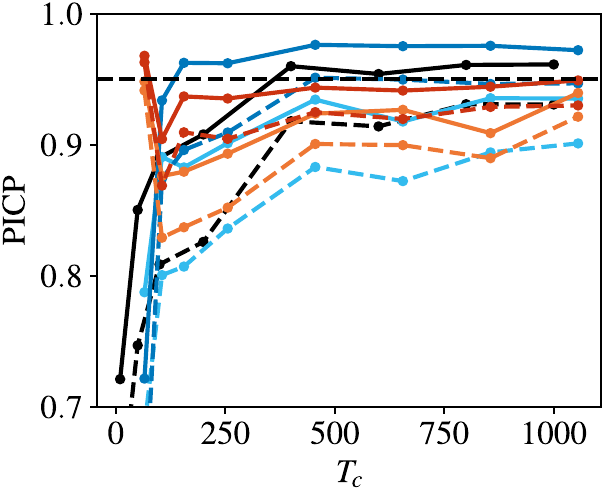}
        \caption{Epistemic uncert. metric ($\rightarrow 0.95$)}
        \label{fig:comparison_mf_architectures_sca_rve_appendix_12_6_d}
    \end{subfigure}
    \begin{subfigure}[t]{0.33\textwidth}
        \centering
        \includegraphics[width=\textwidth]{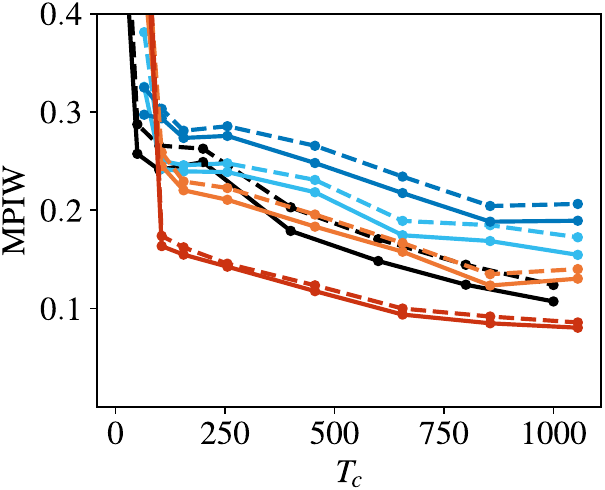}
        \caption{Epistemic uncertainty metric ($\downarrow$)}
        \label{fig:comparison_mf_architectures_sca_rve_appendix_12_6_e}
    \end{subfigure}
    \vskip 5pt
    \begin{minipage}{\textwidth}
        \centering
        \includegraphics[width=\textwidth]{figures/comparison_mf_architectures_sca_rve_appendix_2_1_legend.pdf}
    \end{minipage}
    \caption{Comparison of different multi-fidelity model configurations proposed in this paper for \textbf{Dataset 2} with RNN+VeBRNN architecture under low-fidelity of $N_\text{cluster}=18$}
    \label{fig:mf_bayes_different_architecture_comparison_appendix_12_6}
\end{figure}

In this section, we continue to test the four multi-fidelity model configurations developed in \Cref{sec:mf_model_comparison}, using deterministic training for the low-fidelity data and cooperative training for the high-fidelity data, specifically the RNN+VeBRNN architecture. The results trained based on different low-fidelity data of $\overline{\text{ROM}}$, i.e., $N\text{cluster}=3$ and $N\text{cluster}=18$, are shown in \Cref{fig:mf_bayes_different_architecture_comparison_appendix_2_1} and \Cref{fig:mf_bayes_different_architecture_comparison_appendix_12_6}, respectively.

According to \Cref{fig:mf_bayes_different_architecture_comparison_appendix_2_1} and \Cref{fig:mf_bayes_different_architecture_comparison_appendix_12_6}, the conclusion that the multi-fidelity model performs better with better low-fidelity data resources remains the same for all configurations. Furthermore, employing a cooperative training for the high-fidelity model does not alter the conclusion that \emph{MF-Residual-Hidden} remains the best configuration. Moreover, results on Epistemic TLL and PICP provide strong evidence that \emph{MF-Residual-Hidden} yields more consistent predictions across both ID and OOD data. It is particularly encouraging that the use of a transfer learning based on hidden states gives tighter or similar confidence intervals as shown \Cref{fig:comparison_mf_architectures_sca_rve_appendix_2_1_e} and \Cref{fig:comparison_mf_architectures_sca_rve_appendix_12_6_e}, while leading the PICP close to the preset confidence level of 0.95. Concerning the Wasserstein distance, \emph{MF-Nest-Hidden} is slightly worse than \emph{MF-Nest-Output} and \emph{MF-Nest-Hidden}; however, the prediction shows in \Cref{fig:prediction_of_mf_redisual_on_id_path} that the \emph{MF-Nest-Hidden} already converged to a reasonable level for this deterministic dataset.

In summary, the results in \Cref{sec:configuration_deter_comparison} and \Cref{sec:configuration_bayesian_comparison} demonstrate that \emph{MF-Residual-Hidden} is the best multi-fidelity model configuration among all four configurations proposed in this section. Therefore, we recommend using \emph{MF-Residual-Hidden} for multi-fidelity data-driven plasticity law.

\newpage

\section{Additional experimental results}
\label{sec:additional_experiments}

\subsection{Predictions of RNN on the single-fidelity problem}
\label{sec:additional_sf_problem} 

\begin{figure}[h]
    \centering
    \begin{subfigure}[t]{0.33\textwidth}
        \centering
        \includegraphics[width=\textwidth]{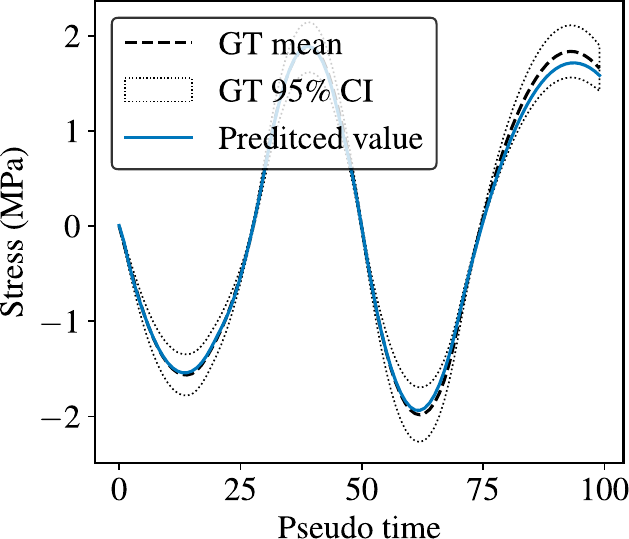}
        \caption{$\sigma_{11}$ }
    \end{subfigure}
    \hspace{0.01\textwidth}
    \begin{subfigure}[t]{0.308\textwidth}
        \centering
        \includegraphics[width=\textwidth]{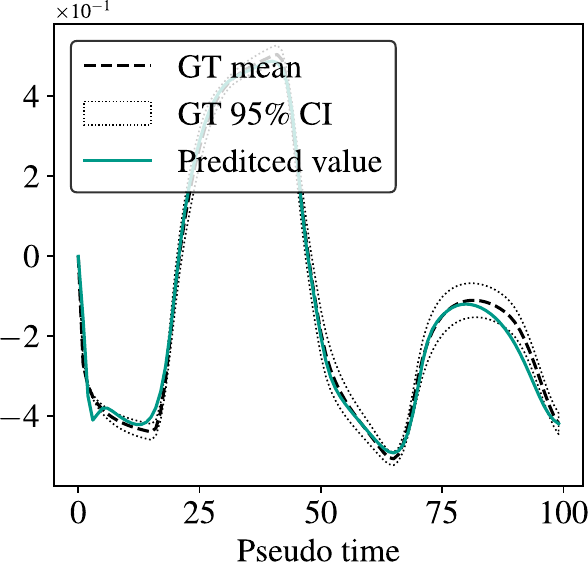}
        \caption{ $\sigma_{12}$ }
    \end{subfigure}
    \hspace{0.01\textwidth}
    \begin{subfigure}[t]{0.308\textwidth}
        \centering
        \includegraphics[width=\textwidth]{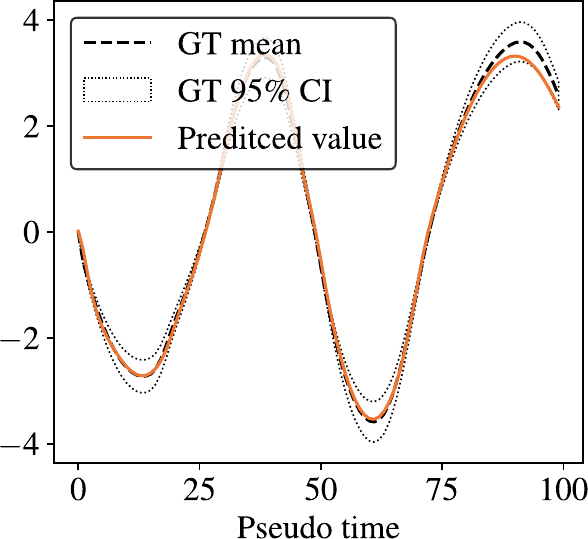}
        \caption{ $\sigma_{22}$ }
    \end{subfigure}
    \caption{Predictions of the RNN with 2000 training paths ($T_c=2000$) on the ID test path }
    \label{fig:sf_mse_training}
\end{figure}

The predictions of RNN with $2000$ training paths obtained by ($\widetilde{\text{DNS}}$) are given in \Cref{fig:sf_mse_training}. 

As shown in \Cref{fig:sf_mse_training}, RNN only predicts a mean estimation for each stress component, and it is clear to see the overfitting in the prediction of $\sigma_{12}$. In addition, it loses the essential information on uncertainty quantification, making it difficult to assess the overall quality of material properties.

\subsection{Additional training results for Dataset 3}
\label{sec:deter_train_mf_problem_2}

\begin{figure}[h]
    \centering
    \begin{subfigure}[t]{0.375\textwidth}
        \centering
        \includegraphics[width=\textwidth]{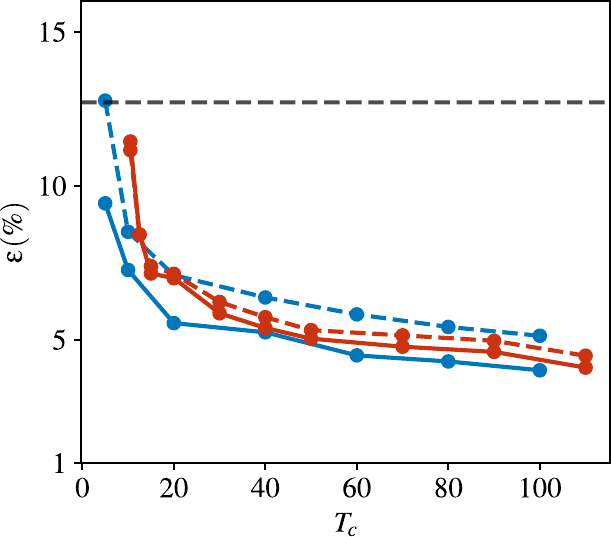}
        \caption{$N_\text{cluster}=3$}
        \label{fig:mf_deter_dns_sve_sca_sve_a}
    \end{subfigure}
    \hspace{0.03\textwidth}
    \begin{subfigure}[t]{0.35\textwidth}
        \centering
        \includegraphics[width=\textwidth]{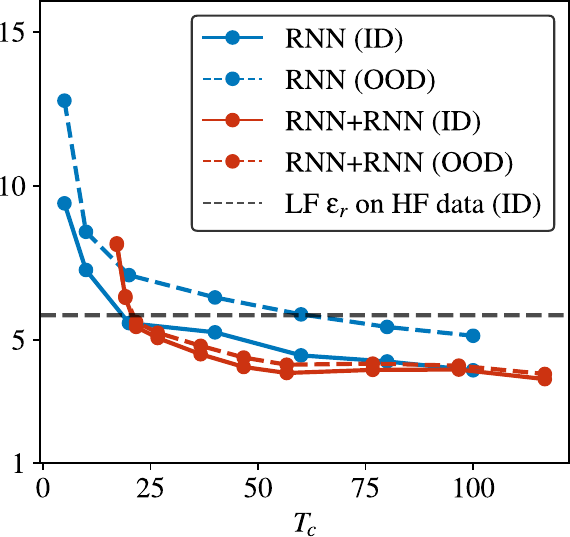}
        \caption{$N_\text{cluster}=18$}
        \label{fig:mf_deter_dns_sve_sca_sve_b}
    \end{subfigure}
    \caption{Results for RNN+RNN model trained on Dataset 3 ($\widetilde{\text{ROM}}$+$\widetilde{\text{DNS}}$) compared to single-fidelity RNN trained on HF data. (a) LF model trained on data obtained by SCA of an SVE with $N_\text{cluster}=3$; (b). LF model trained on data obtained by SCA of an SVE with $N_\text{cluster}=18$.}
    \label{fig:mf_deter_mf_problem_2}
\end{figure}

\begin{figure}[h]
    \centering
    \hspace*{0.04\textwidth} 
    \begin{subfigure}[t]{0.33\textwidth}
        \centering
        \includegraphics[width=\textwidth]{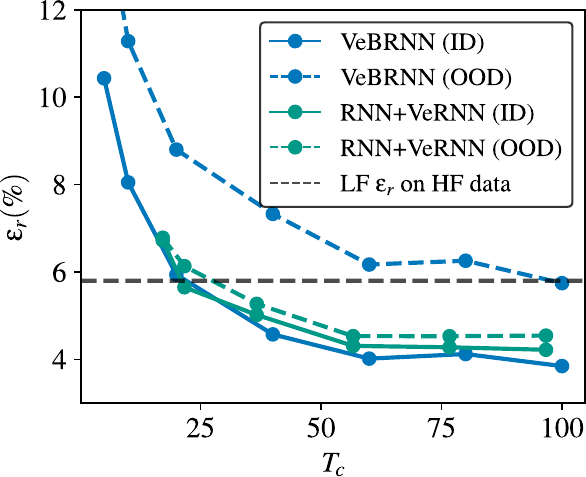}
        \caption{Mean estimate metric ($\downarrow$)}
        \label{fig:mf_uq_dns_rve_dns_sve_rnn_vebrnn_a}
    \end{subfigure}
    \begin{subfigure}[t]{0.33\textwidth}
        \centering
        \includegraphics[width=\textwidth]{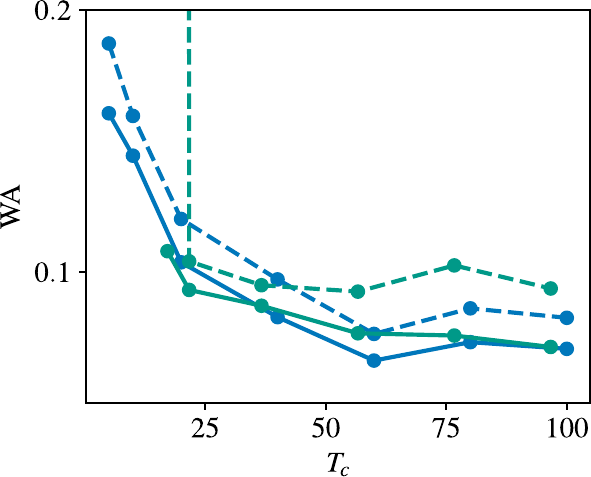}
        \caption{Aleatoric uncertainty metric ($\downarrow$)}
        \label{fig:mf_uq_dns_rve_dns_sve_rnn_vebrnn_b}
    \end{subfigure}
    \vspace{0.3cm}
    \begin{subfigure}[t]{0.33\textwidth}
        \centering
        \includegraphics[width=\textwidth]{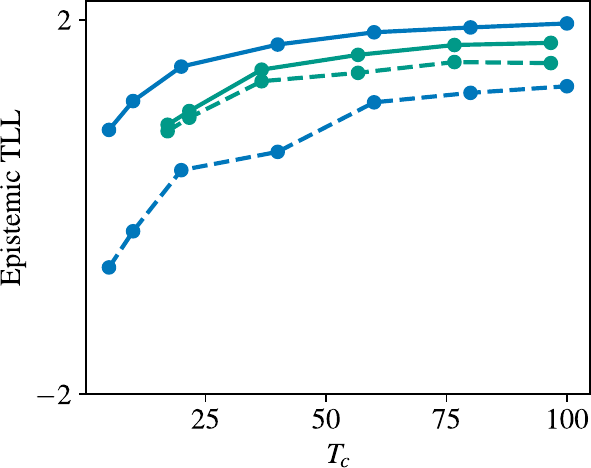}
        \caption{Epistemic uncertainty metric ($\uparrow$)}
        \label{fig:mf_uq_dns_rve_dns_sve_rnn_vebrnn_c}
    \end{subfigure}
    \begin{subfigure}[t]{0.33\textwidth}
        \centering
        \includegraphics[width=\textwidth]{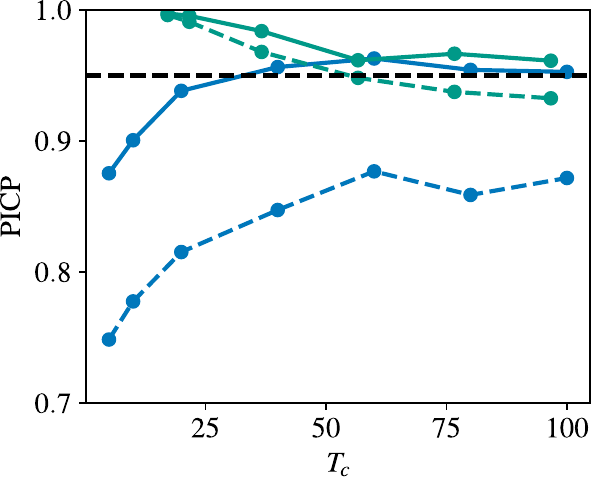}
        \caption{Epistemic uncert. metric ($\rightarrow 0.95$)}
        \label{fig:mf_uq_dns_rve_dns_sve_rnn_vebrnn_d}
    \end{subfigure}
    \begin{subfigure}[t]{0.33\textwidth}
        \centering
        \includegraphics[width=\textwidth]{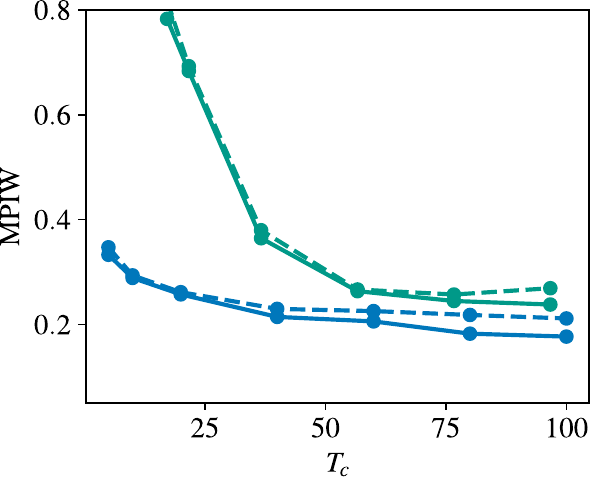}
        \caption{Epistemic uncertainty metric ($\downarrow$)}
        \label{fig:mf_uq_dns_rve_dns_sve_rnn_vebrnn_3}
    \end{subfigure}
    \caption{Results for the RNN+VeBRNN model, when compared with a single-fidelity VeBRNN trained on HF data. The LF data is obtained by SCA of an SVE with $N_\text{cluster}=18$.}

    \label{fig:mf_uq_dns_rve_dns_sve_rnn_vebrnn}
\end{figure}

\Cref{sec:mf_problem_2} presents the VeBRNN+VeBRNN model for Dataset 3, where both fidelity levels are noisy (see also \Cref{fig:aleatoric_uncertainty_sca} for visualizing the noise from both fidelities). In this section, we present the results for the same dataset but considering two additional MF models: (1) a fully deterministic training MF model composed of RNN+RNN shown in \Cref{fig:mf_deter_mf_problem_2}; and (2) a RNN+VeBRNN model where the HF prediction is Bayesian but the LF is deterministic (\Cref{fig:mf_uq_dns_rve_dns_sve_rnn_vebrnn}). For this more challenging scenario, using a lower-quality LF model (i.e., $N_\text{cluster} = 3$) does not lead to a better MF model when compared to the single-fidelity case of an RNN trained solely on HF data. Even with a better LF model (i.e., $N_\text{cluster} = 18$), the benefits remain small, observing only marginal improvements within the range of $T_c = 20$ to $T_c =80$.

\newpage
\section{Hyperparameter settings}
\label{sec:hyper_params}

Bayesian inference is often considered challenging due to the difficulty of optimizing the ELBO in VI, and the frequent occurrence of low mixture rates in MCMC methods. In this paper, we mitigate these challenges by employing the pSGLD sampler. For the hyperparameter settings, we draw on insights from prior studies that address similar problems in a deterministic setting~\cite{Mozaffarplasticity2019, Dekhovich2023}, and define our configurations as follows:

\begin{itemize}
    \item \textbf{Architectures:}
    \begin{itemize}
        \item \textbf{Mean network:} A two-layer GRU architecture with 128 hidden units per layer.
        \item \textbf{Variance network:} A smaller GRU network with one layer and 8 hidden units. Note that this network is only used when the cooperative training strategy is activated.
    \end{itemize}

    \item \textbf{Optimizer / Inference:}
    \begin{itemize}
        \item We use \textit{Adam} with a learning rate of $0.001$ for 1000 epochs during deterministic training and the warm-up phase of the cooperative training strategy.
        \item For variance network training: 
        \begin{itemize}
            \item Datasets with SVEs: learning rate $0.01$, 4000 epochs.
            \item Datasets with RVEs: learning rate $0.01$, 10000 epochs.
        \end{itemize}
        \item For posterior sampling via pSGLD, the learning rate is set to be $0.001$; and we use a burn-in of 50 epochs and draw 100 posterior samples every 10 epochs, totaling 1050 epochs.
    \end{itemize}

    \item \textbf{Training / Validation:}
    \begin{itemize}
        \item For deterministic training, the dataset is split into $80\%/20\%$ for training and validation, respectively.
        \item For cooperative training, no validation split is needed. We fix the iteration count to $K=2$ for all experiments.
    \end{itemize}
\end{itemize}

\end{document}